%% file: main.tex
\documentclass{article}

    \PassOptionsToPackage{numbers, compress}{natbib}

   \usepackage[preprint]{neurips_2024}

\input{00_preamble}

\usepackage[utf8]{inputenc} 
\usepackage[T1]{fontenc}    
\usepackage{hyperref}       
\usepackage{url}            
\usepackage{booktabs}       
\usepackage{amsfonts}       
\usepackage{nicefrac}       
\usepackage{microtype}      
\usepackage{wrapfig}

\definecolor{titlegold}{RGB}{237, 178, 13}
\definecolor{titlepurple}{RGB}{112, 48, 160}
\definecolor{titlecyan}{RGB}{0, 176, 240}
\definecolor{titlepink}{RGB}{215, 49, 136}

\title{{\textsc{\textcolor{titlegold}{C}\textcolor{titlepurple}{t}\textcolor{titlecyan}{r}\textcolor{titlepink}{l}}}-Adapter: An Efficient and Versatile Framework for 
Adapting
Diverse
Controls
to 
Any
Diffusion
Model
}

\newcommand*\samethanks[1][\value{footnote}]{\footnotemark[#1]}

\author{
Han Lin\thanks{equal contribution}
\quad
Jaemin Cho\samethanks{}
\quad
Abhay Zala
\quad
Mohit Bansal\\
UNC Chapel Hill\\
\texttt{\{hanlincs, jmincho, aszala,  mbansal\}@cs.unc.edu} \\
{\tt \normalsize \href{https://ctrl-adapter.github.io}{https://ctrl-adapter.github.io}}
}

\begin{document}

\maketitle

\vspace{-20pt}

\begin{figure}[!htb]
\centering
  \includegraphics[width=0.95\linewidth]{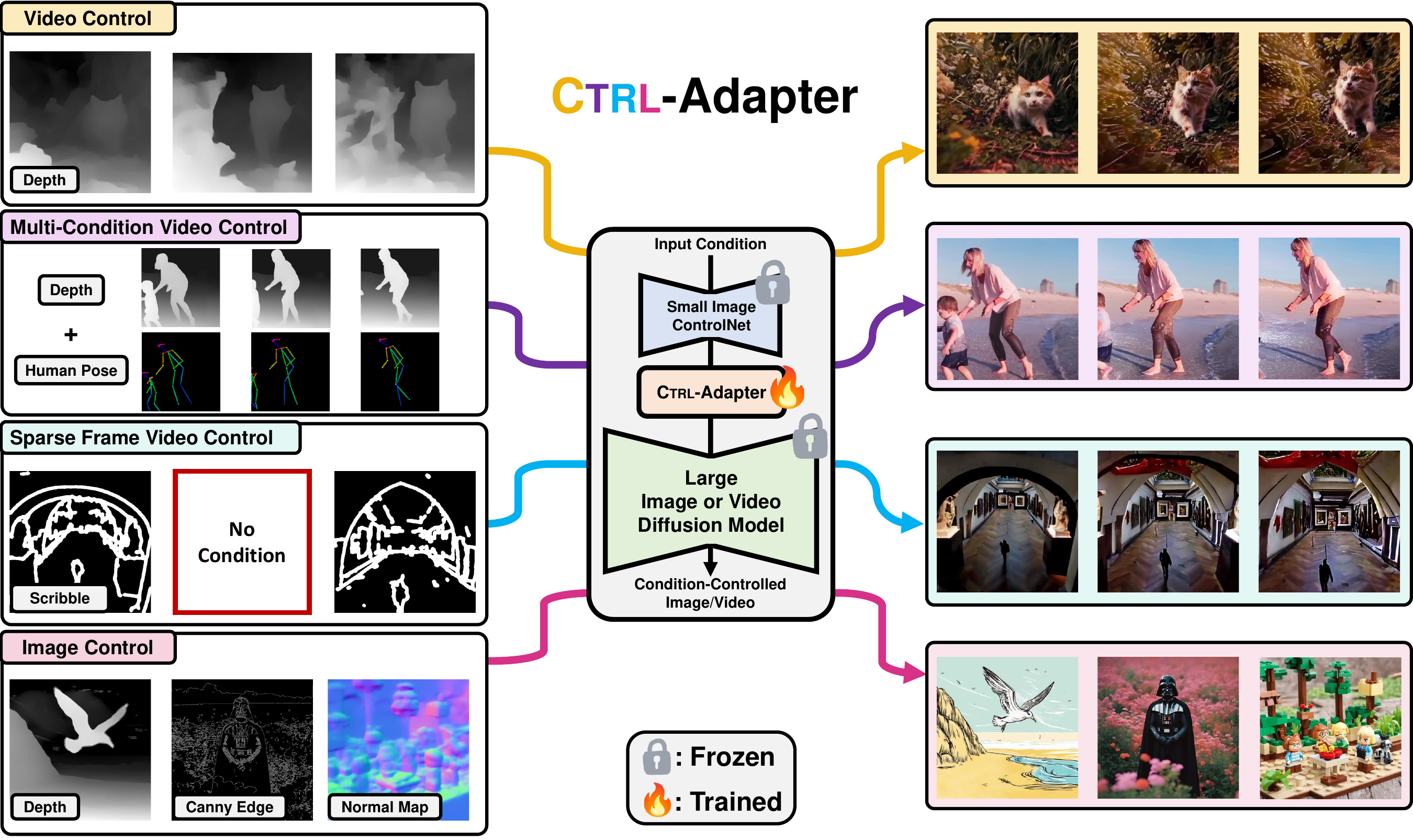}
  \vspace{-2.5mm}
  \caption{
  We propose \textbf{\adaptermethod{}}, an efficient and versatile framework for adding diverse controls to any diffusion model.
  \adaptermethod{} supports a variety of useful applications.
  }
  \label{fig:teaser_video_image_examples}
\end{figure}

\begin{abstract}
\cnet{}s are widely used for adding spatial control to text-to-image diffusion models with different conditions, such as depth maps, scribbles/sketches, and human poses. However, when it comes to controllable video generation, \cnet{}s cannot be directly integrated into new backbones due to feature space mismatches, and training \cnet{}s for new backbones can be a significant burden for many users. Furthermore, applying \cnet{}s independently to different frames cannot effectively maintain object temporal consistency.
To address these challenges,
we introduce \textbf{\adaptermethod{}},
\textit{an efficient and versatile framework that adds diverse controls to any image/video diffusion model through the adaptation of pretrained \cnet{}s.}
\adaptermethod{} offers strong and diverse capabilities, including image and video control, sparse-frame video control, {fine-grained patch-level multi-condition control (via an MoE router)}, zero-shot adaptation to unseen conditions, {and supports a variety of downstream tasks beyond spatial control, including video editing, video style transfer, and text-guided motion control.}
With {six diverse U-Net/DiT-based} image/video diffusion models (SDXL, PixArt-$\alpha$, I2VGen-XL, SVD, {Latte, Hotshot-XL}), \adaptermethod{} matches the performance of pretrained \cnet{}s on COCO and achieves the state-of-the-art on DAVIS 2017 with significantly lower computation ($<$ 10 GPU hours).
\end{abstract}

\section{Introduction}
\label{sec:intro}

Recent diffusion models have achieved significant progress
in generating
high-fidelity
images~\cite{rombach2022high, podell2023sdxl,Saharia2022Imagen,Ramesh2022UnCLIP}
and videos~\cite{blattmann2023stable, girdhar2023emu, chen2024videocrafter2,lin2023videodirectorgpt, long2024videodrafter}
from text descriptions.
As it is often hard to describe every image/video detail only with text,
there have been many works to control diffusion models in a more fine-grained manner
by providing additional condition inputs
such as bounding boxes~\cite{li2023gligen,yang2022reco},
reference object images~\cite{Ruiz2023Dreambooth,Gal2023TextualInversion,Li2023BLIP-Diffusion},
and
segmentation maps~\cite{Gafni2022Make-A-Scene,Avrahami2023SpaText,zhang2023adding}.
Among them,
Zhang~\etal{}~\cite{zhang2023adding} have released a variety of \cnet{} checkpoints based on Stable Diffusion~\cite{rombach2022high} v1.5 (SDv1.5), and the user community has shared many \cnet{}s trained with different input conditions.
Until now, \cnet{} has become one of the most popular methods for controllable image generation.

However, there are challenges when using the
existing pretrained image \cnet{}s
for controllable
video generation.
First, pretrained \cnet{}a cannot be directly plugged into new {backbone} models, and
the cost for training \cnet{}s for new backbone models is a big burden for many users due to high computational costs.
For example, training a \cnet{} for SDv1.5 takes 500-600 A100 GPU hours~\cite{ControlNet_Depth, ControlNet_Canny}.
Second, \cnet{} was originally designed for controllable image generation; hence, applying pretrained image \cnet{}s directly to each video frame independently does not take the temporal consistency across frames into account.

\begin{figure}[t]
  \centering
  \includegraphics[width=.98\linewidth]{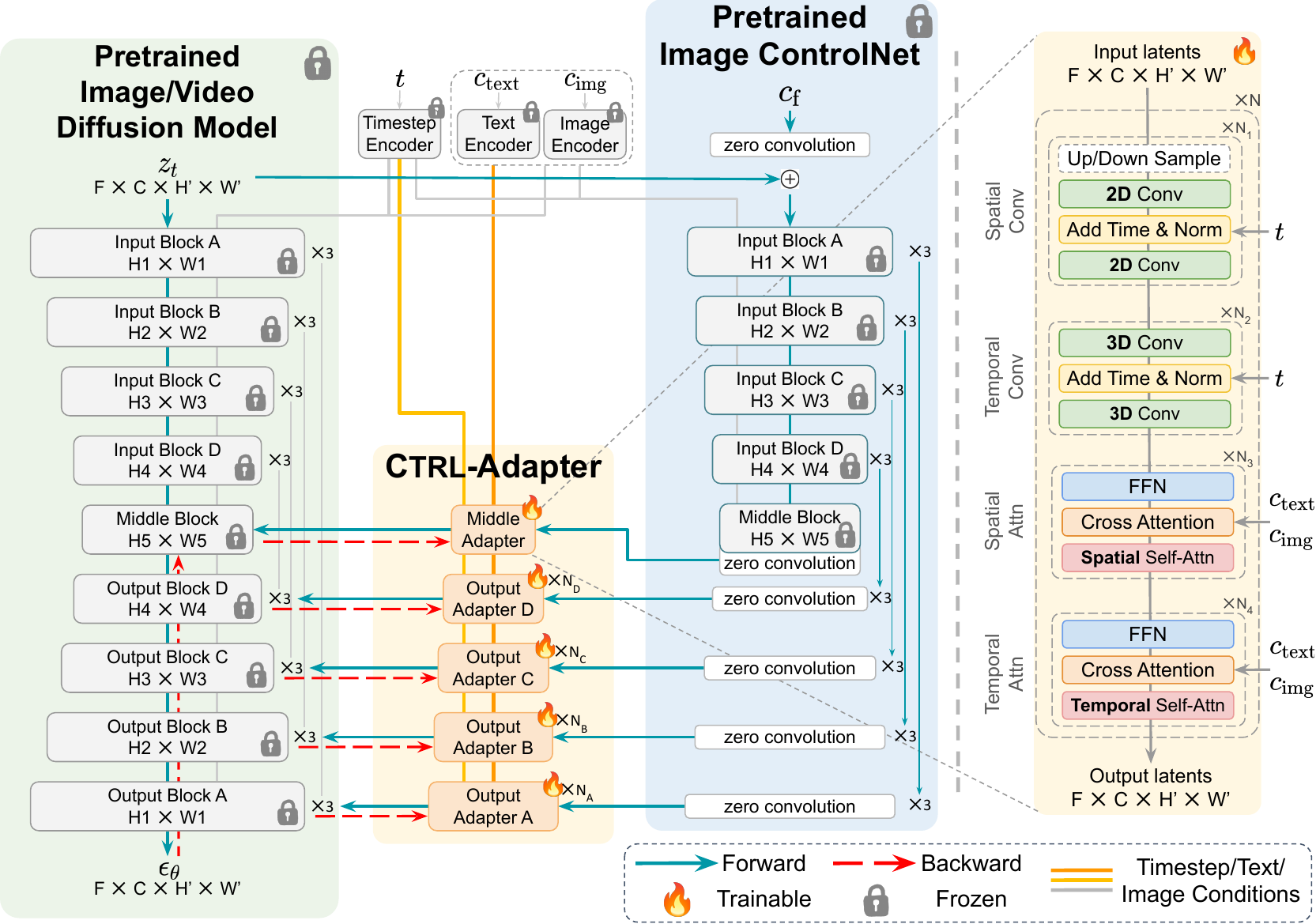}
  \vspace{-2mm}
  \caption{
  \textbf{Left:} \adaptermethod{}  (colored {\color{orange}{orange}}) enables to reuse pretrained image \cnet{}s (colored {\color{cyan}{blue}})
  for new image/video diffusion models (colored {\color{Green}{green}}). 
  \textbf{Right:} Architecture details of \adaptermethod{}. 
  Temporal convolution and attention layers are skipped for image diffusion backbones.
  }
\label{fig:adapter_method} 
\end{figure}

To address this challenge,
we design \textbf{\adaptermethod{}},
\textit{a
novel, flexible
framework that enables the efficient reuse of pretrained \cnet{}s for diverse controls with any new image/video
diffusion models,
by
adapting
pretrained \cnet{}s
(and improving temporal alignment for videos)}.
We illustrate the overall capabilities of \adaptermethod{} framework in \cref{fig:teaser_video_image_examples}.
As shown in \cref{fig:adapter_method}
left,
\adaptermethod{} trains
adapter layers~\cite{Houlsby2019ParameterEfficientTL,sung2022vladapter} to map the features of a pretrained image \cnet{} to a target image/video diffusion model, while keeping the parameters of the \cnet{} and the backbone diffusion model frozen.
As shown in \cref{fig:adapter_method} right,
each \adaptermethod{} consists of four modules: spatial convolution, temporal convolution, spatial attention, and temporal attention. 
The temporal convolution/attention modules effectively fuse the \cnet{} features into image/video diffusion models for better temporal consistency.
Additionally, to ensure robust adaptation of \cnet{}s to backbone models of different noise scales and sparse frame control conditions, we propose skipping the visual latent variable from the \cnet{} inputs. We also introduce inverse timestep sampling to effectively adapt \cnet{}s to new backbones equipped with continuous diffusion timestep samplers.
{For more accurate control beyond a single condition, we designed a novel and powerful Mixture-of-Experts (MoE) router, which allows {fine-grained, patch-level}
composition of spatial feature maps from multiple control conditions via \adaptermethod{}s (see \cref{subsec:method_multi_source}).}

As shown in \Cref{table:comparison_of_model_abilities},
\adaptermethod{} allows many useful capabilities, including image control, video control, video control with sparse frames, multi-condition control, and compatibility with different backbone models,
while previous methods only support a small subset of them (see details in \cref{sec:related_work}).
We demonstrate the effectiveness of \adaptermethod{} through extensive experiments and analyses. It exhibits strong performance when adapting \cnet{}s (pretrained with SDv1.5) to various video and image diffusion backbones, including image-to-video generation -- I2VGen-XL~\cite{zhang2023i2vgen} and Stable Video Diffusion (SVD)~\cite{blattmann2023stable},
text-to-video generation -- Latte~\cite{ma2024latte} and Hotshot-XL~\cite{Mullan_Hotshot-XL_2023},
and text-to-image generation --  SDXL~\cite{podell2023sdxl} and PixArt-$\alpha$~\cite{chen2024pixartalpha}.
{The ability of \adaptermethod{} to seamlessly adapt to DiT-based models such as Latte and PixArt-$\alpha$, which are structurally different from U-Net based \cnet{}s, demonstrates the flexibility of our framework design.}

\begin{figure}[t]
\minipage{0.49\textwidth}
  \includegraphics[width=\linewidth]{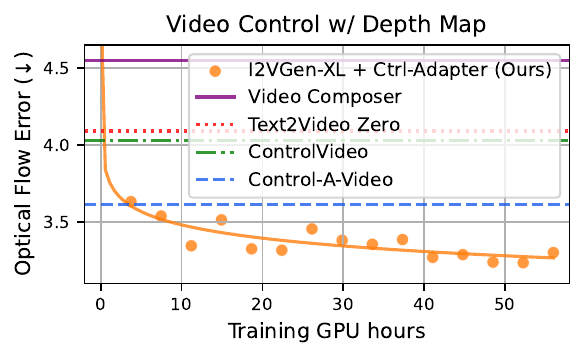}
\endminipage
\minipage{0.49\textwidth}
  \includegraphics[width=\linewidth]{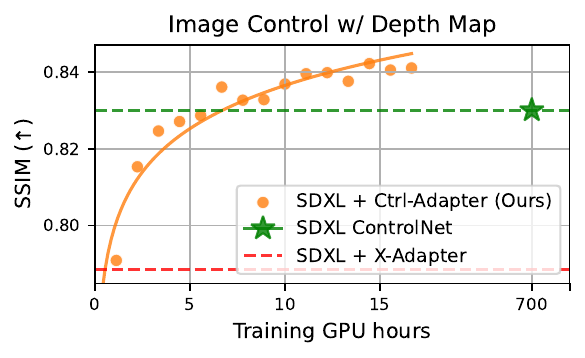}
\endminipage\hfill
  \vspace{-2mm}
  \caption{
 Training speed of \adaptermethod{} for video (left) and image (right) control with depth maps, measured on A100 80GB GPUs. For both video and image controls, \adaptermethod{} trained for 10 GPU hours outperforms strong baselines, including SDXL, which is trained for 700 GPU hours.
  }
  \label{fig:training_statistics_comparison}
\end{figure}

In \cref{sec:results_video_gen_single} and \cref{sec:results_image_gen_single}, we first show that \adaptermethod{} matches the performance of a pretrained image \cnet{} on COCO dataset~\cite{lin2014microsoft} 
and outperforms previous methods in controllable video generation (achieving state-of-the-art performance on the \davis{} 2017 dataset~\cite{pont20172017}) with significantly lower training costs (less than 10 GPU hours, see \cref{fig:training_statistics_comparison}).
Next, we demonstrate that \adaptermethod{} enables more accurate video generation with multiple conditions compared to a single condition.
{Our fine-grained patch-level MoE router consistently outperforms both the equal weights baseline and the global weights MoE router (\cref{sec:results_video_gen_multi}).}
In addition,
we show that skipping the visual latent variable from \cnet{} inputs allows video control only with a few frames of (\ie{}, sparse) conditions, eliminating the need for dense conditions across all frames (\cref{sec:results_sparse_control}).
We also highlight zero-shot adaption -- \adaptermethod{} trained with one condition can easily adapt to another \cnet{} trained with a different condition (\cref{sec:results_zero_shot_generalization}).
{Moreover, our \adaptermethod{} can be flexibly applied to a variety of downstream tasks {beyond spatial control}, including video editing, video style transfer, and text-guided object motion control (\cref{sec:beyond_spatial_control})}.
Lastly, we provide comprehensive ablations for \adaptermethod{} design choices and qualitative examples (\cref{sec:appendix_variants_of_adapter_architecture_design}, \cref{appendix_sec:additional_quant_analysis}, and \cref{appendix_sec:additional_qualitative_analysis}).

\begin{table}[t]
    \caption{
    Overview of the capabilities supported by controllable image/video generation methods.
    }
    \label{table:comparison_of_model_abilities}
    \centering
    \resizebox{0.87\linewidth}{!}{
    \begin{tabular}{l ccccc}
        \toprule
        \multirow{2}{*}{Method}
        & Image & Video & Video Control & Multi-Condition & Compatible w/ \\ 
        & Control & Control & w/ Sparse Frames & Control & Different Backbones\\
        \midrule
        \multicolumn{1}{l}{\textcolor{gray}{Image Control Methods}} \\
        \cnet{}~\cite{zhang2023adding} & \ForestGreencheck & \redcross & \redcross & \redcross & \redcross\\
        Multi-\cnet{}~\cite{zhang2023adding} & \ForestGreencheck & \redcross & \redcross &\ForestGreencheck &  \redcross\\  
        T2I-Adapter~\cite{mou2023t2i} & \ForestGreencheck & \redcross & \redcross &\ForestGreencheck &  \redcross\\  
        Uni-\cnet{}~\cite{zhao2024uni} & \ForestGreencheck & \redcross & \redcross &\ForestGreencheck &  \redcross\\  
        X-Adapter~\cite{ran2023x} & \ForestGreencheck & \redcross & \redcross &\redcross & \ForestGreencheck\\
        \midrule
        \multicolumn{1}{l}{\textcolor{gray}{Video Control Methods}} \\
        ControlVideo~\cite{zhang2023controlvideo} & \redcross & \ForestGreencheck & \redcross &\redcross &  \redcross\\
        VideoComposer~\cite{wang2024videocomposer} & \redcross & \ForestGreencheck & \redcross &\ForestGreencheck &  \redcross\\
        SparseCtrl~\cite{guo2023sparsectrl} & \redcross & \ForestGreencheck &  \ForestGreencheck & \redcross & \redcross \\
        \midrule
        \rowcolor{lightblue}
        \adaptermethod{} (Ours)  & \ForestGreencheck & \ForestGreencheck & \ForestGreencheck & \ForestGreencheck  &  \ForestGreencheck \\
        \bottomrule
    \end{tabular}
    }
\end{table}

\section{Related Works: Adding Control to Diffusion Models}
\label{sec:related_work}
There have been many works using different types of additional inputs to control the image/video diffusion models,
such as bounding boxes~\cite{li2023gligen,yang2022reco},
reference object image~\cite{Ruiz2023Dreambooth,Gal2023TextualInversion,Li2023BLIP-Diffusion},
segmentation map~\cite{Gafni2022Make-A-Scene,Avrahami2023SpaText,zhang2023adding},
sketch~\cite{zhang2023adding},
\etc{},
and combinations of multiple conditions~\cite{kim2023diffblender,qin2023unicontrol,zhao2024uni,wang2024videocomposer}.
As finetuning all the parameters of such image/video diffusion models is computationally expensive,
several methods, such as \cnet{}~\cite{zhang2023adding}, have been proposed to add conditional control capability via parameter-efficient training~\cite{zhang2023adding,simoryu_lora_diffusion,mou2023t2i}.

X-Adapter~\cite{ran2023x} learns an adapter module to reuse \cnet{}s pretrained with a smaller image diffusion model (\eg{}, SDv1.5) for a bigger image diffusion model (\eg{}, SDXL).
While they focus solely on learning an adapter for image control, \adaptermethod{} features architectural designs (\eg{}, temporal convolution/attention layers)
for video generation as well.
In addition, X-Adapter needs to be used with the source image diffusion model (SDv1.5) during both training and inference, whereas \adaptermethod{} does not require the smaller diffusion model for image or video generation, making it more memory and computationally efficient (see \cref{sec:visualized_comparison_different_methods} for details).

SparseCtrl~\cite{guo2023sparsectrl} guides a video diffusion model with conditional inputs of few frames (instead of full frames), to alleviate the cost of collecting video conditions.
Since SparseCtrl involves augmenting \cnet{} with an additional channel for frame masks, it requires training a new variant of \cnet{} from scratch. 
In contrast, we leverage existing image \cnet{}s more efficiently by propagating information through temporal layers in adapters and enabling sparse frame control via skipping the latents from \cnet{} inputs (see \cref{subsec:method_adapter} for details).

Furthermore, compared with previous works that are specially designed for specific condition controls on a single modality (image~\cite{zhang2023adding, qin2023unicontrol} or video~\cite{hu2023videocontrolnet, zhang2023controlvideo}),
our work presents a unified and versatile framework that supports diverse controls, including image control, video control, sparse frame control, 
with significantly lower computational costs by reusing pretrained \cnet{}s (outperforms strong baselines in less than 10 GPU hours, see \cref{fig:training_statistics_comparison}).
\added{To the best of our knowledge, we are also the first work that extends multi-condition video control into fine-grained patch-level composition.}
\Cref{table:comparison_of_model_abilities} compares \adaptermethod{} with other relevant methods.
See \cref{sec:appendix_related_works} for extended related works.

\section{Method}
\label{sec:method}

\subsection{Preliminaries: Latent Diffusion Models and ControlNets}
\label{subsec:method_preliminaries}

\paragraph{Latent Diffusion Models.}
Many recent video generation works utilize latent diffusion models (LDMs)~\cite{rombach2022high} to learn the compact representations of videos.
First, 
given a $F$-frame RGB video $\bm{x}\in\mathbb{R}^{F\times 3 \times H \times W}$,
a video encoder (of a pretrained autoencoder) provides $C$-dimensional latent representation (\ie{}, latents):
$\bm{z}=\mathcal{E}(\bm{x})\in\mathbb{R}^{F\times C \times H' \times W'}$,
where height and width are spatially downsampled
($H' < H$ and $W' < W$).
Next,
in the forward process,
a noise scheduler (\eg{}, DDPM~\cite{ho2020denoising}) adds noise to the latents $\bm{z}$.
Then, in the backward pass, a diffusion model
$\bm{\mathcal{F}_\theta}(\bm{z}_t, t, \bm{c}_\text{text/img})$ learns to gradually denoise the latents,
given 
a diffusion timestep $t$, and
a text prompt $\bm{c}_\text{text}$ (\ie{}, T2V)
and/or an initial frame $\bm{c}_\text{img}$ (\ie{}, I2V) if provided.
The diffusion model is trained with objective: 
$\mathcal{L}_{\text{LDM}}=\mathbb{E}_{\bm{z}, \bm{\epsilon}\sim N(0, \bm{I}), t}\|\bm{\epsilon} - \bm{\epsilon_{\theta}}(\bm{z}_t, t, \bm{c}_\text{text/img})\|_2^2
$, where $\bm{\epsilon}$ and $\bm{\epsilon_\theta}$ represent the added noise to latents and the predicted noise by $\bm{\mathcal{F}_{\theta}}$ respectively. We apply the same objective for \adaptermethod{} training.

\paragraph{\cnet{}s.} 
\cnet{}~\cite{zhang2023adding} is designed to add spatial controls (\eg{}, depth, sketch, segmentation maps, \etc{}) to image diffusion models.
Specifically,
given a pretrained backbone image diffusion model
$\bm{\mathcal{F}_{\theta}}$
that consists of input/middle/output blocks,
\cnet{} has a similar architecture $\bm{\mathcal{F}_{\theta'}}$, where
the input/middle blocks parameters of $\bm{{\theta'}}$ are initialized from $\bm{{\theta}}$,
and the output blocks consist of $1\times1$ convolution layers initialized with zeros.
\cnet{} takes the
diffusion timestep $t$,
text prompt $\bm{c}_\text{text}$, control image $\bm{c}_\text{f}$ (\eg{}, depth map), and the noisy latents $\bm{z_t}$ as inputs,
and the output features are merged into the backbone model $\bm{\mathcal{F}_\theta}$ for final image generation.

\subsection{\adaptermethod{}}
\label{subsec:method_adapter}

We introduce \textbf{\adaptermethod{}}, a novel framework that enables the efficient reuse of existing image \cnet{}s (SDv1.5) for spatial control with new diffusion models.
{\sl We mainly describe our method details in the video generation settings, since \adaptermethod{} can be flexibly adapted to image diffusion models by regarding images as single-frame videos.}

\noindent\textbf{Efficient adaptation of {pretrained} \cnet{}s.}
As shown in \cref{fig:adapter_method} (left),
we train an
adapter module (colored {\color{orange}{orange}}) to map the middle/output blocks of a pretrained \cnet{} (colored {\color{cyan}{blue}}) to the corresponding middle/output blocks of the target video diffusion model (colored {\color{Green}{green}}).
If the target backbone does not have the same number of output blocks
\adaptermethod{} maps the \cnet{} features to the output block that handles the closest height and width of the latents.
We keep all parameters in both the \cnet{} and the target video diffusion model frozen.
Therefore, training a \adaptermethod{} can be significantly more efficient than training a new video \cnet{}.

\noindent\textbf{\adaptermethod{} architecture.}
As shown in \cref{fig:adapter_method} (right),
each block of \adaptermethod{}
consists
of four modules:
spatial convolution, temporal convolution, spatial attention, and temporal attention.
We set the values for $N_1, ..., N_4$ and $N$ as 1 by default.
The temporal convolution and attention modules effectively fuse the \cnet{} features to the video backbone models for better temporal consistency.
Moreover,
the spatial/temporal convolution modules
incorporate the current denoising timestep $t$ 
and
spatial/temporal attention modules
incorporate the conditions (\ie{}, text prompt/initial frame) $\bm{c}_\text{text/img}$. This design allows \adaptermethod{} to dynamically adjust its features according to different denoising stages and the objects generated.
In addition, we skip the temporal convolution/attention modules when adapting to image diffusion models.
See \cref{sec:adapter_architecture_detail} for architecture details of the four modules, and \cref{sec:appendix_variants_of_adapter_architecture_design} for detailed ablation studies on the design choices of \adaptermethod{}.

\noindent\textbf{Adaptation to DiT-based image/video backbones.}
\added{Our \adaptermethod{} can also adapt U-Net based \cnet{}s
to DiT-based image/video generation backbones. One important observation we made is that the spatial features encoded in the U-Net of \cnet{}s and the DiT blocks are structurally different (see \cref{fig:feature_map_visualization}).
Specifically, the representation from U-Net blocks exhibits coarse-to-fine, hierarchical patterns (\eg{}, earlier blocks output smaller size feature maps and control high-level information such as object presence, while later blocks output larger feature maps and control lower-level details like textures), while all DiT blocks handle the feature maps of same sizes.
This indicates that mapping all middle/output blocks of \cnet{} to DiT blocks might not be the optimal solution.
Therefore, we choose to map the feature maps of the largest size in \cnet{} (\ie{}, block A) to the DiT blocks via \adaptermethod{}s, which are followed by zero-convolutions for channel dimension matching. To improve computational efficiency for DiT-based video generation models (\ie{}, Latte~\cite{ma2024latte}), we only insert \adaptermethod{}s into every other DiT block (\ie{}, blocks 2, 4, 6..., 28, see (a) in \cref{fig:method_dit_adapter_latte}).
See \cref{sec:appendix_dit_based_backbones}
for more discussion on \adaptermethod{} designs for DiT.
}

\begin{wrapfigure}[13]{r}{0.5\textwidth}
\vspace{-16pt}
  \centering
  \includegraphics[width=\linewidth]{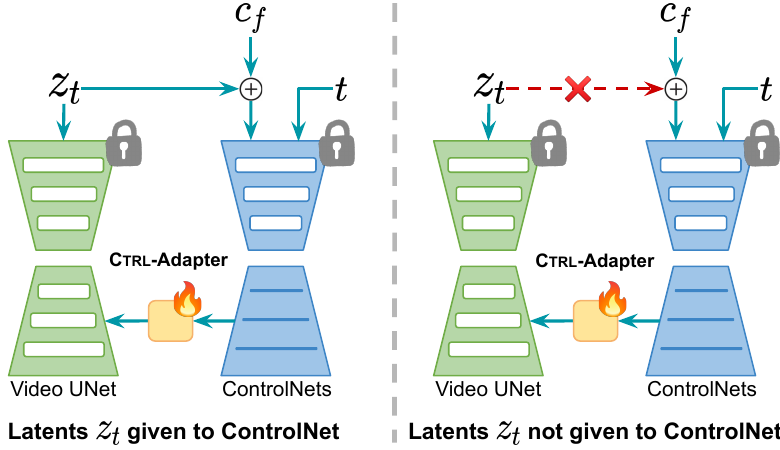}
  \vspace{-6mm}
  \caption{
  \textbf{Left (default):} latent $\bm{z}_t$ is given to \cnet{}.
  \textbf{Right:} latent $\bm{z}_t$ not given to \cnet{}.
  }
\label{fig:method_skip_input} 
\end{wrapfigure}

\noindent\textbf{Skipping the latent from \cnet{} inputs: robust adaption to different noise scales \& sparse frame conditions.}
Although the original \cnet{}s take the latent $\bm{z}_t$ as part of their inputs,
we find that
skipping $\bm{z}_t$ from \cnet{} inputs is effective for \adaptermethod{} in certain settings,
as illustrated in \cref{fig:method_skip_input}.
\textbf{(1) Different noise scales:}
while SDv1.5 samples noise $\bm{\epsilon}$ from $N(\bm{0}, \bm{I})$,
some recent diffusion models~\cite{hoogeboom2023simple,esser2024scaling,blattmann2023stable}
sample noise $\bm{\epsilon}$ of much bigger scale (\eg{} SVD~\cite{blattmann2023stable} sample noise from $\sigma * N(\bm{0}, \bm{I})$, where $\sigma\sim \text{LogNormal}(0.7, 1.6)$; $\sigma\in[0, +\infty]$ and $\mathbb{E}[\sigma]=7.24$).
We find that adding larger-scale $\bm{z}_t$ from the new backbone models to image conditions $\bm{c}_f$ dilutes the $\bm{c}_f$ and makes the \cnet{} outputs less informative,
whereas skipping $\bm{z}_t$ enables the adaptation of such new backbone models.
\textbf{(2) Sparse frame conditions:}
when the image conditions are provided only for the subset of video frames (\ie{}, $\bm{c}_f=\emptyset$ for most frames $f$),
\cnet{} could rely on the information from $\bm{z}_t$ and ignore $\bm{c}_f$ during training.
Skipping $\bm{z}_t$ from \cnet{} inputs also helps the \adaptermethod{} to more effectively handle such sparse frame conditions (see \Cref{table:ablation_skip_latent}).

\noindent\textbf{Inverse timestep sampling: robust adaptation to continuous diffusion timestep samplers.}
While SDv1.5 samples discrete timesteps $t$ uniformly from $\{0, 1, ... 1000\}$,
some recent diffusion models~\cite{esser2024scaling,ma2024sit,stablediffusion} sample timesteps from continuous distributions, \eg{}, SVD~\cite{blattmann2023stable} 
samples timesteps from a \text{LogNormal} distribution.
This gap between discrete and continuous distributions means that we cannot assign the same timestep $t$ to both the video diffusion model and the \cnet{}.
Therefore, we propose {inverse timestep sampling}, an algorithm that {\sl{creates a timestep mapping between the continuous and discrete time distributions}}
(see \cref{alg:timestep_sampling} for PyTorch~\cite{Ansel_PyTorch_2_Faster_2024} code).
\added{The high-level idea of this algorithm is inspired by inverse transform sampling~\cite{Inverse_transform_sampling}. Given the cumulative distribution functions (CDFs) of the continuous timestep distribution $F_\text{cont.}$ and the \cnet{} timestep distribution $F_\text{CNet}$,
we first uniformly sample a value $u$ between $[0, 1]$, and then returns the smallest timesteps $t_\text{cont.}\in[0, \infty]\subseteq\mathbb{R}, t_\text{CNet}\in\{0, 1, ..., 1000\}\subseteq\mathbb{N}$, such that $F_\text{cont.}(t_\text{cont.})\geq u, F_\text{CNet}(t_\text{CNet})\geq u$. This procedure naturally creates a mapping between two distributions. See \cref{sec:pseudocode_inverse_timestep_sampling} for details.}

\begin{figure}[h]
  \centering
  \includegraphics[width=0.99\linewidth]{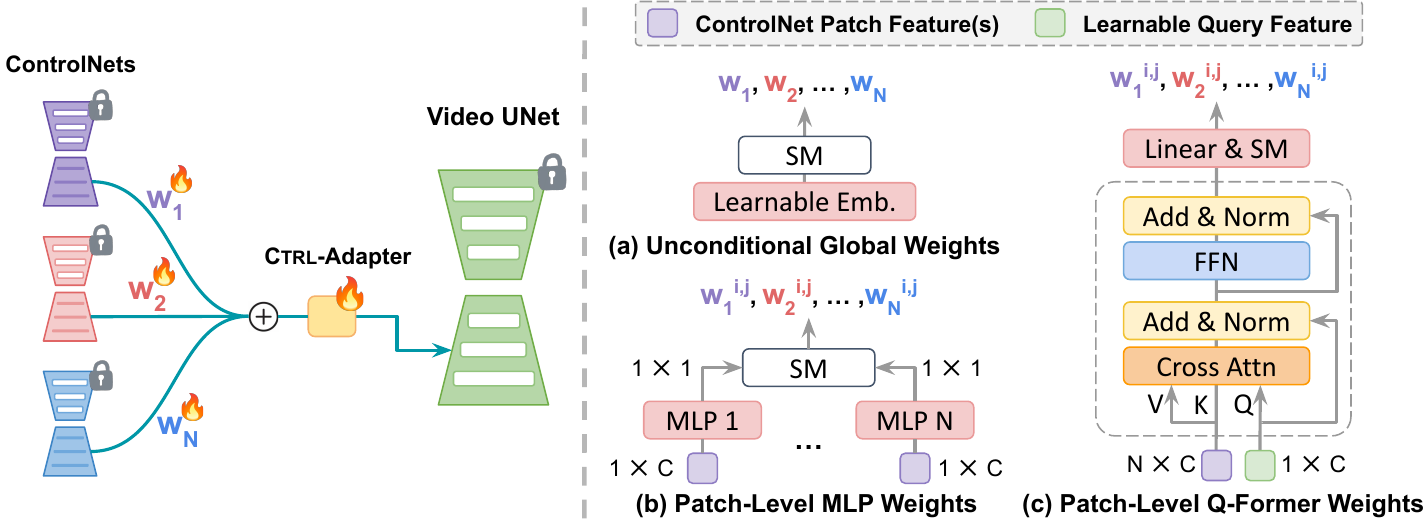}
\vspace{-5pt}
  \caption{
  \textbf{Left:} Framework for multi-condition video generation by combining multiple \cnet{}s. {$w_1, w_2, ..., w_N$ are the weights allocated to each \cnet{}.}
  \textbf{Right:} Three MoE router variants. (a) operates globally,
  while (b) and (c) operate on the fine-grained patch-level. $C$ and $N$ represent feature dimensions and number of \cnet{} experts respectively. {$w_k^{i,j}$ represents the router weights at position ($i, j$) of the $k^\text{th}$ \cnet{} 2D feature map.} SM stands for Softmax.
  }
\label{fig:multi_source_method} 
\end{figure}

\subsection{Multi-Condition Generation via \adaptermethod{} Composition}
\label{subsec:method_multi_source}

Multi-\cnet{}~\cite{zhang2023adding} is proposed for spatial control beyond a single condition. However, this method naively combines different conditions with equal weights during inference time without training. 
\added{For more effective control composition, we first experiment with some simple extensions, such as replacing these fixed weights with unconditional global learnable weights via a lightweight MoE~\cite{Shazeer2017MoE} router (see variant (a) in \cref{fig:multi_source_method} right).
Next, we further propose refining the global weights MoE router into a more fine-grained, patch-level MoE router.
Variant (b) processes the patch-level features of each \cnet{} into a scalar value independently, then uses softmax to assign \cnet{} weights for each patch. Variant (c) takes in all $N$ \cnet{} features associated with a patch, using an architecture design inspired by Q-Former~\cite{li2023blip} to output expert weights.
Comparisons of different variants are discussed in \cref{sec:results_video_gen_multi} and \cref{appendix_sec:different_weighting_schemes}.
}

\section{Experimental Setup}
\label{sec:experiment_setups}

\noindent\textbf{\cnet{}s and Target Diffusion Models.}
We use \cnet{}s trained with SD v1.5.
For target diffusion models,
we experiment with 
two I2V models -- I2VGen-XL~\cite{zhang2023i2vgen} and Stable Video Diffusion (SVD)~\cite{blattmann2023stable},
two T2V models -- Latte~\cite{ma2024latte} and Hotshot-XL~\cite{Mullan_Hotshot-XL_2023},
and two T2I models --  SDXL~\cite{podell2023sdxl} and PixArt-$\alpha$~\cite{chen2024pixartalpha}.
Note that Latte and PixArt-$\alpha$ are generation models based on DiT instead of U-Net.

\noindent\textbf{Training and Evaluation Datasets.}
We use 200K videos sampled from Panda-70M training set~\cite{chen2024panda70m} and 300K images from the LAION POP~\cite{LAION_POP} dataset for video and image \adaptermethod{}s training respectively.
During training, we extract various control conditions (\eg{}, depth map) on-the-fly to simplify the data-preparation process. 
Following previous works~\cite{hu2023videocontrolnet,zhang2023controlvideo},
we evaluate video \adaptermethod{}s on \davis{} 2017~\cite{pont20172017}, and image \adaptermethod{}s on COCO \texttt{val2017} split~\cite{lin2014microsoft}. 
Detailed training and inference setups for the experiments are provided in \cref{sec:appendix_train_inference_details} and \cref{sec:appendix_experiment_setups}.

\noindent\textbf{Evaluation Metrics.} We perform evaluation on two folds: \textbf{visual quality} and \textbf{spatial control}.
Following previous works~\cite{qin2023unicontrol, hu2023videocontrolnet},
we use Frechet Inception Distance (FID)~\cite{heusel2017gans} to measure the visual quality of generated images/videos.
For video datasets,
following VideoControlNet~\cite{hu2023videocontrolnet}, we report the L2 distance between the optical flow error~\cite{ranjan2017optical} between the input and generated videos.
For image datasets,
following Uni-ControlNet~\cite{zhao2024uni}, we report the Structural Similarity (SSIM)~\cite{wang2004image} and mean squared error (MSE) between generated images and ground truth images.

\begin{table}[t]
    \caption{
    Evaluation of video generation with single control condition on \davis{} 2017 dataset.
    The best number in each column is \textbf{bolded}, and the second best is \underline{underscored}.
    }
    \vspace{-2mm}
    \label{table:single_source_video_results}
    \centering
    \resizebox{0.9\linewidth}{!}{
    \begin{tabular}{l cc cc cc}
        \toprule
        \multirow{2}{*}{Method}  & \multicolumn{2}{c}{Depth Map} & \multicolumn{2}{c}{Canny Edge} \\
        \cmidrule(lr){2-3} \cmidrule(lr){4-5} 
        & FID ($\downarrow$) &  Optical Flow Error ($\downarrow$) & FID ($\downarrow$) &  Optical Flow Error ($\downarrow$) \\
        \midrule
        Text2Video-Zero~\cite{khachatryan2023text2video} & 19.46 & 4.09 & 17.80 & 3.77 \\
        ControlVideo~\cite{zhang2023controlvideo}  & 27.84 & 4.03 & 25.58 & 3.73 \\
        Control-A-Video~\cite{chen2023control}  & 22.16 & 3.61 & 22.82 & 3.44  \\
        VideoComposer~\cite{wang2024videocomposer} & 22.09 & 4.55 & - & - \\
        \midrule
        \multicolumn{1}{l}{\textcolor{gray}{Hotshot-XL backbone}} \\
        SDXL \cnet{}~\cite{von-platen-etal-2022-diffusers} & 45.35 & 4.21 & 25.40 & 4.43 \\
        \rowcolor{lightblue}
        SDv1.5 \cnet{} + \adaptermethod{} (Ours) & 14.63 & 3.94 & 20.83 & 4.15 \\
        \midrule 
        \multicolumn{1}{l}{\textcolor{gray}{Latte backbone (DiT-Based)}} \\
        \rowcolor{lightblue}
        SDv1.5 \cnet{} + \adaptermethod{} (Ours) & 16.92 & 3.98 & 17.87 & {\underline{2.73}} \\
        \midrule 
        \multicolumn{1}{l}{\textcolor{gray}{I2VGen-XL backbone}} \\
        \rowcolor{lightblue}
        SDv1.5 \cnet{} + \adaptermethod{} (Ours) & 7.43 & {\underline{3.20}} & 
        {\underline{6.42}} & {3.37} \\
        \midrule 
        \multicolumn{1}{l}{\textcolor{gray}{SVD backbone}} \\
        SVD Temporal \cnet{}~\cite{strawberry_svdtemporalcontrolnet_2023}
        & {\underline{4.91}} &  4.84 & - & - \\
        \rowcolor{lightblue}
        SDv1.5 \cnet{} + \adaptermethod{} (Ours) & {\textbf{3.82}} & {\textbf{2.96}} & {\textbf{3.96}} & {\textbf{2.39}} \\
        \bottomrule
    \end{tabular}
    }
\end{table}

\section{Results and Analysis}
\label{sec:results}

\subsection{Video Generation with Single Condition}
\label{sec:results_video_gen_single}

\begin{wrapfigure}[19]{r}{0.43\textwidth}
\vspace{-90pt}
  \centering
\includegraphics[width=\linewidth]{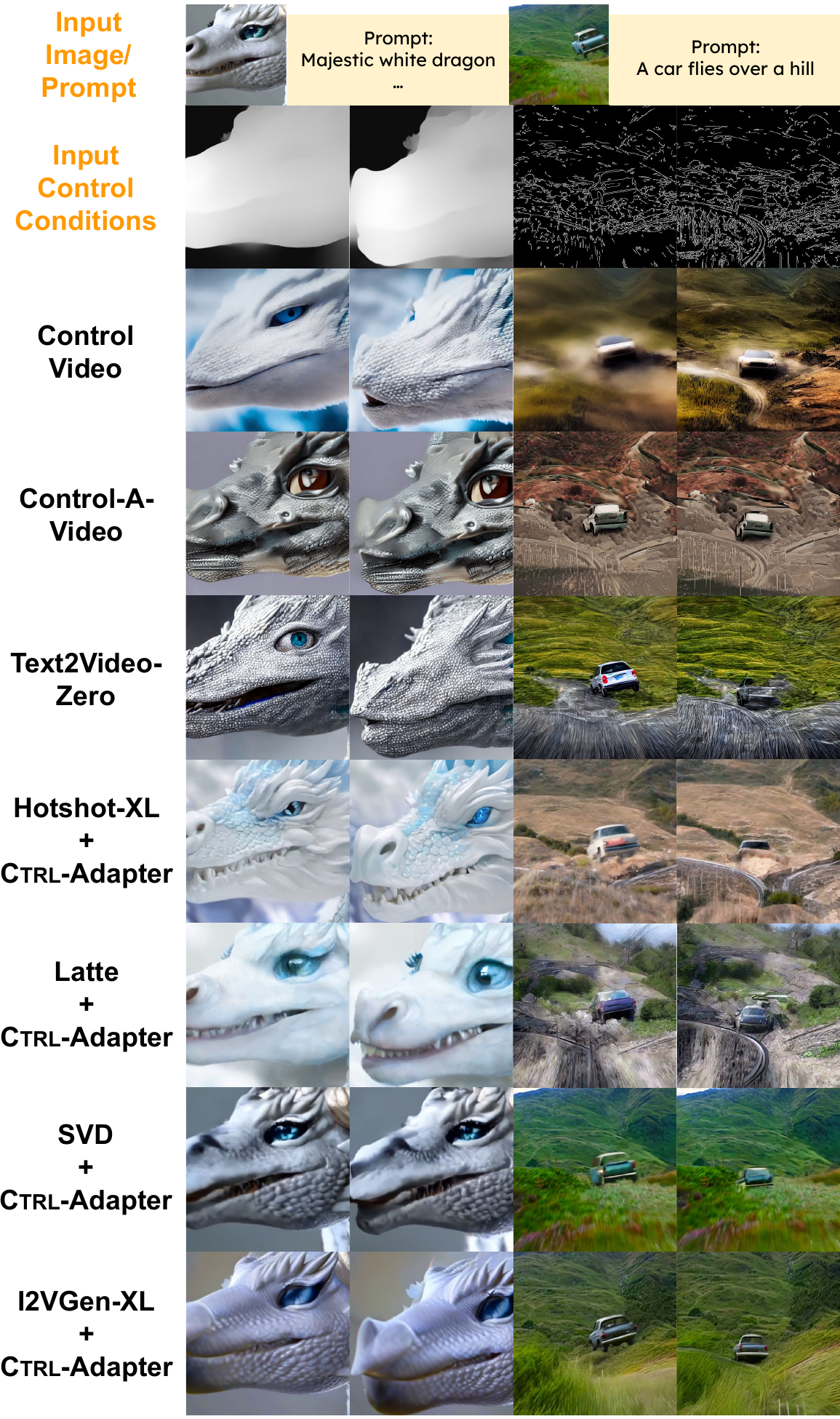}
  \vspace{-6mm}
  \caption{Single-cond. video generation.
  }
\label{fig:bigger_movements} 
\end{wrapfigure}

We compare SDv1.5 \cnet{} + \adaptermethod{} built on Hotshot-XL,  I2VGen-XL, SVD, and Latte with  video control methods including Text2Video-Zero~\cite{khachatryan2023text2video}, Control-A-Video~\cite{chen2023control}, ControlVideo~\cite{zhang2023controlvideo}, and VideoComposer~\cite{wang2024videocomposer}.
As the spatial layers of Hotshot-XL are initialized with SDXL and remain frozen, the SDXL \cnet{}s are directly compatible with Hotshot-XL, so we include Hotshot-XL + SDXL \cnet{} as a baseline.
We also experiment with
a temporal \cnet{}~\cite{strawberry_svdtemporalcontrolnet_2023} trained with SVD.

\Cref{table:single_source_video_results}
shows that
in both depth map and canny edge input conditions,
\adaptermethod{}s on I2VGen-XL and SVD
outperforms all previous strong video control methods
in visual quality (FID) and spatial control (optical flow error) metrics.
Note that it takes $<$ 10 GPU hours for \adaptermethod{} to outperform the baselines 
(see \cref{fig:training_statistics_comparison}).
{
In \cref{fig:bigger_movements} and \cref{appendix_subsec:video_gen_examples}, we visualize the comparison between \adaptermethod{} and other video control baselines.
We study visual quality-spatial control trade-off in  \cref{sec:tradeoff_visual_quality_spatial_control}. 
}

\begin{table}[t]
\vspace{-2mm}
    \caption{
    Evaluation of image generation with single control condition on COCO val2017 split. 
    The best number in each column is \textbf{bolded}, and the second best is \underline{underscored}.
    }
    \vspace{-2mm}\label{table:single_source_image_results}
    \centering
    \resizebox{0.95\linewidth}{!}{
    \begin{tabular}{l ccc cc cc cc}
        \toprule
        \multirow{2}{*}{Method}  & \multicolumn{3}{c}{Depth Map} & \multicolumn{2}{c}{Canny Edge} & \multicolumn{2}{c}{Soft Edge / HED} \\
        \cmidrule(lr){2-4} \cmidrule(lr){5-6} \cmidrule(lr){7-8} 
        & FID ($\downarrow$) & MSE ($\downarrow$) & SSIM ($\uparrow$) & FID ($\downarrow$) &  SSIM ($\uparrow$) & FID ($\downarrow$) &  SSIM ($\uparrow$) \\
        \midrule
        \multicolumn{1}{l}{\textcolor{gray}{SDv1.4 or v1.5 backbone}} \\
        SDv1.5 \cnet{}~\cite{zhang2023adding} & 21.25 &  87.57 & - & 18.90 &  0.4828 & 26.59 & 0.4719 \\
        T2I-Adapter~\cite{mou2023t2i} & 21.35 & 89.82 & - &18.98 &  0.4422 & - & -\\
        GLIGEN~\cite{li2023gligen} & 21.46  & 88.22 & - & 24.74 &  0.4226 & 28.57 & 0.4015 \\
        Uni-ControlNet~\cite{zhao2024uni} & 21.20  & 91.05 & - & {\underline{17.79}} & {{0.4911}} & {\underline{17.86}} & 0.5197 \\
        \midrule 
        \multicolumn{1}{l}{\textcolor{gray}{SDXL backbone}} \\
        SDXL \cnet{}~\cite{von-platen-etal-2022-diffusers} & {\textbf{17.91}}  &  {\underline{86.95}} & {{0.8363}} & {\textbf{17.21}} & 0.4458 & - & -\\
        SDv1.5 \cnet{} + X-Adapter~\cite{ran2023x} & 20.71 &  90.08 & 0.7885 & 19.71 &  0.3002 & - & - \\
        \rowcolor{lightblue}
        SDv1.5 \cnet{} + \adaptermethod{} (Ours) & {\underline{19.26}} & {{87.54}} & {\textbf{0.8534}} & 21.04 & {\underline{0.5806}} & 18.08 & 0.6454 \\
        \midrule
        \multicolumn{1}{l}{\textcolor{gray}{PixArt-$\alpha$ backbone (DiT-Based)}} \\
        PixArt-$\delta$ ControlNet~\cite{chen2024pixart_delta}  & - & - & - & - & - & 20.41 & {\textbf{0.6938}} \\
        \rowcolor{lightblue}
        SDv1.5 \cnet{} + \adaptermethod{} (Ours) & 22.54 & \textbf{84.78} & \underline{0.8496} & 18.75 & \textbf{0.6359} & {\textbf{17.52}} & {\underline{0.6812}} \\
        \bottomrule
    \end{tabular}
    }
\end{table}

\begin{wrapfigure}[10]{r}{0.55\textwidth}
\vspace{-18pt}
  \centering
  \includegraphics[width=\linewidth]{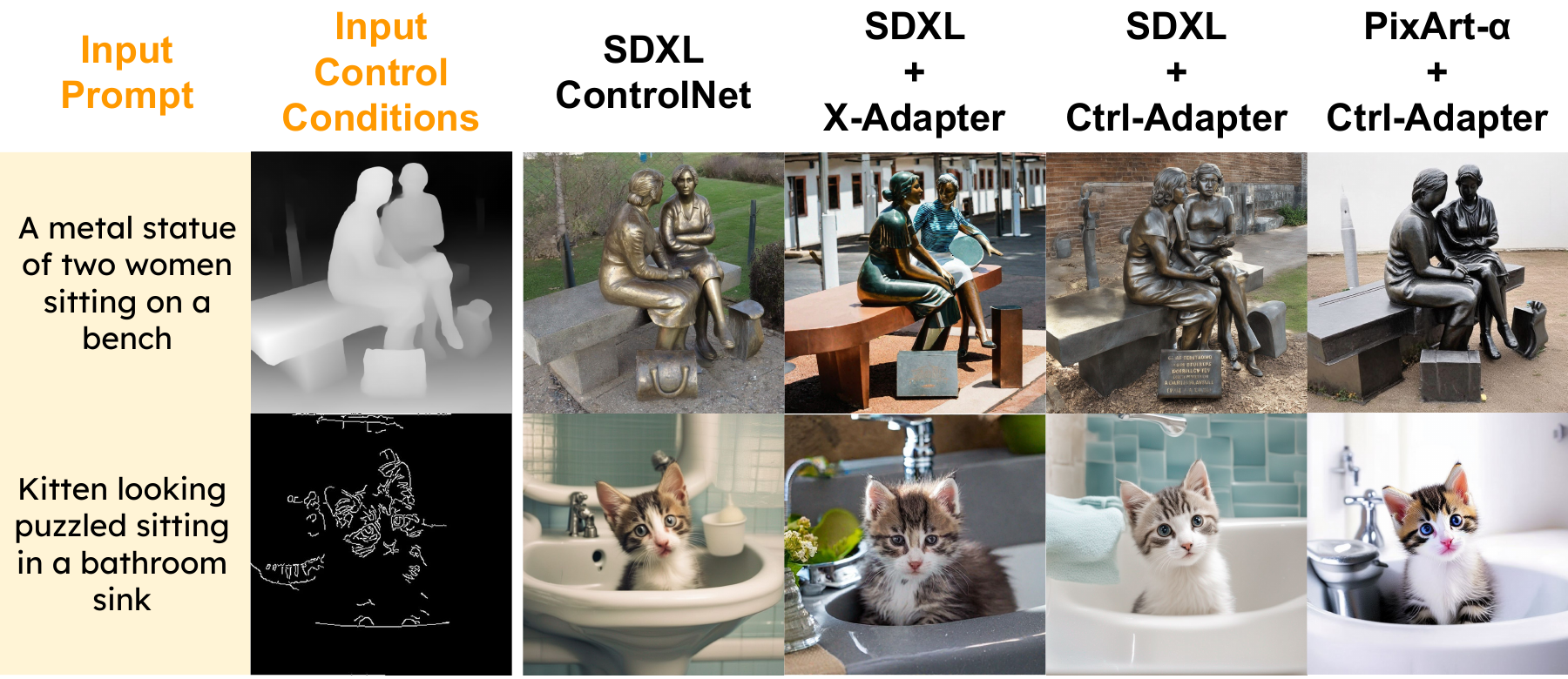}
  \vspace{-6mm}
  \caption{Generated images on COCO val2017 split.
  }
\label{fig:single_source_image_generation_examples} 
\end{wrapfigure}

\subsection{Image Generation with Single Condition}
\label{sec:results_image_gen_single}
We compare SDv1.5 \cnet{} + \adaptermethod{} with controllable image generation methods that use SDv1.4, SDv1.5, SDXL, and PixArt-$\alpha$ as backbones, 
including 
pretrained SDv1.5/SDXL \cnet{}s~\cite{zhang2023adding, von-platen-etal-2022-diffusers}, 
T2I-Adapter~\cite{mou2023t2i}, 
GLIGEN~\cite{li2023gligen},
Uni-ControlNet~\cite{zhao2024uni},
X-Adapter~\cite{ran2023x},
and
PixArt-$\delta$ ControlNet~\cite{chen2024pixart_delta}.

As shown in \Cref{table:single_source_image_results},
\adaptermethod{}
outperforms baselines with SDv1.4/v1.5 backbones in almost all metrics.
When compared to the baselines with SDXL backbones,
\adaptermethod{} outperforms X-Adapter in most metrics,
and matches (in FID/MSE with depth map inputs) or outperforms SDXL \cnet{} (in SSIM with depth map and canny edge inputs).
Note that
SDXL \cnet{} was trained for much longer than \adaptermethod{} (700 \textit{vs.} 44 A100 GPU hours)
and it takes less than 10 GPU hours for \adaptermethod{}
to outperform the SDXL depth \cnet{} in SSIM (see \cref{fig:training_statistics_comparison}).
\added{In addition, when applied to DiT-based backbone (\ie{}, PixArt-$\alpha$), \adaptermethod{} achieves good
improvement in FID (17.52 ours \textit{vs.} 20.41 PixArt-$\delta$ \cnet{} on soft edge)
and competitive SSIM score.
In \cref{fig:bigger_movements} and \cref{fig:single_source_image_generation_examples}, we visualize the comparison between \adaptermethod{} and other image control baselines. See \cref{appendix_subsec:image_gen_examples} for more visualizations.
}

\begin{table}[t]
    \caption{
    Comparison of different weighting methods (see \cref{fig:multi_source_method} right part for details) for
    multi-condition video generation.
    The control sources are abbreviated as D (depth map), C (canny edge), N (surface normal), S (softedge), Seg (semantic segmentation map), L (line art), and P (human pose).
    }
    \label{tab:multi-source-video-gen}
    \centering
    \resizebox{0.99\linewidth}{!}{
    \begin{tabular}{l cc cc cc cc}
        \toprule
        & \multicolumn{2}{c}{D+C} & \multicolumn{2}{c}{D+P} & \multicolumn{2}{c}{D+C+N+S} & \multicolumn{2}{c}{D+C+N+S+Seg+L+P} \\
        \cmidrule(lr){2-3} \cmidrule(lr){4-5} \cmidrule(lr){6-7} \cmidrule(lr){8-9} 
        & FID ($\downarrow$) & Flow Error ($\downarrow$) & FID ($\downarrow$) & Flow Error ($\downarrow$)& FID ($\downarrow$) & Flow Error ($\downarrow$) & FID ($\downarrow$) & Flow Error ($\downarrow$) \\
        \midrule
        {\textcolor{gray}{Baseline:}} Equal Weights  & 8.50 & 2.84 & 11.32 & 3.48 & 8.75 & 2.40 & 9.48 & 2.93 \\
        \midrule
        (a) Unconditional Global Weights  & 9.14 & 2.89 & 10.98 & 3.32 & 8.39 & 2.36 & 8.18 & 2.48 \\
        \rowcolor{lightblue}
        (b) Patch-Level MLP Weights  & 8.40 & \textbf{2.34} & 9.37 & \textbf{3.17} & 7.87 & \textbf{2.11} & 8.26 & \textbf{2.00} \\
        \rowcolor{lightblue}
        (c) Patch-Level Q-Former Weights  & \textbf{7.54} & 2.39 & \textbf{9.22} & 3.22 & \textbf{7.72} & 2.31 & \textbf{8.00} & 2.08 \\
        \bottomrule
    \end{tabular}
    }
\end{table}

\subsection{Video Generation with Multiple Control Conditions}
\label{sec:results_video_gen_multi}

As described in \cref{subsec:method_multi_source},
users can achieve multi-source control by simply
combining the control features of multiple \cnet{}s via our \adaptermethod{}.
\added{
\Cref{tab:multi-source-video-gen} shows the result in two folds: firstly, patch-level MoE routers (\ie{}, variants b and c in \cref{fig:multi_source_method}) consistently outperforms the equal weights baseline as well as the unconditional global weights (\ie{}, variant a in \cref{fig:multi_source_method}), which proves the effectiveness of patch-level fine-grained control composition. Secondly, as shown in (b) and (c),  control with more conditions almost always yields better spatial control and visual quality
than control with a single condition.
\cref{fig:multi_source_examples} and \cref{fig:multi_source_examples_2} show that multi-condition composition provides more accurate control compared to a single condition.
\Cref{tab:multi-source-video-gen-all-results} extends (a) by conditioning on image/text/timestep embeddings. 
}

\begin{figure}[h]
  \centering
  \includegraphics[width=.89\linewidth]{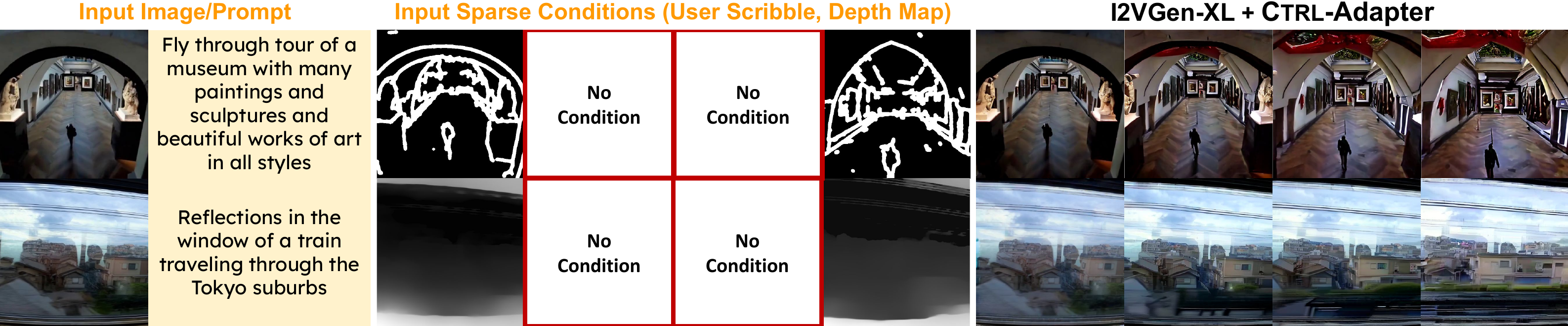}
  \vspace{-2mm}
  \caption{Video generation from sparse frame conditions with \adaptermethod{} on I2VGen-XL (which generates 16 frames in total). We only provide controls for the 1st, 6th, 11th, and 16th frames.
  }
  \vspace{-4mm}
\label{fig:sparse_control_examples} 
\end{figure}

\subsection{Video Generation with Sparse Frames as Control Condition}
\label{sec:results_sparse_control}

We experiment \adaptermethod{} with providing sparse frame conditions 
using I2VGen-XL as backbone.
During each training step, we first randomly select an integer $k\in\{1, ..., N\}$, where $N$ is equal to the total number of output frames (\eg{}, $N=16$ for I2VGen-XL). Next, we randomly select $k$ key frames from $N$ total frames.
We then extract these key frames' depth maps and user scribbles as control conditions.
we do not give the latents $\bm{z}$ and only give the $k$ frames to \cnet{}.
In \cref{fig:sparse_control_examples}, we can see that I2VGen-XL with our \adaptermethod{} can correctly generate videos that follow the control conditions for the given 4 sparse key frames and make reasonable interpolations on the frames without conditions.
In \cref{sec:ablation_skipping_latent},
we show that skipping the latent from \cnet{} inputs is important in improving the sparse control capability.

\subsection{Zero-Shot Generalization on Unseen Conditions}
\label{sec:results_zero_shot_generalization}

\begin{figure}[h]
  \centering
  \includegraphics[width=.85\linewidth]{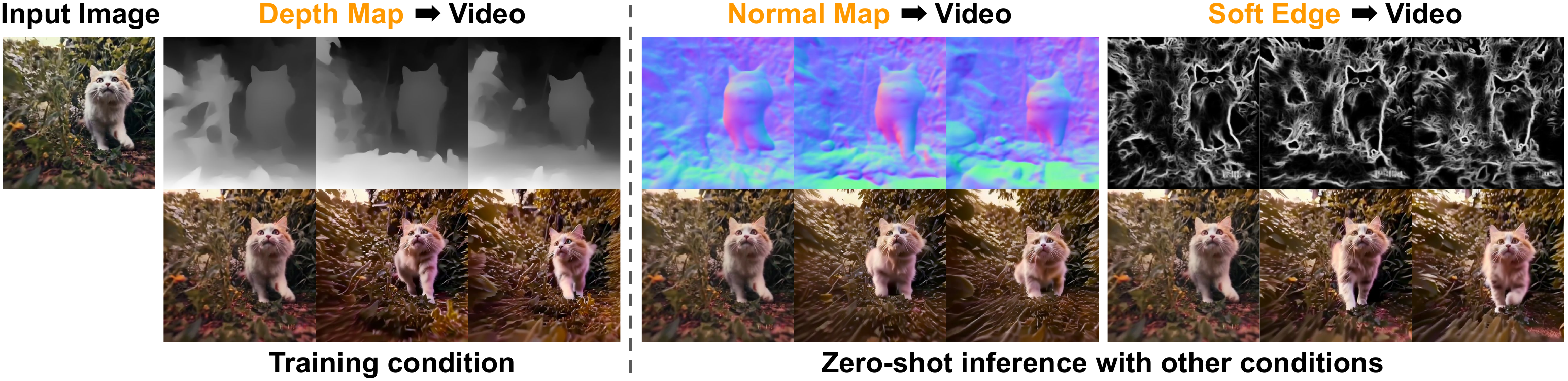}
  \vspace{-2mm}
  \caption{
  Zero-shot transfer of \adaptermethod{} trained only on depth maps to unseen conditions.
  }
\label{fig:zero-shot-inference-unseen-conditions} 
\end{figure}

\cnet{} can be understood as an image feature extractor that maps different types of controls to the unified representation space of backbone generation models.
This begs an interesting question:
\textit{``Does \adaptermethod{} learn general feature mapping from one (smaller) backbone to another (larger) backbone?''}
To answer this question,
we experiment by directly plugging \adaptermethod{} to \cnet{}s
that are \textbf{not seen during training}.
In \cref{fig:zero-shot-inference-unseen-conditions},
we observe the \adaptermethod{} trained on depth maps can adapt to normal map and soft edge \cnet{}s 
in a zero-shot manner. Quantitative analysis of different training strategies based on such observation is illustrated in \cref{sec:ablation_invidual_or_unified_adapter}.

\section{Downstream Tasks Beyond Spatial Control}
\label{sec:beyond_spatial_control}

\begin{figure}[h]
  \centering
  \includegraphics[width=.9\linewidth]{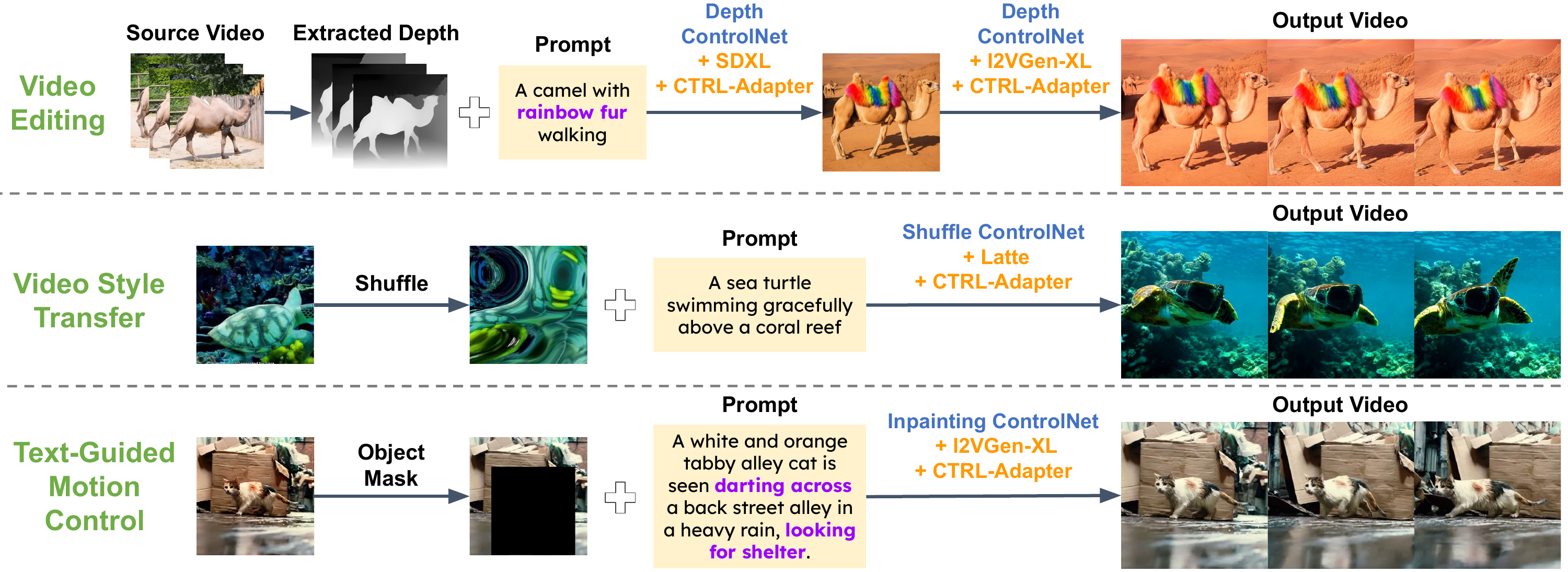}
  \caption{
  Illustration of additional downstream tasks supported by our \adaptermethod{} framework.
  }
\label{fig:down_stream_tasks_main} 
\end{figure}

\added{Here, we aim to {\sl{qualitatively}} explore how other types of \cnet{}s can be seamlessly integrated into our framework to enable a wide variety of downstream tasks {beyond spatial control}.
As shown in \cref{fig:down_stream_tasks_main}, we can achieve video editing by combining image and video \adaptermethod{}s with user edited prompts; video style transfer via shuffle \cnet{} $+$ \adaptermethod{}; and text-guided motion control for masked object via inpainting \cnet{} $+$ \adaptermethod{}. See \cref{appendix_subsec:down_stream_task_examples} for details.
}

\section{Conclusion}
\label{sec:conclusion}

We propose \adaptermethod{},
an efficient, powerful, and versatile framework that adds diverse controls to any image/video diffusion model.
Training an \adaptermethod{} is significantly more efficient than training a \cnet{} for a new backbone, and it can outperform or match strong baselines in visual quality and spatial control.
\adaptermethod{} not only provides many useful capabilities including image/video control, sparse frame control, multi-condition control, and zero-shot adaption to unseen conditions, but also can be easily and flexibly integrated into a variety of downstream tasks.

\section*{Acknowledgments}
This work was supported by DARPA ECOLE Program No. HR00112390060, NSF-AI Engage Institute DRL-2112635, DARPA Machine Commonsense (MCS) Grant N66001-19-2-4031, ARO Award W911NF2110220, ONR Grant N00014-23-1-2356, and a Bloomberg Data Science Ph.D. Fellowship. The views contained in this article are those of the authors and not of the funding agency.

{
    \small
    \bibliographystyle{abbrv}
    \bibliography{main}
}

\renewcommand\thesubsection{\Alph{subsection}}

\input{99_appendix}

\end{document}

%% file: 00_preamble.tex
\usepackage[dvipsnames]{xcolor}

\newcommand{\added}[1]{\textcolor{black}{{#1}}}

\usepackage{algorithm}

\usepackage[frozencache=true,cachedir=minted-cache]{minted}

\usepackage{amsmath,amsfonts,bm}
\usepackage{graphicx}
\usepackage{multirow}
\usepackage{enumitem}
\usepackage{makecell}

\definecolor{cvprblue}{rgb}{0.21,0.49,0.74}
\usepackage[pagebackref,breaklinks,colorlinks,citecolor=cvprblue]{hyperref}

\definecolor{lightblue}{rgb}{0.90, 0.95, 0.98}
\newcolumntype{a}{>{\columncolor{lightblue}}c}

\newcommand{\adaptermethod}{{\textsc{Ctrl}-Adapter}}
\newcommand{\cnet}{{ControlNet}}

\newcommand{\davis}{{DAVIS}}
\newcommand{\sdxl}{{SDXL}}

\newcommand{\svd}{{SVD}}

\newcommand{\eg}[1]{\emph{e.g}.}
\newcommand{\ie}[1]{\emph{i.e}.}
\newcommand{\etc}[1]{\emph{etc}.}
\newcommand{\etal}[1]{\emph{et al}.}

\usepackage[page,header]{appendix}
\usepackage{titletoc}

\definecolor{Red}{rgb}{0.6,0,0}
\definecolor{Blue}{rgb}{0,0,0.8}
\definecolor{Green}{rgb}{0.2,0.8,0}
\definecolor{airforceblue}{rgb}{0.36, 0.54, 0.66}
\definecolor{ao(english)}{rgb}{0.0, 0.5, 0.0}
\definecolor{azure(colorwheel)}{rgb}{0.0, 0.5, 1.0}
\definecolor{crimson}{rgb}{0.86, 0.08, 0.24}
\definecolor{darkcerulean}{rgb}{0.03, 0.27, 0.49}
\definecolor{cobalt}{rgb}{0.0, 0.28, 0.67}
\definecolor{rosegold}{rgb}{0.72, 0.43, 0.47}
\definecolor{orange-red}{rgb}{1.0, 0.27, 0.0}
\definecolor{mountainmeadow}{rgb}{0.19, 0.73, 0.56}
\definecolor{malachite}{rgb}{0.04, 0.85, 0.32}
\definecolor{darkblue}{rgb}{0.0, 0.0, 0.55}
\definecolor{customblue}{rgb}{0.2, 0.35, 0.8}

\hypersetup{colorlinks,linkcolor={blue},citecolor={mountainmeadow},urlcolor={magenta}}

\usepackage[capitalize]{cleveref}
\crefname{section}{Sec.}{Secs.}
\Crefname{section}{Section}{Sections}
\Crefname{table}{Table}{Tables}
\crefname{table}{Tab.}{Tabs.}

\usepackage{colortbl}

\usepackage{pifont}
\newcommand*\colourcheck[1]{%
  \expandafter\newcommand\csname #1check\endcsname{\textcolor{#1}{\ding{52}}}%
}
\colourcheck{blue}
\colourcheck{green}
\colourcheck{red}
\colourcheck{olive}
\colourcheck{ForestGreen}

\newcommand*\colourcross[1]{%
  \expandafter\newcommand\csname #1cross\endcsname{\textcolor{#1}{\ding{56}}}%
}
\colourcross{blue}
\colourcross{green}
\colourcross{red}

%% file: 99_appendix.tex
\newpage

{
\begin{center}
    \textbf{{{\Large Appendix}}}
\end{center}
}

\renewcommand\thesection{\Alph{section}}
\renewcommand\thesubsection{\Alph{section}.\arabic{subsection}}

\appendix
\startcontents[sections]
\printcontents[sections]{l}{1}{\setcounter{tocdepth}{3}}

\vspace{20pt}

\section{Background}

\subsection{Extended Related Works}
\label{sec:appendix_related_works}

\paragraph{Text-to-video and image-to-video generation models.}
Generating videos from text descriptions or images (\eg{}, initial video frames) based on deep learning
and has increasingly gained much attention.
Early works for this task~\cite{li2017video,li2018storygan,Zhao_2018_ECCV,stylegan_v}
have commonly used
variational autoencoders (VAEs)~\cite{kingma2013auto} and generative adversarial networks (GANs)~\cite{goodfellow2020generative},
while
most of recent video generation works are based on
denoising diffusion models~\cite{ho2020denoising,Sohl-Dickstein2015}.
Powered by large-scale training, recent video diffusion models demonstrate impressive performance in generating highly realistic videos from text descriptions~\cite{he2022lvdm,ho2022imagen-video,singer2022make,zhou2022magicvideo,khachatryan2023text2video,wang2023gen,yin2023nuwa-xl,wang2023modelscope,Mullan_Hotshot-XL_2023,sora_2024, gupta2023photorealistic, menapace2024snap}
or initial video frames (\ie{}, images)~\cite{blattmann2023stable,zhang2023i2vgen, guo2023animatediff, xing2023dynamicrafter}.

\paragraph{Adding control to image/video diffusion models.}

While recent image/video diffusion models demonstrate impressive performance in generating highly realistic images/videos from text descriptions,
it is hard to describe every detail of images/videos
only with text or first frame image.
Instead, there have been many works using different types of additional inputs to control the image/video diffusion models,
such as bounding boxes~\cite{li2023gligen,yang2022reco},
reference object image~\cite{Ruiz2023Dreambooth,Gal2023TextualInversion,Li2023BLIP-Diffusion},
segmentation map~\cite{Gafni2022Make-A-Scene,Avrahami2023SpaText,zhang2023adding},
sketch~\cite{zhang2023adding},
\etc{},
and combinations of multiple conditions~\cite{kim2023diffblender,qin2023unicontrol,zhao2024uni,wang2024videocomposer}.
As finetuning all the parameters of such image/video diffusion models is computationally expensive,
several methods, such as \cnet{}~\cite{zhang2023adding}, have been proposed to add conditional control capability via parameter-efficient training~\cite{zhang2023adding,simoryu_lora_diffusion,mou2023t2i}.
X-Adapter~\cite{ran2023x} learns an adapter module to reuse \cnet{}s pretrained with a smaller image diffusion model (\eg{}, SDv1.5) for a bigger image diffusion model (\eg{}, SDXL).
While they focus solely on learning an adapter for image control, \adaptermethod{} features architectural designs (\eg{}, temporal convolution/attention layers)
for video generation as well.
In addition, X-Adapter needs the smaller image diffusion model (SDv1.5) during training and inference, whereas \adaptermethod{} doesn't need the smaller diffusion model at all (for image/video generation), hence being more memory and computationally efficient (see \cref{sec:visualized_comparison_different_methods} for details).
SparseCtrl~\cite{guo2023sparsectrl} guides a video diffusion model with conditional inputs of few frames (instead of full frames), to alleviate the cost of collecting video conditions.
Since SparseCtrl involves augmenting \cnet{} with an additional channel for frame masks, it requires training a new variant of \cnet{} from scratch. 
In contrast, we leverage existing image \cnet{}s more efficiently by propagating information through temporal layers in adapters and enabling sparse frame control via skipping the latents from \cnet{} inputs (see \cref{subsec:method_adapter} for details).
Furthermore, compared with previous works that are specially designed for specific condition controls on a single modality (image~\cite{zhang2023adding, qin2023unicontrol} or video~\cite{hu2023videocontrolnet, zhang2023controlvideo}),
our work presents a unified and versatile framework that supports diverse controls, including image control, video control, sparse frame control, and multi-source control,
with significantly lower computational costs by reusing pretrained \cnet{}s (\eg{}, \adaptermethod{} outperforms baselines in less than 10 GPU hours).
\Cref{table:comparison_of_model_abilities} summarizes the 
comparison of \adaptermethod{} with related works.

\subsection{Extended Preliminaries: LDM and \cnet{}}
\label{sec:appendix_preliminaries}

\paragraph{Latent Diffusion Models.}
Many recent video generation works are based on
latent diffusion models (LDMs)~\cite{rombach2022high},
where a diffusion model learns the temporal dynamics of compact latent representations of videos.
First, 
given a $F$-frame RGB video $\bm{x}\in\mathbb{R}^{F\times 3 \times H \times W}$,
a video encoder (of a pretrained autoencoder) provides $C$-dimensional latent representation (\ie{}, latents):
$\bm{z}=\mathcal{E}(\bm{x})\in\mathbb{R}^{F\times C \times H' \times W'}$,
where height and width are spatially downsampled
($H' < H$ and $W' < W$).
Next,
in the forward process,
a noise scheduler such as DDPM~\cite{ho2020denoising}
gradually adds noise to the latents $\bm{z}$:
$q(\bm{z}_t|\bm{z}_{t-1})=N(\bm{z}_t; \sqrt{1-\beta_t}\bm{z}_{t-1}, \beta_t \bm{I})$,
where $\beta_t\in(0,1)$ is the variance schedule with $t\in\{1, ..., T\}$.
Then, in the backward pass, a diffusion model (usually a U-Net architecture) $\bm{\mathcal{F}_\theta}(\bm{z}_t, t, \bm{c}_\text{text/img})$ learns to gradually denoise the latents,
given 
a diffusion timestep $t$, and
a text prompt $\bm{c}_\text{text}$ (\ie{}, T2V)
or an initial frame $\bm{c}_\text{img}$ (\ie{}, I2V) if provided.
The diffusion model is trained with following objective: 
$\mathcal{L}_{\text{LDM}}=\mathbb{E}_{\bm{z}, \bm{\epsilon}\sim N(0, \bm{I}), t}\|\bm{\epsilon} - \bm{\epsilon_{\theta}}(\bm{z}_t, t, \bm{c}_\text{text/img})\|_2^2
$, where $\bm{\epsilon}$ and $\bm{\epsilon_\theta}$ represent the added noise to latents and the predicted noise by $\bm{\mathcal{F}_{\theta}}$ respectively. We apply the same objective for \adaptermethod{} training.

\paragraph{\cnet{}s.} 
\cnet{}~\cite{zhang2023adding} is designed to add spatial controls to image diffusion models in the form of different guidance images (\eg{}, depth, sketch, segmentation maps, \etc{}).
Specifically,
given a pretrained backbone image diffusion model
$\bm{\mathcal{F}_{\theta}}$
that consists of input/middle/output blocks,
\cnet{} has a similar architecture $\bm{\mathcal{F}_{\theta'}}$, where
the input/middle blocks parameters of $\bm{{\theta'}}$ are initialized from $\bm{{\theta}}$,
and the output blocks consist of 1x1 convolution layers initialized with zeros.
\cnet{} takes the
diffusion timestep $t$,
text prompt $\bm{c}_\text{text}$,
control image $\bm{c}_\text{f}$ (\eg{}, depth image), and the noisy latents $\bm{z_t}$ as inputs,
and provides the features that are merged into middle/output blocks of backbone image model $\bm{\mathcal{F}_\theta}$ to generate the final image.
The authors of \cnet{} have released a variety of \cnet{} checkpoints based on Stable Diffusion~\cite{rombach2022high} v1.5 (SDv1.5) and the user community have also shared many \cnet{}s trained with different input conditions based on SDv1.5.
However, these \cnet{}s cannot be used with more recently released bigger and stronger image/video diffusion models, such as SDXL~\cite{podell2023sdxl} and I2VGen-XL~\cite{zhang2023i2vgen}.
Moreover, the input/middle blocks of the \cnet{} are in the same size with those of the diffusion backbones (\ie{}, if the backbone model gets bigger, \cnet{} also gets bigger).
Due to this, it becomes increasingly difficult to train new \cnet{}s for each bigger and newer model that is released over time.
To address this, we introduce \adaptermethod{} for efficient adaption of existing \cnet{}s for new diffusion models.

\section{\adaptermethod{} Method and Architecture Details}
\label{sec:appendix_method}

\subsection{\adaptermethod{} Architecture Details}
\label{sec:adapter_architecture_detail}

\begin{figure}[t]
  \centering
  \includegraphics[width=.9\linewidth]{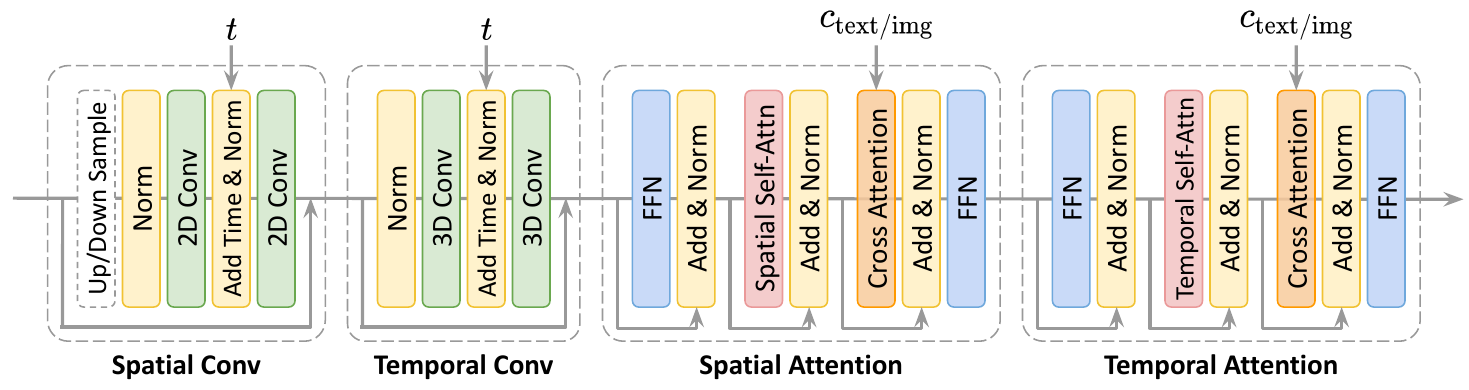}
  \caption{
  Detailed architecture of \adaptermethod{} blocks.
  }
\label{fig:method_video_architecture} 
\end{figure}

In \cref{fig:method_video_architecture},
we illustrate the detailed architecture of \adaptermethod{} blocks.
See \cref{fig:adapter_method} for how the \adaptermethod{} blocks are used to adapt \cnet{}s to image/video diffusion models. \cref{fig:method_video_architecture} is an extended version of
\cref{fig:adapter_method} (right) with more detailed visualizations, including skip connections, normalization layers in each module, and the linear projection layers (\ie{}, FFN) in each spatial/temporal attention modules.

\subsection{PyTorch Implementation for Inverse Timestep Sampling}
\label{sec:pseudocode_inverse_timestep_sampling}

In \cref{alg:timestep_sampling}, we provide the PyTorch~\cite{Ansel_PyTorch_2_Faster_2024} implementation of inverse timestep sampling, described in \cref{subsec:method_adapter}.
In the example, inverse time stamping adapts to the SVD~\cite{blattmann2023stable} backbone.

During each training step, the procedure for this algorithm can be summarized as follows:
\begin{itemize}[leftmargin=1.5em]
    \item Sample a variable $u$ from \text{Uniform}$[0, 1]$. See line 19 in function \texttt{inverse\_timestamp\_sample}.
    \item Sample noise scale $\sigma_\text{cont.}$ via inverse transform sampling~\cite{Inverse_transform_sampling}; \ie{}, we derive the inverse cumulative density function of $\sigma_\text{cont.}$ and sample $\sigma_\text{cont.}$ by sampling $u$: $\sigma_\text{cont.}=F_\text{cont.}^{-1}(u)$. See function \texttt{sample\_sigma} and line 21 in function \texttt{inverse\_timestamp\_sample}.
    \item Given a preconditioning function $g_\text{cont.}$  that maps noise scale to timestep (typically associated with the continuous-time noise sampler),
we can compute $t_\text{cont.}=g_\text{cont.}(\sigma_\text{cont.})$. See function \texttt{sigma\_to\_timestep} and line 23 in \texttt{inverse\_timestamp\_sample}.
\item Set the timesteps and noise scales for both \cnet{} and our \adaptermethod{} as $t_\text{CNet}=\text{round}(1000u)$ and $\sigma_\text{CNet}=u$ respectively, where 1000 represents the denoising timestep range over which the \cnet{} is trained. See line 25 in \texttt{inverse\_timestamp\_sample}.
\end{itemize}

During inference, we follow the similar sampling strategy, with the only change in the first step. Instead of uniformly sample a single value for $u$, we uniformly sample $k$ equidistant values for $u$ within $[0, 1]$  and derive corresponding $t_\text{cont./CNet}$ and $\sigma_\text{cont./CNet}$ as inputs for denoising steps, where $k$ here is the number of denoising steps during inference.

\input{timestep_pseudocode}

\subsection{Comparison of \adaptermethod{} Variants and Related Methods}
\label{sec:visualized_comparison_different_methods}

\begin{figure}[h]
  \centering
  \includegraphics[width=.75\linewidth]{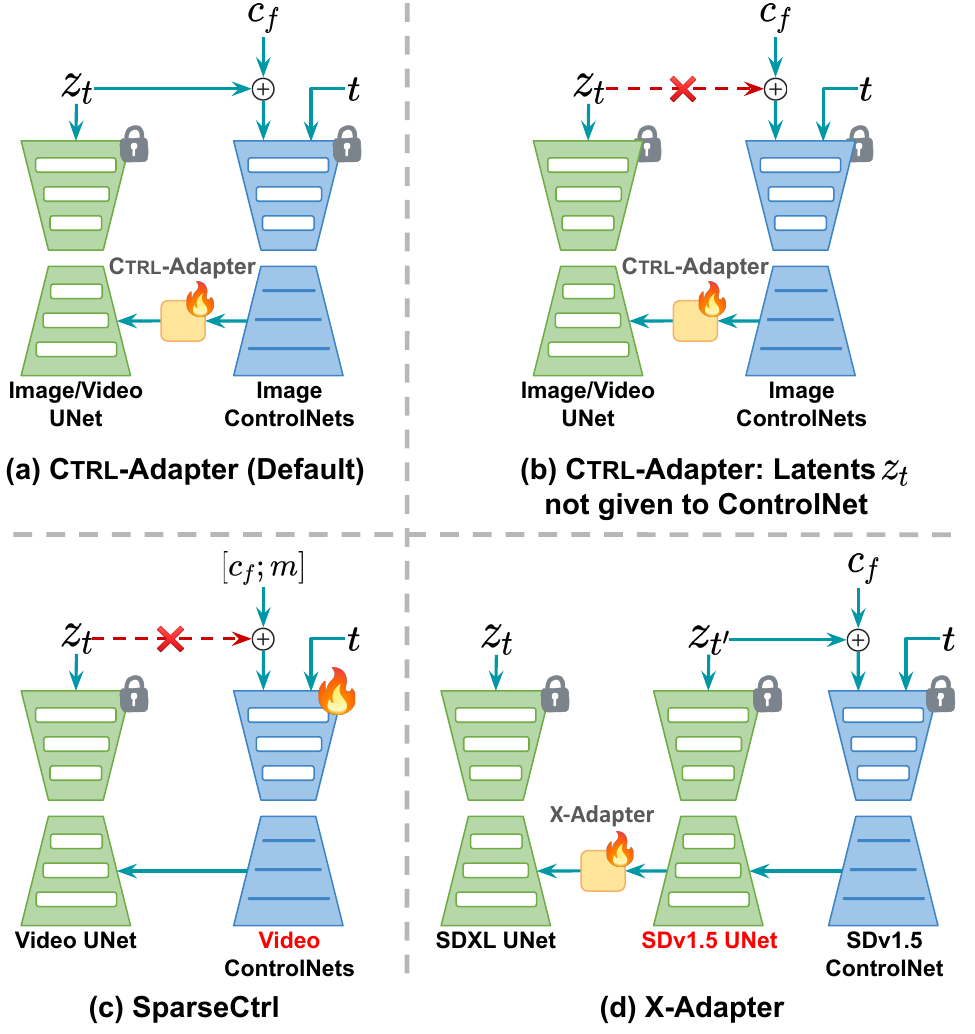}
  \caption{
  Comparison of giving different inputs to \cnet{}, where $\bm{z}_t, \bm{c}_\text{f}$, and $t$ represent latents, input control features, and timesteps respectively.
  \textbf{(a):} Default \adaptermethod{} design.
  \textbf{(b):} Variant of \adaptermethod{} where latents $\bm{z}_t$ are not given to \cnet{} (see \cref{subsec:method_adapter} for details).
  \textbf{(c):} SparseCtrl~\cite{guo2023sparsectrl} trains a video \cnet{} with control conditions $c_f$ and frame masks $m$ as inputs. 
  \textbf{(d):} X-Adapter~\cite{ran2023x} needs SDv1.5 U-Net as well as SDv1.5 \cnet{} during training and inference, whereas \adaptermethod{} doesn't need to SDv1.5 U-Net at all.
  }
\label{fig:appendix_comparison_of_all_model_architectures} 
\end{figure}

In \cref{fig:appendix_comparison_of_all_model_architectures},
we compare the variants of \adaptermethod{} designs (with latent is given / not given to \cnet{}; see \cref{subsec:method_adapter} for details) and two related methods: SparseCtrl~\cite{guo2023sparsectrl} and X-Adapter~\cite{ran2023x}.
Unlike \adaptermethod{} that leverages the pretrained image \cnet{}s,
SparseCtrl (\cref{fig:appendix_comparison_of_all_model_architectures} c) trains a video \cnet{} with control conditions $c_f$ and frame masks $m$ as inputs. 
While X-Adapter (\cref{fig:appendix_comparison_of_all_model_architectures} d) needs SDv1.5 U-Net as well as SDv1.5 \cnet{} during training and inference,
\adaptermethod{} doesn't need to SDv1.5 U-Net at all.

\section{Training and Inference Details}
\label{sec:appendix_train_inference_details}


\paragraph{Model architectures.} Detailed illustration of our \adaptermethod{} architecture has been provided across several parts of our paper, including \cref{sec:method}, \cref{sec:adapter_architecture_detail}, \cref{sec:appendix_variants_of_adapter_architecture_design}, and \cref{appendix_sec:additional_quant_analysis}. In addition, for all the backbone models used in this paper, we kept all their parameters frozen and made no modifications.

\paragraph{Training details.} We use a learning rate of $5\times e^{-5}$; AdamW~\cite{loshchilov2018decoupled} optimizer with values for $\beta_1$, $\beta_2$, $\epsilon$, and weight decay as 0.9, 0.999, $1\times e^{-8}$, and $1\times e^{-2}$ respectively. We set the max gradient norm as 1. 
All our experiments are trained on 4 A100 80GB GPUs with batch size of 1. Detailed study of
per-GPU training memory for different model architecture variants are shown in \cref{fig:ablation_bubble_plot_adapter_architecture_sdxl_i2vgenxl}. {\sl{Please note that other than mixed-precision training with data type \texttt{bfloat16}, we didn't use any additional methods to speed up the training/inference clock time, or to save GPU memory.}} To be more specific, \textbf{we didn't use any of the following methods}: \texttt{xformers}~\cite{xFormers2022}, gradient checkpointing, 8bit Adam optimizer, and DeepSpeed~\cite{Rasley2020DeepSpeedSO}. In addition, to make our framework easy to use directly from raw input images/videos, we extract all control condition images/frames \textbf{on-the-fly} during training. We train the image and video \adaptermethod{}s for 80k and 40k steps respectively, which can be finished in 24 hours measured by training clock time. The fast convergence of our method is shown in \cref{fig:training_statistics_comparison}.

\paragraph{Inference details.} 
All inference can be done on a single A6000 GPU with 48GB memory. During inference, we use the default hyper-parameters for each backbone model, including the number of frames to generate, the number of denoising steps, and classifier-free guidance scale, \etc{}.

\paragraph{Safeguards.}
When we generate images during inference, we also activate the NSFW filter of the backbone models. This ensures that users are protected from unnecessary exposure to explicit or objectionable materials. For training, the datasets we used~\cite{chen2024panda70m, LAION_POP} both filter out the image/video samples with harmful contents. For example, as stated in the ``Risk mitigation'' section of Panda70M paper, they used the internal automatic pipeline to filter out the video
samples with harmful or violent language and texts that
include drugs or hateful speech. They also use the NLTK
framework to replace all people’s names with "person". LAION-POP dataset is also created by filtering out samples based on the safety tags (using a customized trained NSFW classifier that they built).

\section{Experimental Setup}
\label{sec:appendix_experiment_setups}

\subsection{\cnet{}s and Target Diffusion Models}
\label{source_target_diffusions}

\paragraph{\cnet{}s.}
We use \cnet{}s trained with SDv1.5.\footnote{\url{https://huggingface.co/lllyasviel/ControlNet}}
SDv1.5 has the most number of publicly released \cnet{}s
and has a much smaller training cost
compared to recent image/video diffusion models.
Note that unlike X-Adapter~\cite{ran2023x}, \adaptermethod{} does not need to load the source diffusion model (SDv1.5) during training or inference (see (a) and (d) in \cref{fig:appendix_comparison_of_all_model_architectures} for model architecture comparison).

\paragraph{Target diffusion models (where \cnet{}s are to be adapted).}
For video generation models, we experiment with two text-to-video generation models -- Latte~\cite{ma2024latte} and Hotshot-XL~\cite{Mullan_Hotshot-XL_2023},
and two image-to-video generation models -- I2VGen-XL~\cite{zhang2023i2vgen} and Stable Video Diffusion (SVD)~\cite{blattmann2023stable}. 
For image generation model, we experiment with PixArt-$\alpha$~\cite{chen2024pixartalpha} and the base model in SDXL~\cite{podell2023sdxl}.
For all models, we use their default settings during training and inference (\eg{}, number of output frames, resolution, number of denoising steps, classifier-free guidance scale, \etc{}).

\subsection{Training Datasets for \adaptermethod{}}
\label{sec:training_datasets}

\paragraph{Video datasets.}
For training \adaptermethod{} for video diffusion models, we download around 1.5M videos randomly sampled from the Panda-70M training set~\cite{chen2024panda70m}.
Following recent works~\cite{blattmann2023stable, dai2023emu}, we filter out videos of static scenes by removing videos whose average optical flow~\cite{farneback2003two, opencv_library} magnitude is below a certain threshold.
Concretely,
we 
use the Gunnar Farneback's algorithm\footnote{ \url{https://docs.opencv.org/4.x/d4/dee/tutorial_optical_flow.html}}~\cite{Farnebck2003TwoFrameME}
at 2FPS, calculate the averaged the optical flow for each video and re-scale it between 0 and 1, and filter out videos whose average optical flow error is below a threshold of 0.25.
This process gives us a total of 200K remaining videos.

\paragraph{Image datasets.}
For training \adaptermethod{} for image diffusion models,
we use 300K images randomly sampled from LAION POP,\footnote{\url{https://laion.ai/blog/laion-pop/}} which is a subset of LAION 5B~\cite{schuhmann2022laion} dataset and contains 600K images in total with aesthetic values of at least 0.5 and a minimum resolution of 768 pixels on the shortest side.
As suggested by the authors, we use the image captions generated with  CogVLM~\cite{wang2023cogvlm}.

\subsection{Input Conditions}
\label{sec:input_conditions}

We extract various input conditions
from the video and image datasets described above.

\begin{itemize}[leftmargin=1.5em]
    \item \textbf{Depth map:}
    As recommended in Midas\footnote{\url{https://github.com/isl-org/MiDaS}}~\cite{ranftl2020towards},
    we employ \texttt{dpt\_swin2\_large\_384} for the best speed-performance trade-off. 
    \item \textbf{Canny edge, surface normal, line art, softedge, and user sketching/scribbles:}
    Following \cnet{}~\cite{zhang2023adding}, we utilize the same annotator implemented in the \texttt{controlnet\_aux}\footnote{\url{https://github.com/huggingface/controlnet_aux}} library.
    \item \textbf{Semantic segmentation map:} To obtain higher-quality segmentation maps
    than 
    UPerNet~\cite{xiao2018unified} used in \cnet{},
    we employ SegFormer~\cite{xie2021segformer} \texttt{segformer-b5-finetuned-ade-640-640} finetuned on ADE20k dataset at 640$\times$640 resolution.
    \item \textbf{Human pose:} We employ ViTPose~\cite{xie2021segformer} \texttt{ViTPose\_huge\_simple\_coco} to improve both processing speed and estimation quality, compared to OpenPose ~\cite{cao2017realtime} used in \cnet{}.
\end{itemize}

\subsection{Evaluation Datasets}
\label{sec:evaluation_datasets}

\paragraph{Video datasets.}
Following previous works~\cite{hu2023videocontrolnet,zhang2023controlvideo},
we evaluate our video ControlNet adapters on \davis{} 2017~\cite{pont20172017}, a public benchmark dataset
also used in other controllable video generation works~\cite{hu2023videocontrolnet}.
We first combine all video sequences from  \texttt{TrainVal}, \texttt{Test-Dev 2017} and \texttt{Test-Challenge 2017}.
Then, we chunk each video into smaller clips, with the number of frames in each clip being the same as the default number of frames generated by each video backbone (\eg{}, 8 frames for Hotshot-XL, 16 frames for I2VGen-XL, and 14 frames for SVD). This process results in a total of 1281 video clips of 8 frames, 697 clips of 14 frames, and 608 video clips of 16 frames.
\paragraph{Image datasets.} We evaluate our image ControlNet adapters on COCO \texttt{val2017} split~\cite{lin2014microsoft}, which contains 5k images that cover diverse range of daily objects. We resize and center crop the images to 1024 by 1024 for SDXL evaluation.

\subsection{Evaluation Metrics}
\label{sec:evaluation_metrics}

\paragraph{Visual quality.}
Following previous works~\cite{qin2023unicontrol, hu2023videocontrolnet},
we use Frechet Inception Distance (FID)~\cite{heusel2017gans} to measure the distribution distance between our generated images/videos and input images/videos.\footnote{To be consistent with the numbers reported in Uni-ControlNet~\cite{zhao2024uni}, we use pytorch-fid (\url{https://github.com/mseitzer/pytorch-fid}) in \Cref{table:single_source_image_results}. For other results, we use clean-fid~\cite{parmar2021cleanfid} (\url{https://github.com/GaParmar/clean-fid}) which is more robust to aliasing artifacts.}

\paragraph{Spatial control.}
For video datasets,
following VideoControlNet~\cite{hu2023videocontrolnet}, we report the L2 distance between the optical flow~\cite{ranjan2017optical} of the input video and the generated video (Optical Flow Error).
For image datasets,
following Uni-ControlNet~\cite{zhao2024uni}, we report the Structural Similarity (SSIM)~\cite{wang2004image}\footnote{\url{https://scikit-image.org/docs/stable/auto_examples/transform/plot_ssim.html}} and mean squared error (MSE)\footnote{\url{https://scikit-learn.org/stable/modules/classes.html\#module-sklearn.metrics}} between generated images and ground truth images.

\section{Variants of \adaptermethod{} Architecture Design}
\label{sec:appendix_variants_of_adapter_architecture_design}

\subsection{\adaptermethod{} Design Ablations}
\label{sec:additional_adapter_design_ablations}

\subsubsection{Combinations of components within each \adaptermethod{}}

\begin{figure}[t]

\minipage{0.49\textwidth}
  \includegraphics[width=\linewidth]{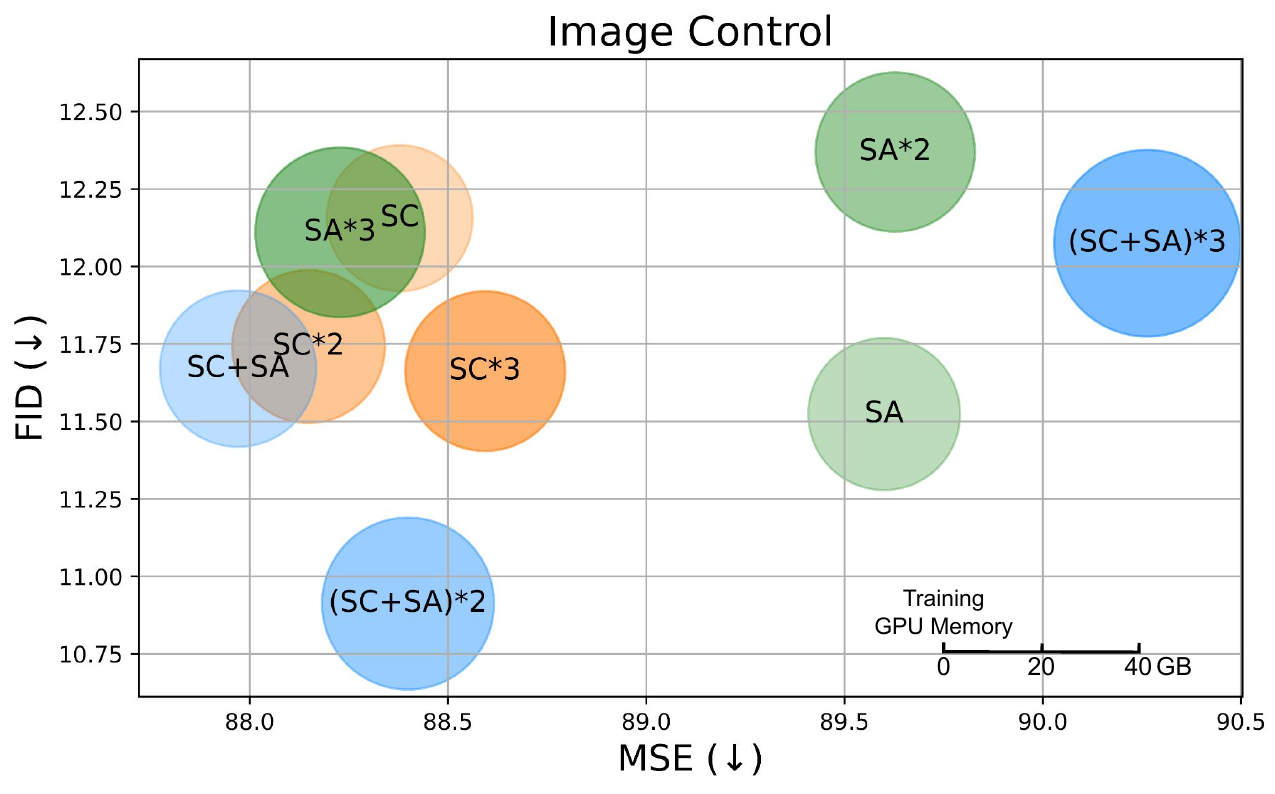}
\endminipage\hfill
\minipage{0.49\textwidth}
  \includegraphics[width=\linewidth]{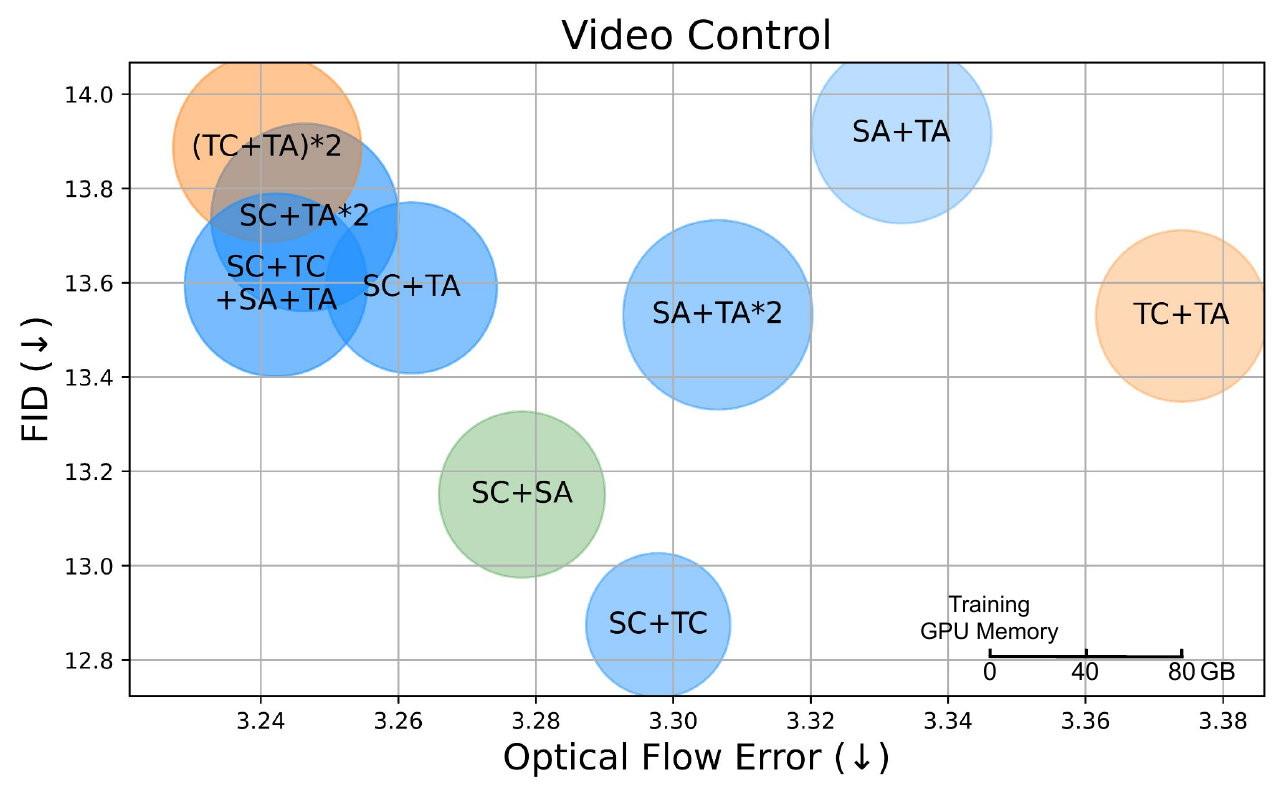}
\endminipage\hfill
  \caption{
  Comparison of different architecture of \adaptermethod{} for image and video control,
  measured with visual quality (FID) and spatial control (MSE/optical flow error) metrics. The metrics are calculated from 1000 randomly selected COCO val2017 images and 150 videos from DAVIS 2017 dataset respectively.
  \textbf{Left:} image control on SDXL backbone.
  \textbf{Right:} video control on I2VGen-XL backbone.
  For both plots, data points in the \textbf{bottom-left} are ideal.
  SC, TC, SA, and TA: Spatial Convolution, Temporal Convolution, Spatial Attention, and Temporal Attention.
  $^*N$ represents the number of blocks in each \adaptermethod{}.
  The diameters of bubbles represent training GPU memory.
  }
\label{fig:ablation_bubble_plot_adapter_architecture_sdxl_i2vgenxl}
\end{figure}

\begin{figure}[t]
\minipage{0.45\textwidth}
  \includegraphics[width=\linewidth]{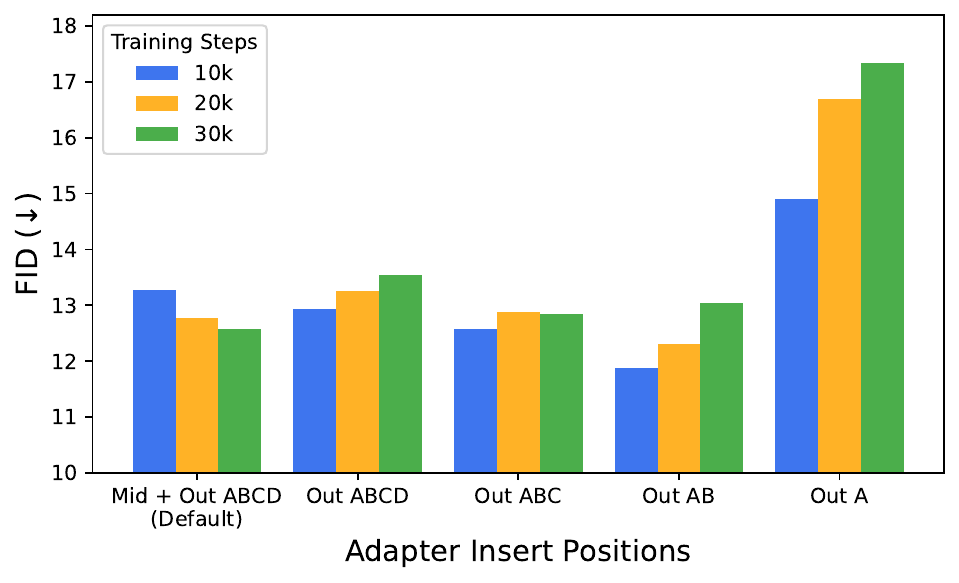}
\endminipage\hfill
\minipage{0.45\textwidth}
  \includegraphics[width=\linewidth]{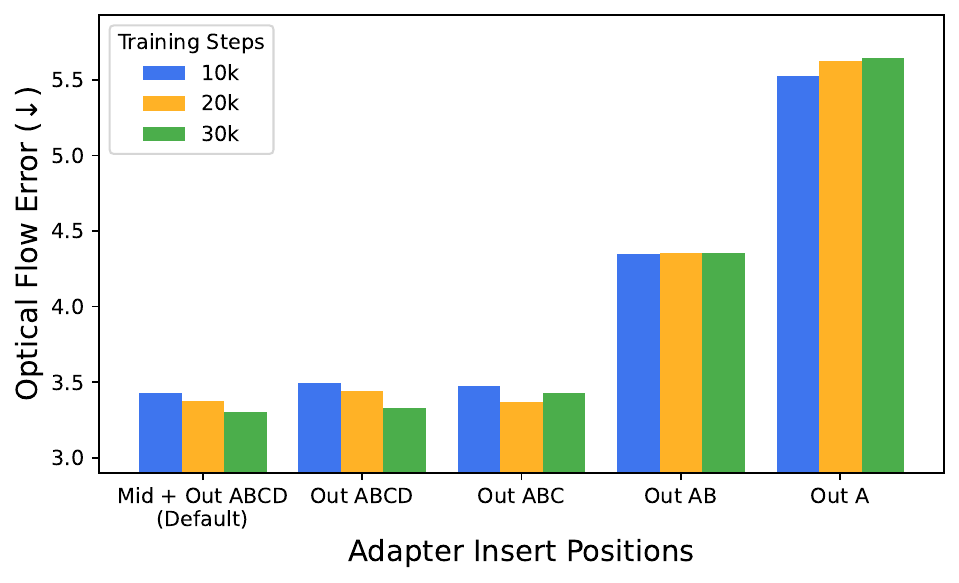}
\endminipage\hfill
  \caption{Comparison of inserting \adaptermethod{} to different U-Net blocks. `Mid' represents the middle block, whereas `Out ABCD' represents output blocks A, B, C, and D. The metrics are calculated from 150 videos from \davis{} 2017 dataset.
  }
  \label{fig:appendix_i2vgenxl_abcdm}
\end{figure}

\begin{figure}[t]
  \centering
  \includegraphics[width=.99\linewidth]{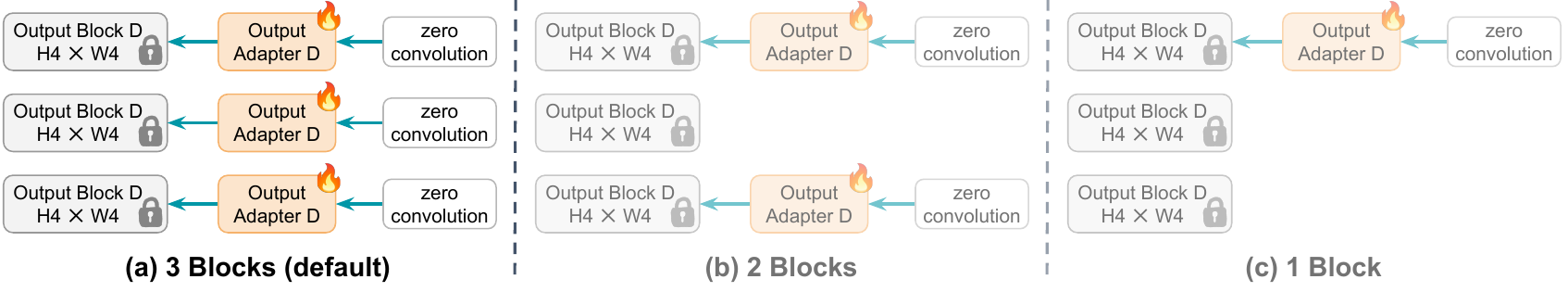}
  \caption{Comparison of inserting different numbers of \adaptermethod{}s 
  to the backbone diffusion U-Net's output blocks. We use output block D here for illustration. We insert three \adaptermethod{}s to the output blocks of the same feature map size by default.
  }
\label{fig:appendix_ablation_number_of_blocks} 
\end{figure}

\begin{figure}[h]
  \centering
  \includegraphics[width=.45\linewidth]{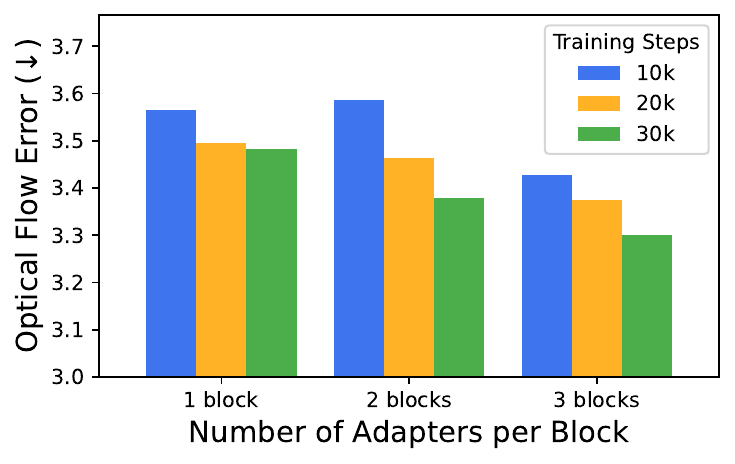}
  \vspace{-2mm}
  \caption{Comparison of inserting different numbers of \adaptermethod{}s to each U-Net output block. The metrics are calculated from 150 videos from DAVIS 2017 dataset. We insert 3 \adaptermethod{}s to each output block by default.
  }
\label{fig:appendix_i2vgenxl_123blocks} 
\end{figure}

As described in \cref{subsec:method_adapter},
each \adaptermethod{} module consists of four components:
spatial convolution (SC), temporal convolution (TC), spatial attention (SA), and temporal attention (TA).
We experiment with different architecture combinations of the adapter components for image and video control, and plot the results in \cref{fig:ablation_bubble_plot_adapter_architecture_sdxl_i2vgenxl}.
Compared to X-Adapter~\cite{ran2023x}, which uses a stack of three spatial convolution modules (\ie{}, ResNet~\cite{resnet2016} blocks) for adapters, and VideoComposer~\cite{wang2024videocomposer}, which employs spatial convolution + temporal attention for spatiotemporal condition encoder,
we explore a richer combination that enhances global understanding of spatial information through spatial attention and improves temporal ability via a combination of temporal convolution and temporal attention.
For \textbf{image control} (\cref{fig:ablation_bubble_plot_adapter_architecture_sdxl_i2vgenxl} left),
we find that the combining of SC+SA is more effective than stacking SC or SA layers only.
Stacking SC+SA twice further improves the visual quality (FID) slightly but hurts the spatial control (MSE) as a tradeoff. Stacking SC+SA three times hurts the performance due to insufficient training. We use the single SC+SA layer for image \adaptermethod{} by default.
For \textbf{video control} (\cref{fig:ablation_bubble_plot_adapter_architecture_sdxl_i2vgenxl} right),
we find that SC+TC+SA+TA shows the best balance of visual quality (FID) and spatial control (optical flow error). Notably,
we find that the combinations with both temporal layers, SC+TC+SA+TA and (TC+TA)*2, achieve the lowest optical flow error.
We use SC+TC+SA+TA for video \adaptermethod{} by default.

\subsubsection{Where to fuse \adaptermethod{} outputs in backbone diffusion}

We compare the integration of \adaptermethod{} outputs at different positions of video diffusion backbone model.
As illustrated in 
\cref{fig:adapter_method},
we experiment with integrating \adaptermethod{} outputs to different positions of I2VGen-XL's  U-Net: middle block, output block A, output block B, output block C, and output block D.
Specifically, we compared our default design (Mid + Out ABCD) with four other variants (Out ABCD, Out ABC, Out AB, and Out A) that gradually remove \adaptermethod{}s from the middle block and output blocks at positions from B to D.
As shown in \cref{fig:appendix_i2vgenxl_abcdm}, removing the \adaptermethod{}s from the middle block and the output block D does not lead to a noticeable increase in FID or optical flow error (\ie{}, the performances of `Mid+Out ABCD', `Out ABCD', and `Out ABC' are similar in both left and right plots).
However, \cref{fig:appendix_i2vgenxl_abcdm} (right) shows that removing \adaptermethod{}s from block C causes a significant increase in optical flow error.
Therefore, we recommend users retain our \adaptermethod{}s in the mid and output blocks A/B/C to ensure good performance.

\subsubsection{Number of \adaptermethod{}s in each output block position}
As illustrated in \cref{fig:adapter_method}, there are three output blocks for each feature map dimension in the video diffusion model (represented by $\times3$ in each output block).
Here, we conduct an ablation study by adding \adaptermethod{}s to only one or two of the three output blocks of the same feature size.
The motivation is that using fewer \adaptermethod{}s can almost linearly decrease the number of trainable parameters, thereby reducing GPU memory usage during training.
We visualize the architectural changes with output block D as an example in \cref{fig:appendix_ablation_number_of_blocks}. We insert \adaptermethod{}s for three blocks as our default setting. As observed in \cref{fig:appendix_i2vgenxl_123blocks}, reducing the number of \adaptermethod{}s increases the optical flow error. Therefore, we recommend adding \adaptermethod{}s to each output block to maintain optimal performance.

\subsection{Adaptation to DiT-Based Backbones}
\label{sec:appendix_dit_based_backbones}

\begin{figure}[t]
  \centering
\includegraphics[width=.99\linewidth]{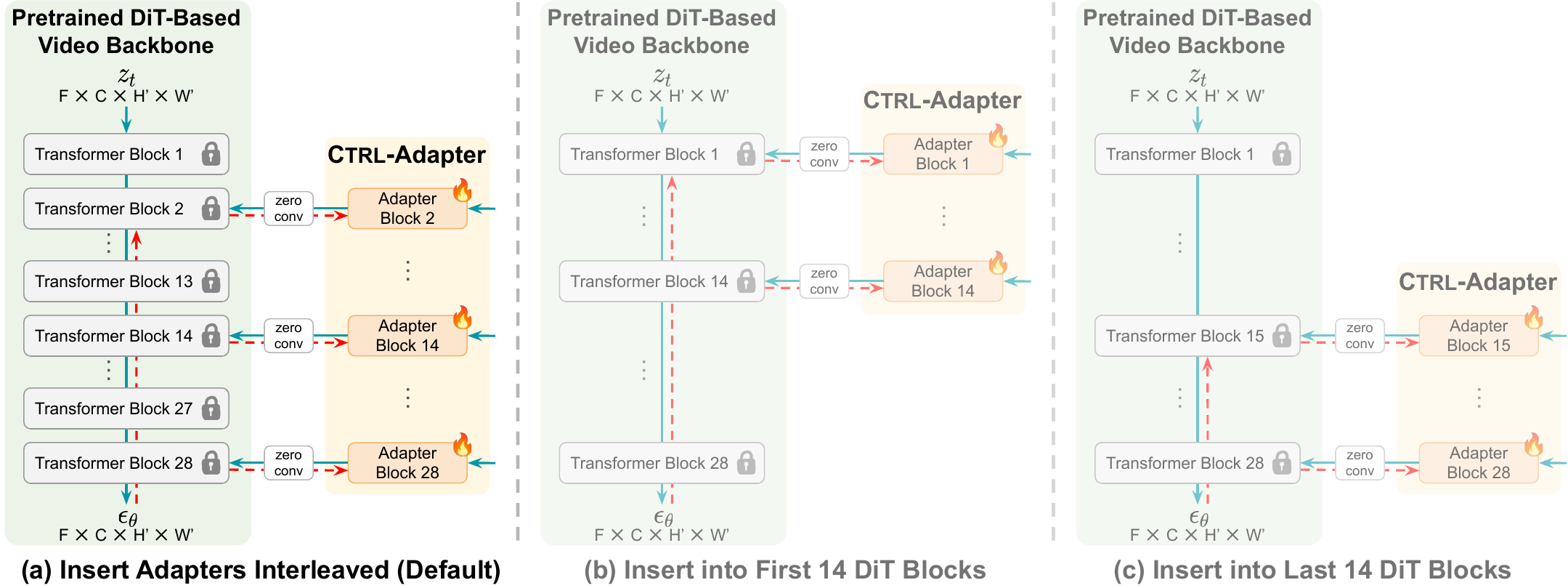}
  \caption{Visualization of different routing methods for combining multiple \cnet{} outputs.
  We use (a) as our default setting, and show the settings (b) and (c) as ablations.
  }
\label{fig:method_dit_adapter_latte} 
\end{figure}

As illustrated in \cref{sec:method}, we have observed that the spatial features encoded in the U-Net of \cnet{}s and
the DiT blocks are structurally different (see \cref{fig:feature_map_visualization} for visualization of such observation). Therefore, mapping all
middle/output blocks of \cnet{} to DiT blocks might not be the optimal solution. In \cref{fig:method_dit_adapter_latte}, we implement three different strategies to insert \adaptermethod{}s to the DiT blocks. Specifically, variant (a) inserts \adaptermethod{}s interleavingly into the DiT blocks, while variant (b) and (c) insert \adaptermethod{}s to the first 14 and the last 14 DiT blocks respectively. In \Cref{table:ablation_adapter_location_latte}, we perform quantitative analysis of these three variants on the DiT-based video generation model, Latte~\cite{ma2024latte}, with soft edge as control condition. As we can see, inserting \adaptermethod{}s interleavingly into the DiT blocks gives the best performance. This is consistent with our finding: since all DiT blocks encode global information of the generated objects, it is optimal to treat these blocks equally, rather than inserting \adaptermethod{}s only at the beginning or end. Between locations A and B, we use location A as our default setting because its feature map size ($64 \times 64$) directly matches the features of the DiT blocks (also $64 \times 64$) without resizing.

After finalizing where to insert \adaptermethod{}s, the next question is which block(s) of the \cnet{} we should create \adaptermethod{}s to map from. In \Cref{table:ablation_adapter_location_pixart}, we implement several variants on the DiT-based image generation model, PixArt-$\alpha$~\cite{chen2024pixartalpha}, including mapping from the block(s) at location A, location B, location C, and location D, respectively (see \cref{fig:adapter_method} for the definitions of these locations). As we can see, mapping from location A or location B gives the best performance. Again, this is consistent with our findings in \cref{fig:feature_map_visualization}, since feature maps at locations C and D are too coarse to be informative. Moreover, we implemented two additional variants: (1) combining the \cnet{} features from locations A and B (\ie{}, Output Blocks A+B), and (2) mapping more blocks from the same location (\ie{}, the second and third columns in \Cref{table:ablation_adapter_location_pixart}). However, neither of these approaches provides sufficient gain compared to mapping a single block from location A or B. Therefore, we use mapping one block from location A as our default setting in our main paper.

\begin{table}[t]
    \caption{
Ablation of inserting \adaptermethod{}s to different DiT blocks in Latte~\cite{ma2024latte}. Visualization of the three architecture variants (interleaved, first half, and second half) are shown in \cref{fig:method_dit_adapter_latte}. We use soft edge as control condition here for evaluation.
    }
    \label{table:ablation_adapter_location_latte}
    \centering
    \resizebox{0.9\linewidth}{!}{
    \begin{tabular}{cc cc cc}
        \toprule
         \multicolumn{2}{c}{(a) Interleave (default)} & \multicolumn{2}{c}{(b) First Half} & \multicolumn{2}{c}{(c) Second Half} \\
        \cmidrule(lr){1-2} \cmidrule(lr){3-4} \cmidrule(lr){5-6}
        FID ($\downarrow$) &  Optical Flow Error ($\downarrow$) & FID ($\downarrow$) &  Optical Flow Error ($\downarrow$) & FID ($\downarrow$) &  Optical Flow Error ($\downarrow$)\\
        \midrule
         \textbf{18.32} & \textbf{2.98} & 19.66 & 3.09 & 23.18 & 3.31 \\
        \bottomrule
    \end{tabular}
    }
\end{table}

\begin{table}[t]
    \caption{
Ablation study of mapping \cnet{} features from different locations, and mapping different number of blocks from the same location to the DiT blocks.
The best numbers in each row are \textbf{bolded}, and the best numbers in each column are \underline{underscored}.
    }
    \label{table:ablation_adapter_location_pixart}
    \centering
    \resizebox{0.79\linewidth}{!}{
    \begin{tabular}{l cc cc cc}
        \toprule
        \multirow{2}{*}{Insert Locations}  & \multicolumn{2}{c}{1 Block (default)} & \multicolumn{2}{c}{2 Blocks} & \multicolumn{2}{c}{3 Blocks} \\
        \cmidrule(lr){2-3} \cmidrule(lr){4-5} \cmidrule(lr){6-7}
        & FID ($\downarrow$) &  SSIM ($\downarrow$) & FID ($\downarrow$) &  SSIM ($\downarrow$) & FID ($\downarrow$) &  SSIM ($\downarrow$)\\
        \midrule
        \rowcolor{lightblue}    
        Output Block A (default) & \textbf{17.90} & \textbf{0.6802} & 19.08 & 0.6971 & \underline{19.28} & 0.6855 \\
        Output Block B & \textbf{18.23} & \underline{0.6712} & \underline{18.61} & \underline{0.6720} & 21.47 & \textbf{\underline{0.6549}} \\
        Output Blocks A+B & \underline{17.52} & 0.6812 & - & - & - & -\\
        Output Block C & 22.22 & 0.5273 & - & - & - & - \\
        Output Block D & 34.16 & 0.3506 & - & - & - & -\\
        \bottomrule
    \end{tabular}
    }
\end{table}

\subsection{Skipping Latent from \cnet{} Inputs}
\label{sec:ablation_skipping_latent}

\begin{table}[h]
    \caption{
    Skipping latent from \cnet{} inputs helps \adaptermethod{} for (1) adaptation to backbone models with different noise scales and (2) video control with sparse frame conditions. We evaluate SVD and I2VGen-XL on depth maps and scribbles as control conditions respectively.
    }
    \label{table:ablation_skip_latent}
    \centering
    \resizebox{0.85\linewidth}{!}{
    \begin{tabular}{l c cc}
        \toprule
        \rowcolor{white}
        \multirow{1}{*}{Method}  & \multirow{1}{*}{Latent $\bm{z}$ is given to \cnet{}}
        & FID ($\downarrow$) & Optical Flow Error ($\downarrow$)   \\
        \midrule 
        \textcolor{gray}{Adaptation to different noise scales}
         \\
        SVD~\cite{blattmann2023stable} + \adaptermethod{} & \ForestGreencheck & 4.48 & 2.77   \\       
        \rowcolor{lightblue}
        SVD~\cite{blattmann2023stable} + \adaptermethod{} & \redcross & \textbf{3.82} & \textbf{2.96}  \\
        \midrule
        \textcolor{gray}{Sparse frame conditions}\\
        I2VGen-XL~\cite{zhang2023i2vgen} + \adaptermethod{} & \ForestGreencheck & 7.20 & 5.13\\
        \rowcolor{lightblue}
        I2VGen-XL~\cite{zhang2023i2vgen} + \adaptermethod{} & \redcross & \textbf{5.98} &  \textbf{4.88} \\
        
        \bottomrule
    \end{tabular}
    }
\end{table}

As described in \cref{subsec:method_adapter},
we find skipping the latent $z$ from \cnet{} inputs can help \adaptermethod{} to more robustly handle
(1) adaption to the backbone with noise scales different from SDv1.5, such as SVD
and
(2) video control with sparse frame conditions.
For the first scenario, we can see from \Cref{table:ablation_skip_latent} that skipping latents in SVD leads to better visual quality (lower FID), but slightly worse spatial control (higher optical flow error). This is reasonable since skipping the noisy latents can avoid introducing large noise into the \cnet{}, but it also risks losing information encoded in the latents.
For the second scenario, skipping latents results in both better visual quality and better spatial control, as adding dense noisy latents can make the sparse control conditions less informative.

\subsection{Different weighing modules for multi-condition generation}
\label{appendix_sec:different_weighting_schemes}

\begin{table}[t]
    \caption{
    Comparison of global weighting methods for
    multi-condition video generation (see \cref{fig:appendix_moe_router_architectures} for visualization of the additional weighting methods (a.2, a.3, and a.4) developed based on (a.1) unconditional global weights).
    The control sources are abbreviated as D (depth map), C (canny edge), N (surface normal), S (softedge), Seg (semantic segmentation map), L (line art), and P (human pose).
    }
    \label{tab:multi-source-video-gen-all-results}
    \centering
    \resizebox{0.99\linewidth}{!}{
    \begin{tabular}{l cc cc cc cc}
        \toprule
        & \multicolumn{2}{c}{D+C} & \multicolumn{2}{c}{D+P} & \multicolumn{2}{c}{D+C+N+S} & \multicolumn{2}{c}{D+C+N+S+Seg+L+P} \\
        \cmidrule(lr){2-3} \cmidrule(lr){4-5} \cmidrule(lr){6-7} \cmidrule(lr){8-9} 
        & FID ($\downarrow$) & Flow Error ($\downarrow$) & FID ($\downarrow$) & Flow Error ($\downarrow$)& FID ($\downarrow$) & Flow Error ($\downarrow$) & FID ($\downarrow$) & Flow Error ($\downarrow$) \\
        \midrule
        \multicolumn{1}{l}{\textcolor{gray}{Baseline}} \\
        \rowcolor{lightblue}
        Equal Weights  & 8.50 & 2.84 & 11.32 & 3.48 & 8.75 & 2.40 & 9.48 & 2.93 \\
        \midrule
        \multicolumn{1}{l}{\textcolor{gray}{Global MoE Router}} \\
        \rowcolor{lightblue}
        (a.1) Unconditional Global Weights  & 9.14 & 2.89 & 10.98 & 3.32 & 8.39 & 2.36 & 8.18 & 2.48 \\
        (a.2) Timestep Emb. Weights & 9.41 & 3.51 & 11.13 & 3.35 & 9.51 & 2.78 & 8.17 & 2.45   \\
        (a.3) Text/Image Emb. Weights  & 8.73 & 3.16 & 11.35 & 3.37  & 7.91 & 2.76 & 8.83 & 2.48 \\
        (a.4) Timestep + Text/Image Emb. Weights & 8.64 & 3.31 & 10.69 & 3.43 &  8.09 & 2.69 & 8.51 & 2.43  \\
        \midrule
        \multicolumn{1}{l}{\textcolor{gray}{Patch-Level MoE Router}} \\
        \rowcolor{lightblue}
        (b) MLP Weights  & 8.40 & \textbf{2.34} & 9.37 & \textbf{3.17} & 7.87 & \textbf{2.11} & 8.26 & \textbf{2.00} \\
        \rowcolor{lightblue}
        (c) Q-Former Weights  & \textbf{7.54} & 2.39 & \textbf{9.22} & 3.22 & \textbf{7.72} & 2.31 & \textbf{8.00} & {2.08} \\
        \bottomrule
    \end{tabular}
    }
\end{table}

\begin{figure}[t]
  \centering
  \includegraphics[width=.9\linewidth]{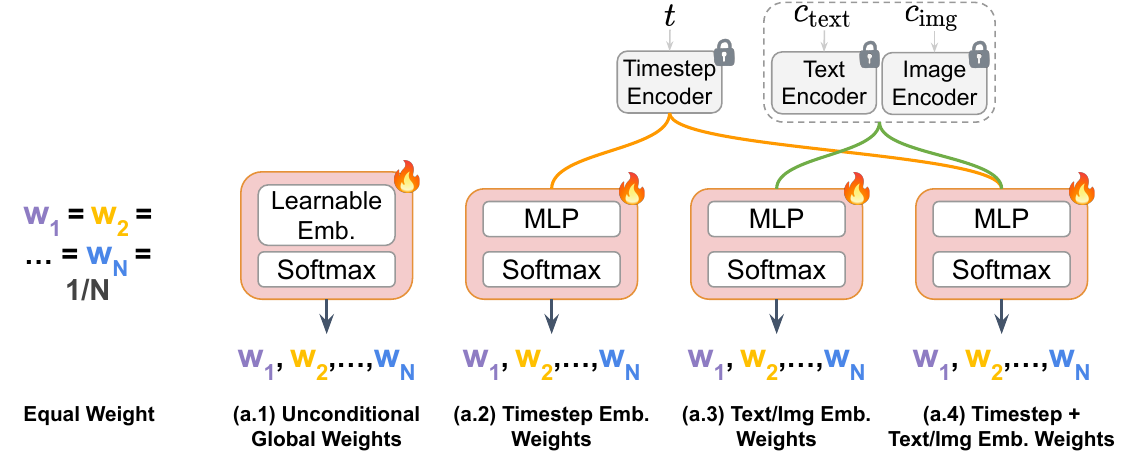}
  \vspace{-2mm}
  \caption{Visualization of different {\color{orange}\textbf{global}} MoE routing methods.
  }
\label{fig:appendix_moe_router_architectures} 
\end{figure}

\begin{table}[t]
    \caption{
    Ablation of training a unified \adaptermethod{} v.s. training an individual \adaptermethod{} for each condition. The results are evaluated with SDXL as the backbone model.
    }
    \label{table:ablation_training_strategy}
    \centering
    \resizebox{0.7\linewidth}{!}{
    \begin{tabular}{l cc cc}
        \toprule
        \multirow{2}{*}{Training Strategy}  & \multicolumn{2}{c}{Depth Map} & \multicolumn{2}{c}{Canny Edge} \\
        \cmidrule(lr){2-3} \cmidrule(lr){4-5} 
        & FID ($\downarrow$) &  SSIM ($\uparrow$) & FID ($\downarrow$) &  SSIM ($\uparrow$) \\
        \midrule
        \rowcolor{lightblue}  
        Individual \adaptermethod{} (default) & \textbf{19.26} & \textbf{0.8534} & \textbf{21.04} & \textbf{0.5806} \\
        Unified \adaptermethod{}  & 19.95 & 0.8437  & 22.31 & 0.5684 \\
        \bottomrule
    \end{tabular}
    }
\end{table}

For multi-condition generation described in \cref{subsec:method_multi_source},
in addition to the simple unconditional global weights,
we also experimented with 
learning a router module that takes additional inputs such as diffusion time steps and image/text embeddings and outputs weights for different \cnet{}s.
Specifically, we introduce three variants based on (a.1) unconditional global weights, which are 
(a.2) MLP router - taking timestep as inputs;
(a.3) MLP router - taking image/text embedding as inputs;
and
(a.4) MLP router - taking timestep and image/text embedding as inputs.
The MoE router in these variants are constructed as a 3-layer MLP.
We illustrate the five methods in \cref{fig:appendix_moe_router_architectures}.

\Cref{tab:multi-source-video-gen-all-results}
show that 
all four global weighting schemes for fusing different \cnet{} outputs
perform effectively, and no specific method outperforms other methods with significant margins in all settings. With no surprise, patch-level MoE router performs consistently better than global MoE router in all control settings. 
Testing the effectiveness of incorporating text/image/timestep embeddings to patch-level MoE routers are left for future work.

\section{Additional Quantitative Analysis}
\label{appendix_sec:additional_quant_analysis}

\subsection{Train individual \adaptermethod{} v.s. train a unified \adaptermethod{}}\label{sec:ablation_invidual_or_unified_adapter}

In our main paper, we train \adaptermethod{} for each control conditions. An interesting question to ask is: {\sl{can we have a single and unified \adaptermethod{} that works for all control conditions?}}
In this part, we perform such analysis with SDXL as the backbone model on a new training strategy. Specifically, during each training step, we randomly choose one control condition from a pool of 8 control conditions, including depth map, canny edge, soft edge, normal map, semantic segmentation, line art, user scribbles, and human pose. We train this variant with the same hyper-parameter settings as the depth map or canny edge \adaptermethod{} mentioned in our main paper.

As shown in \Cref{table:ablation_training_strategy}, training a unified \adaptermethod{} across all control conditions does suffer from a performance decrease. However, this decrease is not very significant. Therefore, from a practical point of view, if a user has computational constraints but still needs to work on multiple control conditions, training a unified \adaptermethod{} can be a viable workaround.

\subsection{Trade-off between Visual Quality and Spatial Control}
\label{sec:tradeoff_visual_quality_spatial_control}

\begin{figure}[h]
\minipage{0.45\textwidth}
  \includegraphics[width=\linewidth]{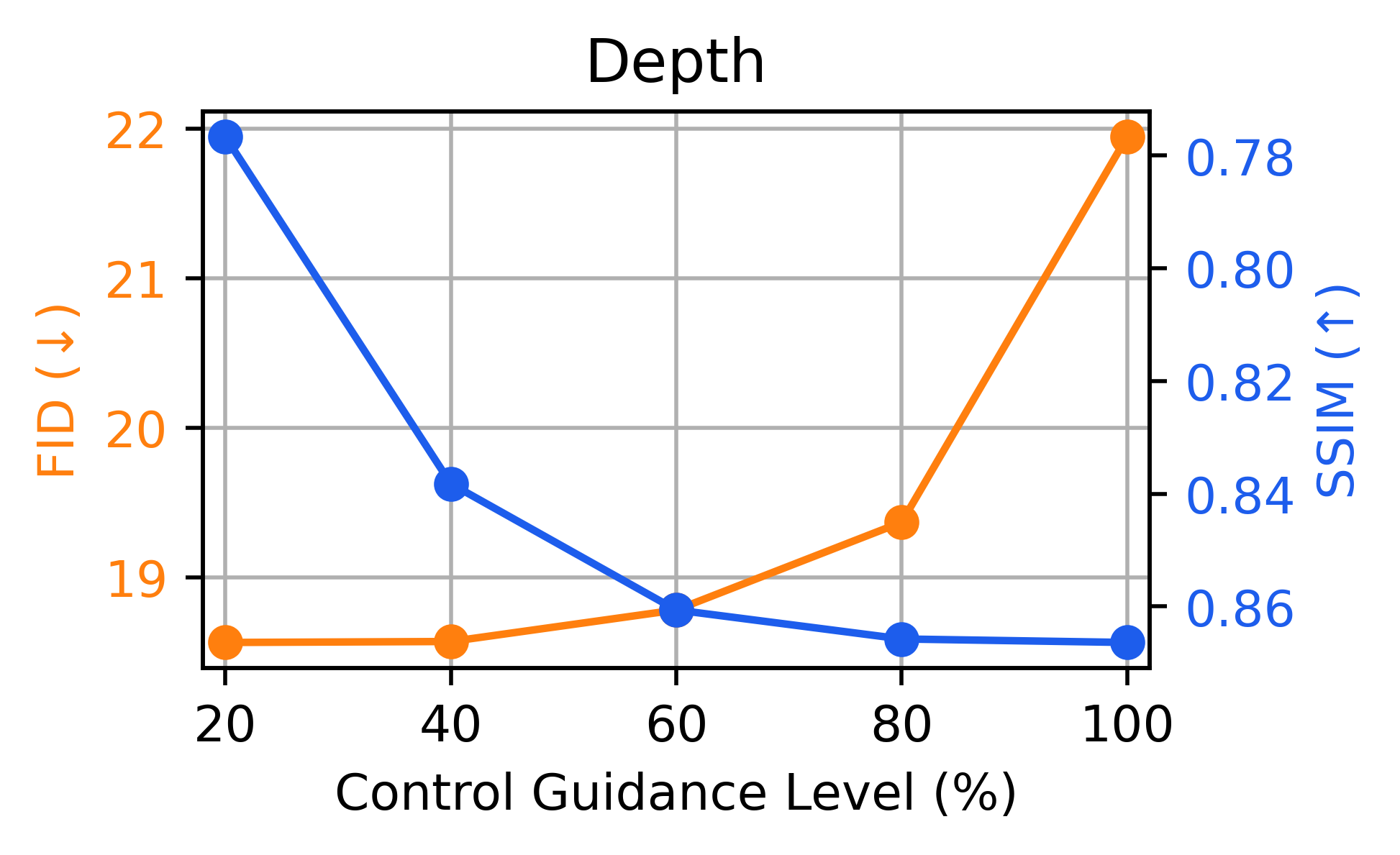}
\endminipage\hfill
\minipage{0.45\textwidth}
  \includegraphics[width=\linewidth]{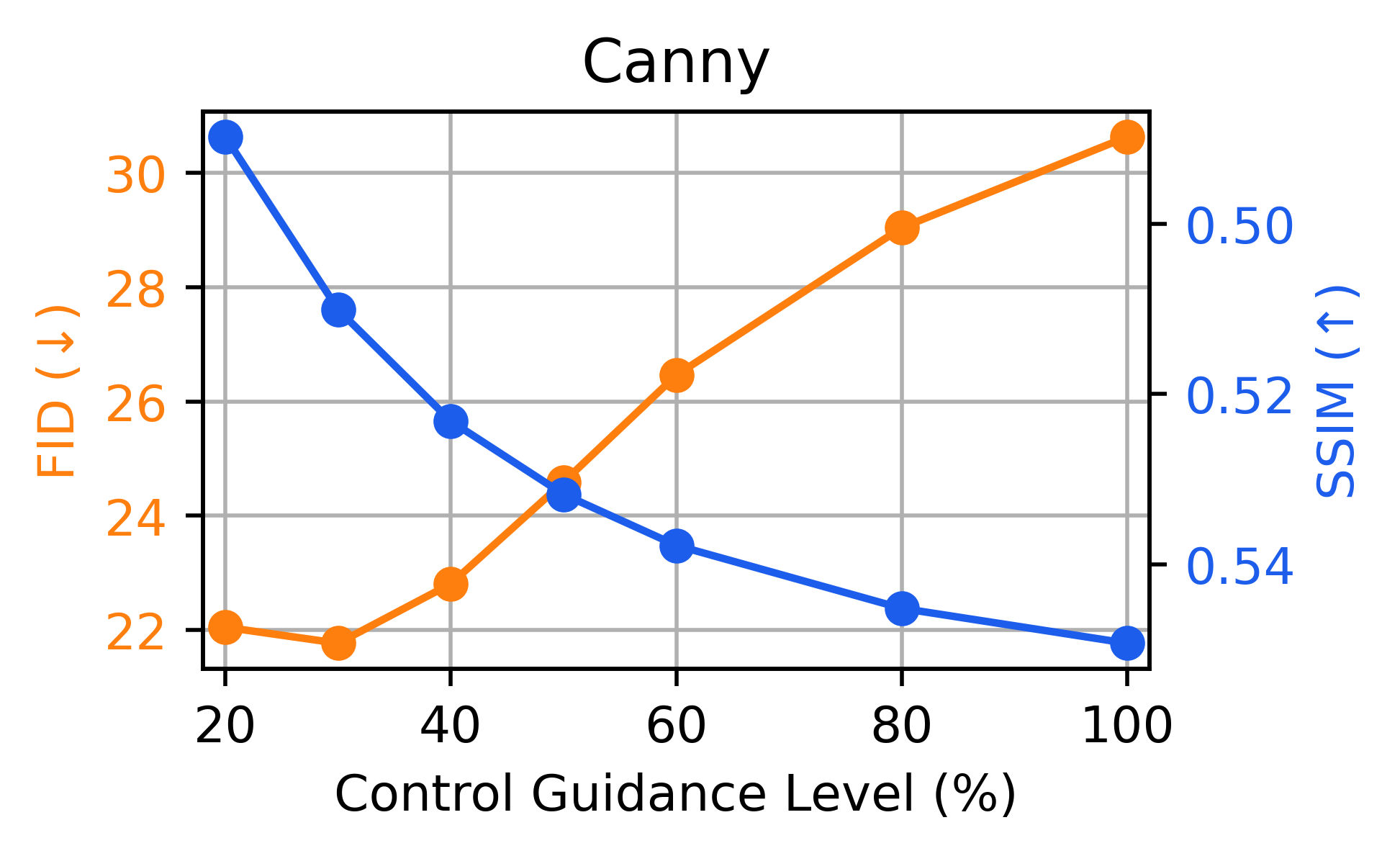}
\endminipage\hfill
  \caption{Trade-off between generated visual quality (FID) and spatial control accuracy (SSIM) on \textbf{SDXL}. Control guidance level of $x$ represents that we apply \adaptermethod{} in the first $x\%$ of the denoising steps during inference. A control guidance level between $30\%$ and $60\%$ usually achieves the best balance between image quality and spatial control accuracy.
  }
\label{fig:appendix_sdxl_canny_fid_ssim_tradeoff}
\end{figure}

\begin{figure}[h]
\minipage{0.45\textwidth}
  \includegraphics[width=\linewidth]{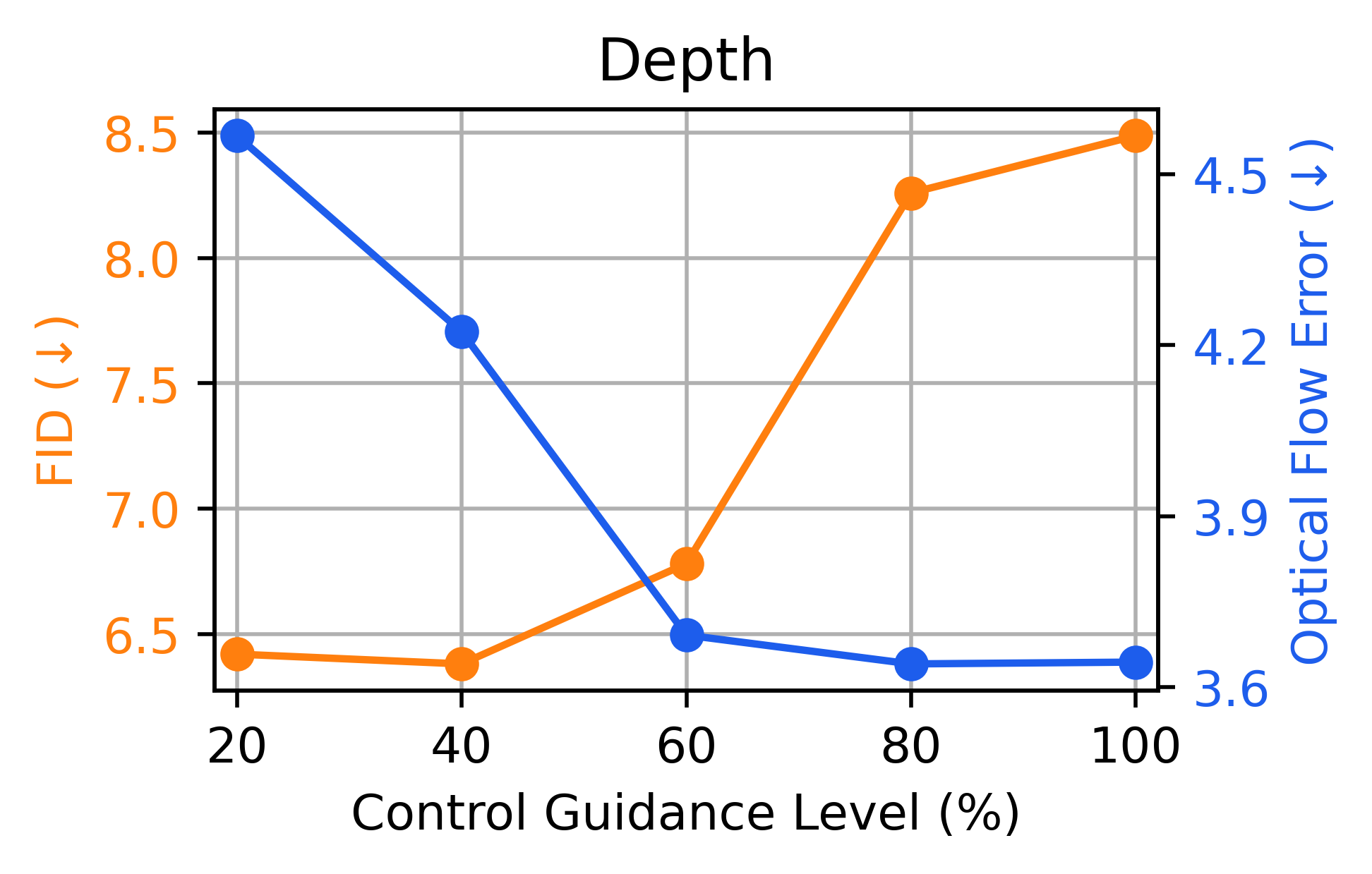}
\endminipage\hfill
\minipage{0.45\textwidth}
  \includegraphics[width=\linewidth]{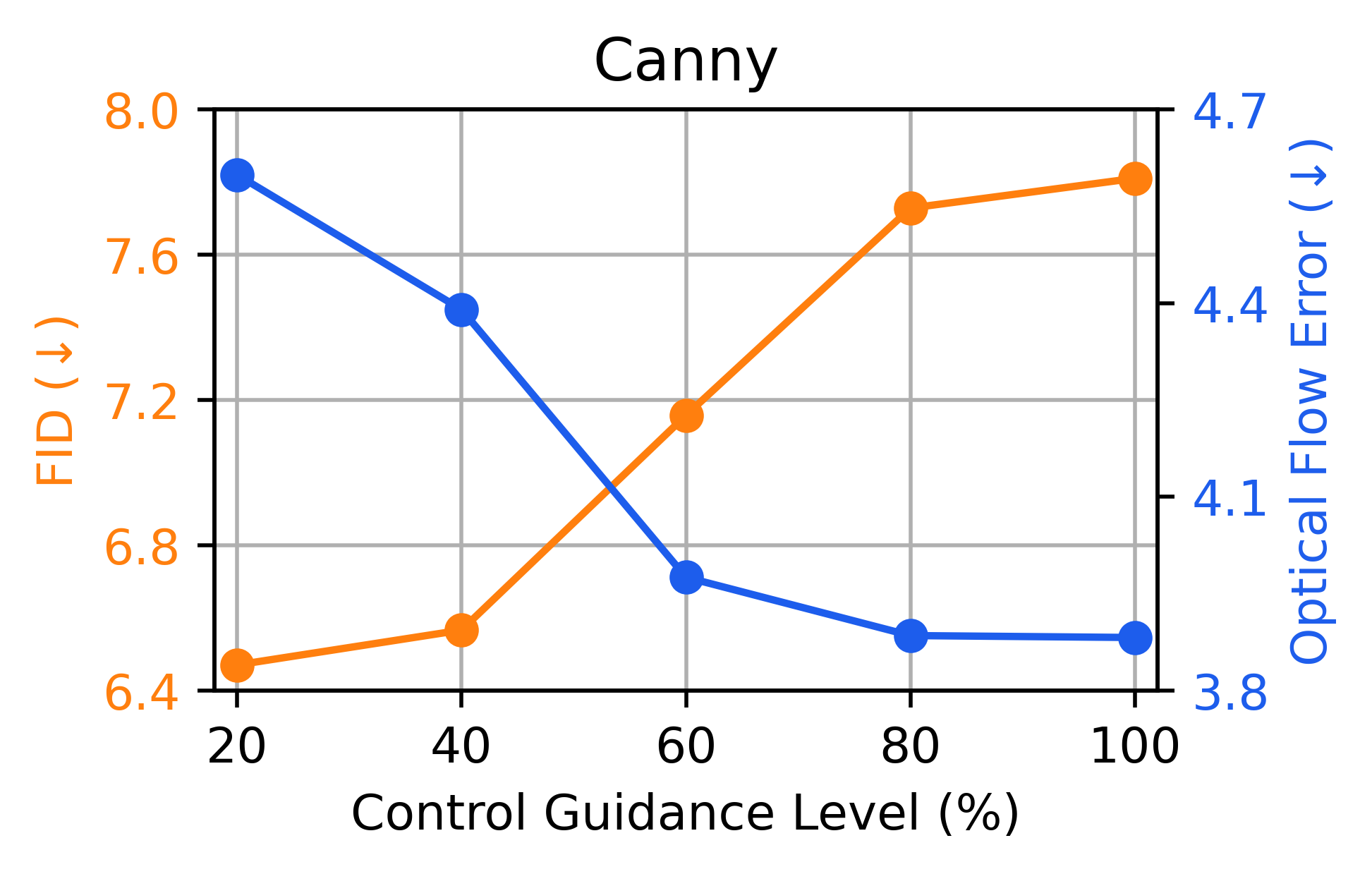}
\endminipage\hfill
  \caption{Trade-off between generated visual quality (FID) and spatial control accuracy (Optical Flow Error) on \textbf{SVD}. Control guidance level of $x$ represents that we apply \adaptermethod{} in the first $x\%$ of the denoising steps during inference. A control guidance level between $40\%$ and $60\%$ usually achieves the best balance between image quality and spatial control accuracy
  }
  \label{fig:appendix_svd_fid_flow_error_tradeoff}
\end{figure}

\begin{figure}[h]
\minipage{0.45\textwidth}
  \includegraphics[width=\linewidth]{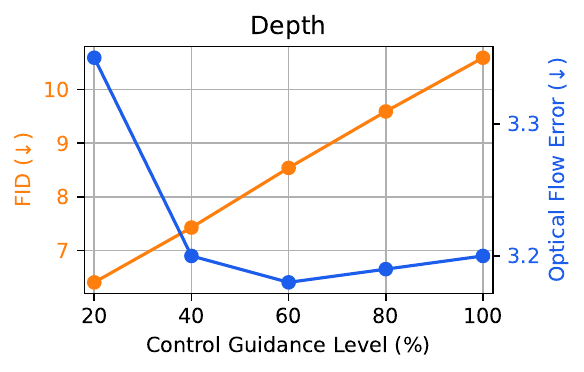}
\endminipage\hfill
\minipage{0.45\textwidth}
  \includegraphics[width=\linewidth]{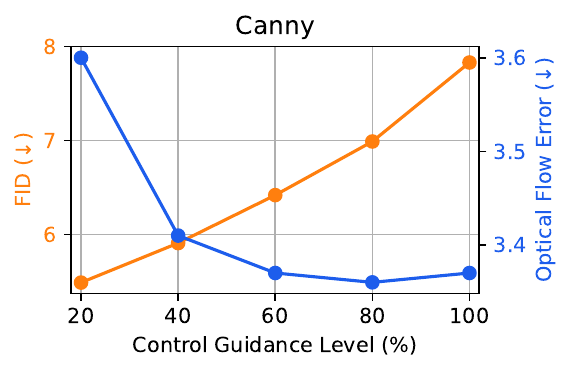}
\endminipage\hfill
  \caption{Trade-off between generated visual quality (FID) and spatial control accuracy (Optical Flow Error) on \textbf{I2VGen-XL}. Control guidance level of $x$ represents that we apply \adaptermethod{} in the first $x\%$ of the denoising steps during inference. A control guidance level between $40\%$ and $60\%$ usually achieves the best balance between image quality and spatial control accuracy.
  }
  \label{fig:appendix_i2vgenxl_fid_flow_error_tradeoff}
\end{figure}

In \cref{fig:appendix_sdxl_canny_fid_ssim_tradeoff},
\cref{fig:appendix_svd_fid_flow_error_tradeoff},
and 
\cref{fig:appendix_i2vgenxl_fid_flow_error_tradeoff},
we show 
the visual quality (FID) and spatial control (SSIM/optical flow error) metrics
with different numbers of denoising steps with spatial control (with the fusion of \adaptermethod{} outputs) on \sdxl{}, \svd{}, and I2VGen-XL backbones respectively.
Specifically, suppose we use $N$ denoising steps during inference, a control guidance level of $x\%$ means that we fuse \adaptermethod{} features to the video diffusion U-Net during the first $x\%\times N$ denoising steps, followed by $(100-x)\%\times N$ regular denoising steps.
In all experiments, we find that increasing the number of denoising steps with spatial control improves the spatial control accuracies (SSIM/optical flow error) but hurts visual quality (FID).

\clearpage
\section{Additional Qualitative Analysis}
\label{appendix_sec:additional_qualitative_analysis}

\subsection{Visualization of Spatial Feature Maps}
\label{sec:appendix_visualization_spatial_feature_maps}

\begin{figure}[h]
  \centering
\includegraphics[width=.99\linewidth]{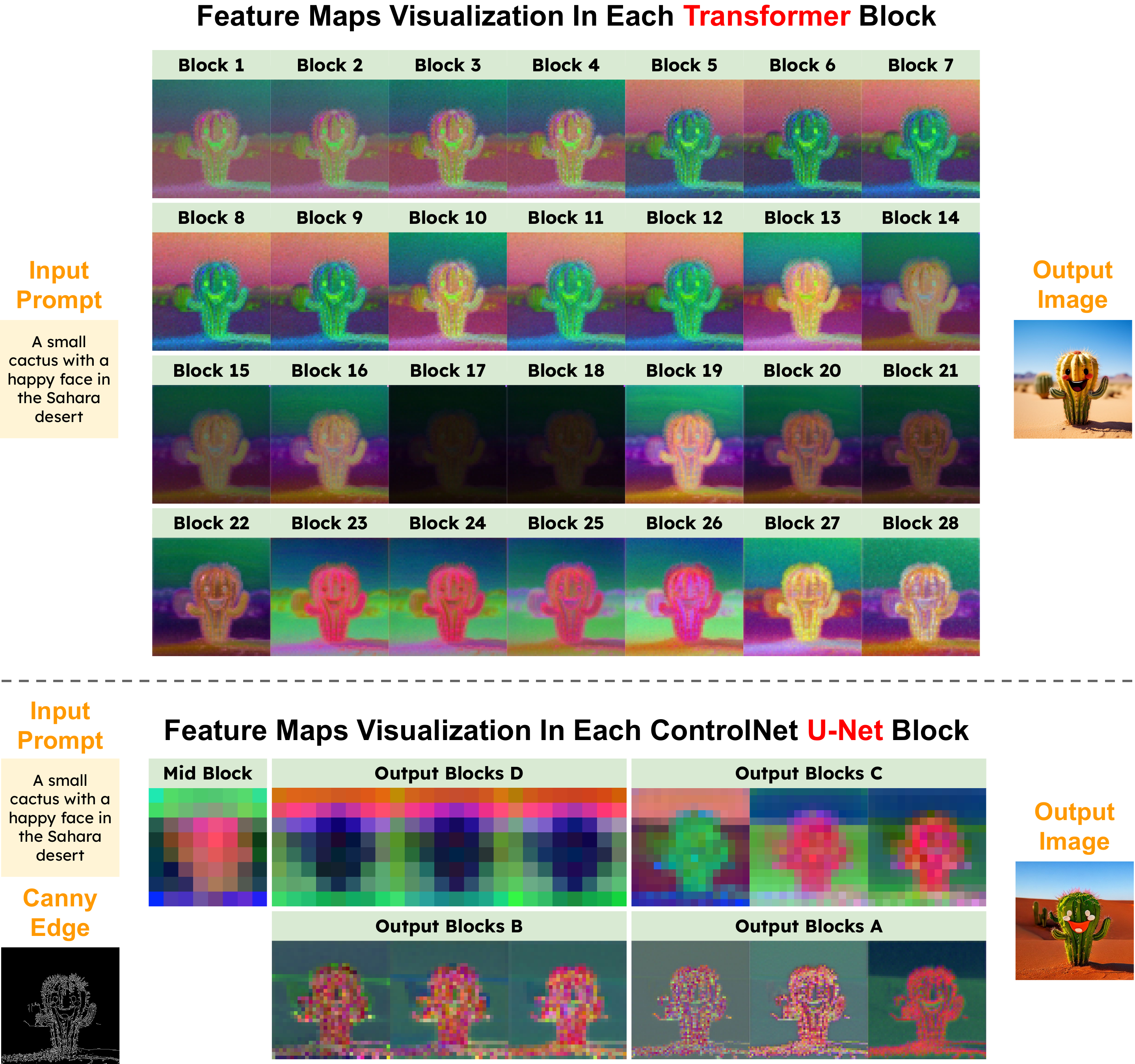}
  \caption{Visualization of spatial feature maps in PixArt-$\alpha$ and canny edge \cnet{}.
  We first extract the spatial features from different DiT blocks and U-Net middle/output blocks at the last denoising step during inference. For each block, we applied PCA to the extracted features and visualized the top three leading components. Almost all 28 DiT blocks capture global and semantic information about the object "cactus", while the U-Net blocks in \cnet{} demonstrate a coarse-to-fine pattern as the feature map size increases. 
  }
\label{fig:feature_map_visualization} 
\end{figure}

As mentioned in \cref{sec:method} and \cref{sec:appendix_dit_based_backbones}, the spatial features encoded in the U-Net of \cnet{}s and the DiT blocks are structurally different. We visualize this difference in \cref{fig:feature_map_visualization}. For the DiT-based model, we use PixArt-$\alpha$ as a representative. We follow the visualization method mentioned in \cite{tumanyan2023plug}. Specifically, we first extract the spatial features from different DiT blocks and U-Net middle/output blocks at the last denoising step during inference. For each block, we applied PCA to the extracted features and visualized the top three leading components.

As shown in \cref{fig:feature_map_visualization}, almost all 28 DiT blocks capture global and semantic information about the object "cactus". This observation is consistent with the findings in \cite{guo2024dit}. On the other hand, the U-Net blocks in \cnet{} demonstrate a coarse-to-fine pattern as the feature map size increases. This indicates that mapping output blocks A/B of \cnet{} to DiT blocks is a better option compared to using middle or output blocks C/D of the \cnet{}.

\clearpage
\subsection{Fast Training Convergence}
\label{appendix_sec:fast_training_convergence}

\begin{figure}[h]
  \centering
  \includegraphics[width=.99\linewidth]{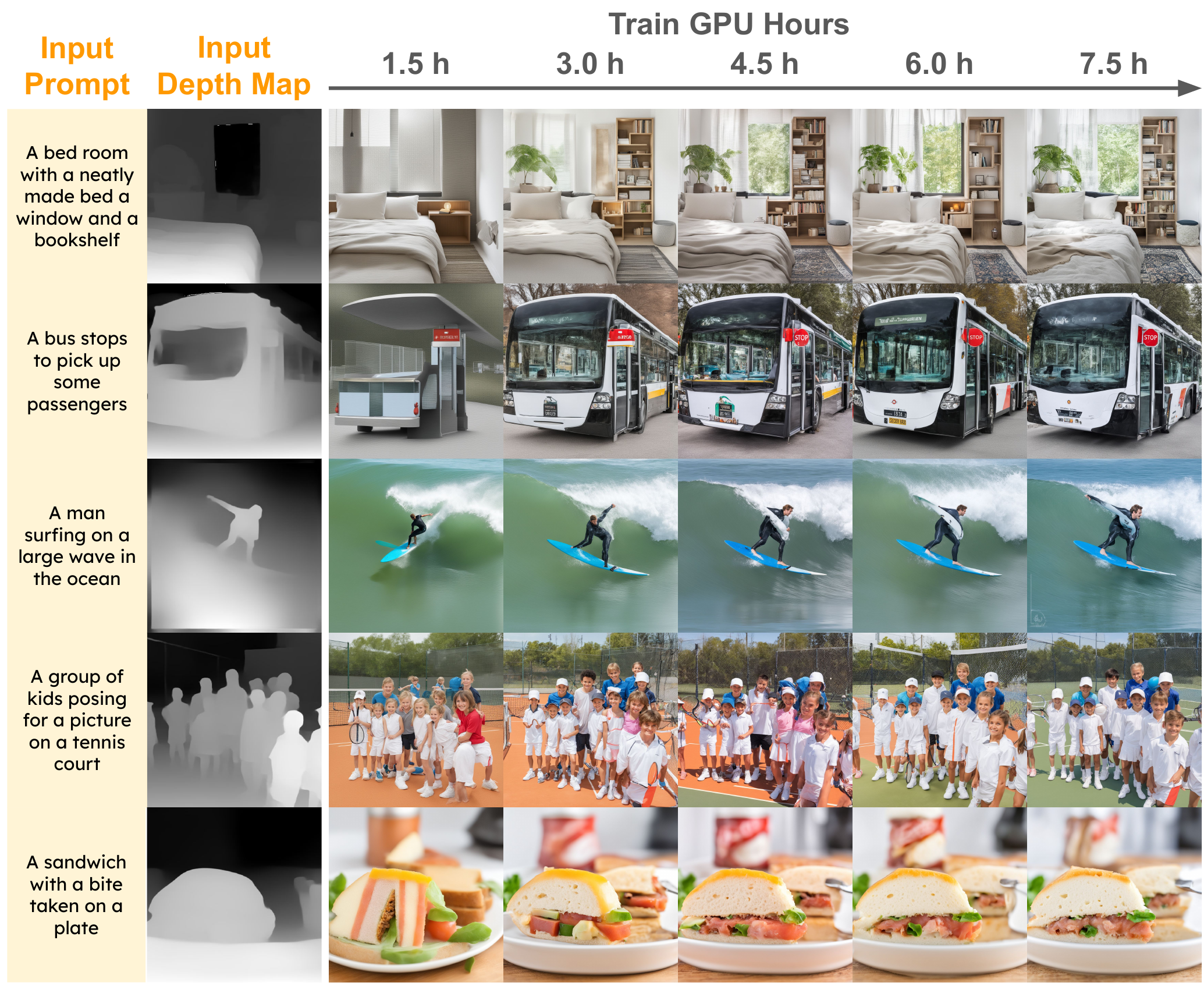}
  \caption{
  Training efficiency of \adaptermethod{} on SDXL backbone.
  Total training GPU hours are measured on 4 A100 80GB GPUs, with batch size per GPU equal to 1.
  }
\label{fig:training_progress_sdxl} 
\end{figure}

In addition to the quantitative results shown in \cref{fig:training_statistics_comparison}, we provide a more straightforward visualization for SDXL depth \cnet{} + \adaptermethod{} training. The training speed test is performed on 4 A100 80GB GPUs, with a batch size of 1 per GPU. As shown in \cref{fig:training_progress_sdxl}, for relatively easy examples (\ie{}, bedroom, sandwich, bus), our \adaptermethod{} training can converge within 4.5 GPU hours (which is equivalent to around 1.125 hours measured in training clock time). For complex examples and those requiring fine details (\ie{}, surfing man, group of kids), our \adaptermethod{} can also converge within around 6 to 7.5 GPU hours (which is equivalent to 1.5 to 1.875 hours measured in training clock time), which proves the training efficiency of our \adaptermethod{}.

\clearpage
\section{Additional Visualization Examples}
\label{appendix_sec:additional_visualization_examples}

We provide more qualitative examples in this section.

\subsection{Video Generation Visualization Examples}
\label{appendix_subsec:video_gen_examples}

\begin{figure}[h]
  \centering
  \includegraphics[width=.99\linewidth]{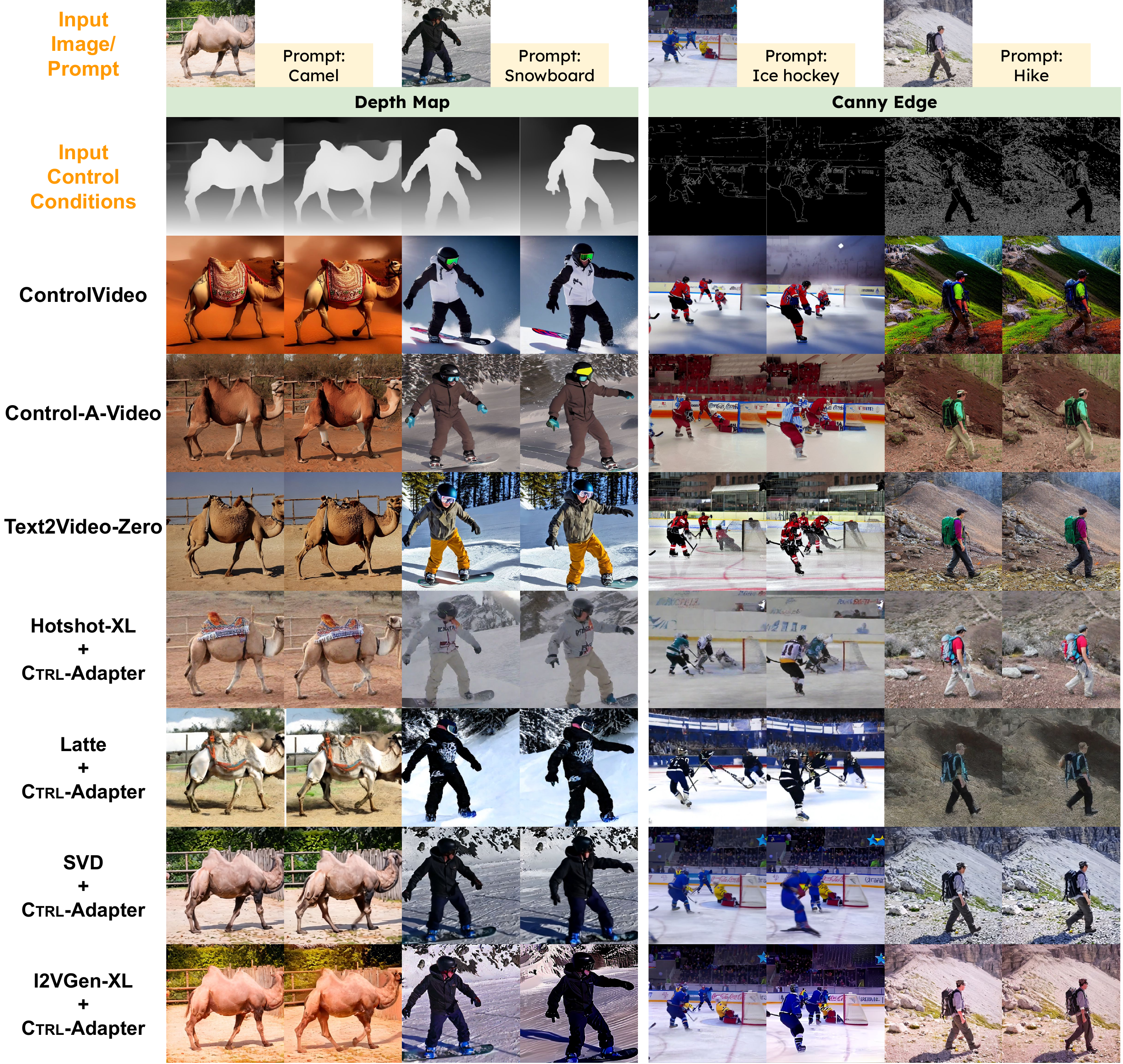}
  \caption{Videos generated from different video control methods and \adaptermethod{} on \davis{} 2017, using depth map (left) and canny edge (right) conditions.
  }
\label{fig:video_visualization_depth_canny_main1} 
\end{figure}

In \cref{fig:video_visualization_depth_canny_main1},
we show video generation results on COCO val2017 split using depth map and canny edge as control conditions. We visualize baseline methods as well as \adaptermethod{}s built on top of Hotshot-XL~\cite{Mullan_Hotshot-XL_2023}, SVD~\cite{blattmann2023stable}, I2VGen-XL~\cite{zhang2023i2vgen}, and Latte~\cite{ma2024latte}.

\begin{figure}[h]
  \centering
  \includegraphics[width=.85\linewidth]{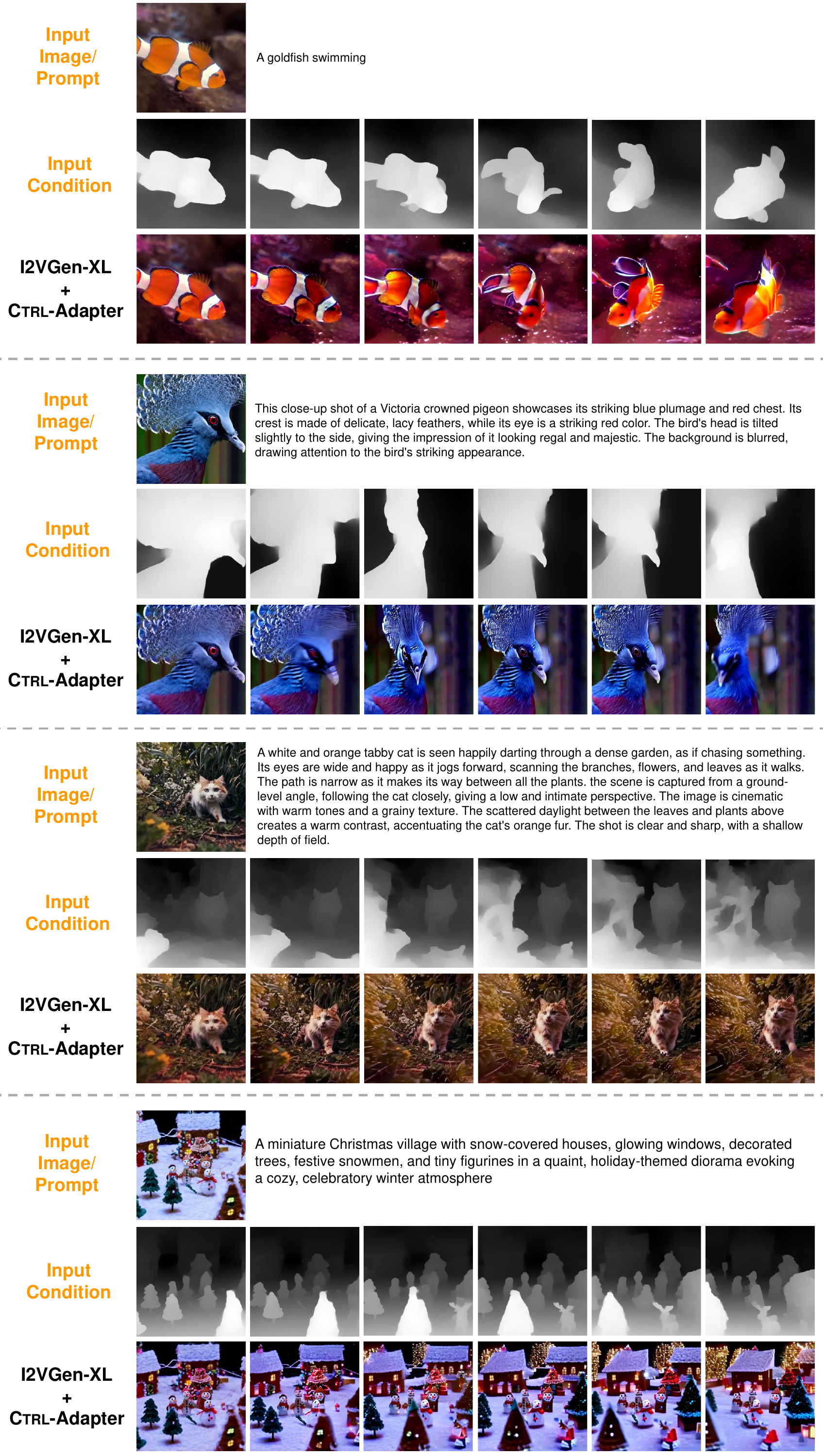}
  \caption{
  Video generation
  with {\color{Blue}\textbf{I2VGen-XL + \adaptermethod{}}} using {\color{orange} \textbf{depth map}} as a control condition.
  }
\label{fig:video_control_depth} 
\end{figure}

\begin{figure}[h]
  \centering
  \includegraphics[width=.85\linewidth]{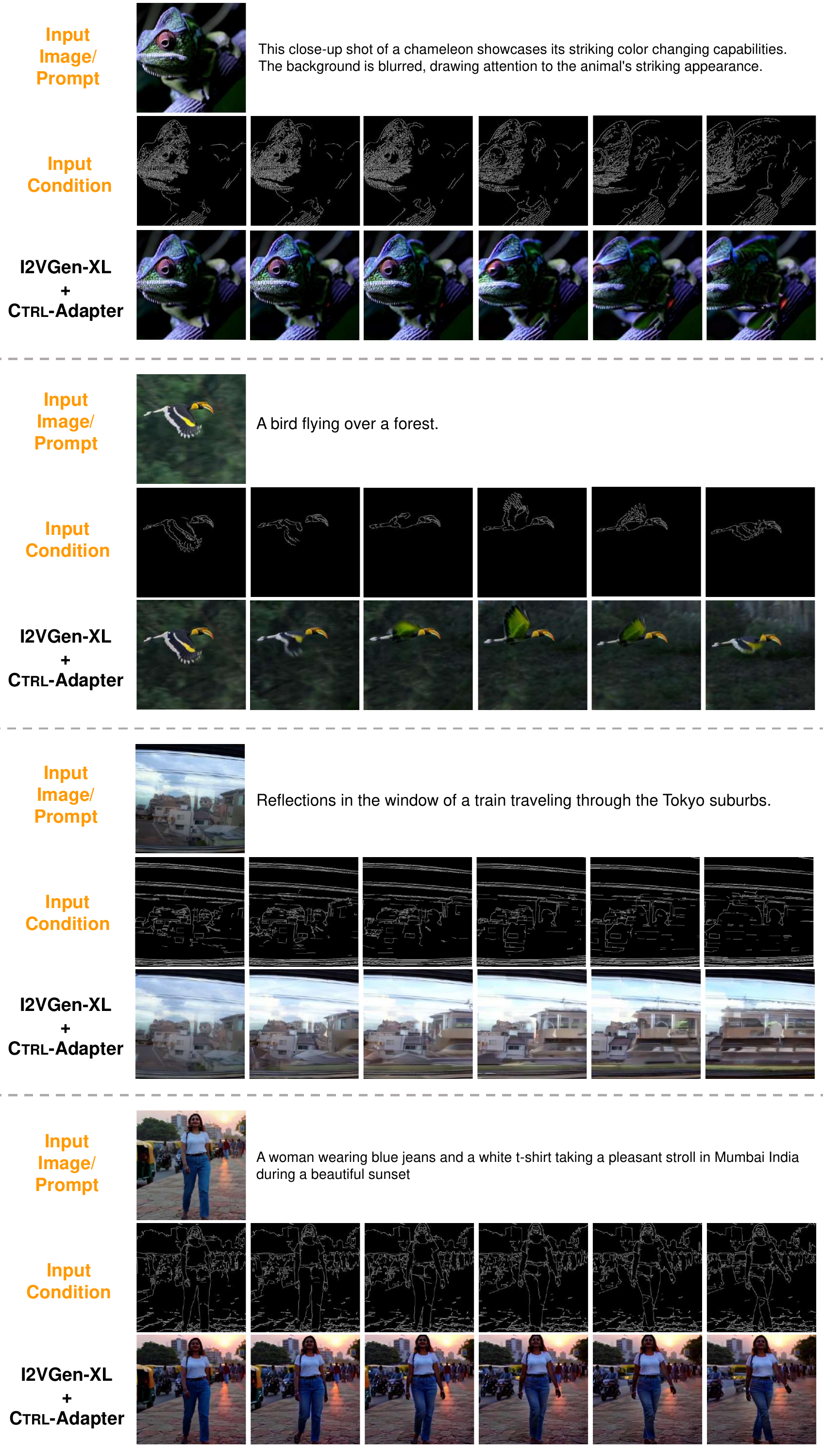}
  \caption{
  Video generation
  with {\color{Blue}\textbf{I2VGen-XL + \adaptermethod{}}} using {\color{orange} \textbf{canny edge}} as control condition.
  }
\label{fig:video_control_canny} 
\end{figure}

In \cref{fig:video_control_depth} and \cref{fig:video_control_canny}, we show video generation results with I2VGen-XL using depth map and canny edge extracted from videos from Sora\footnote{\url{https://openai.com/sora}} and the internet.

\begin{figure}[h]
  \centering
  \includegraphics[width=.82\linewidth]{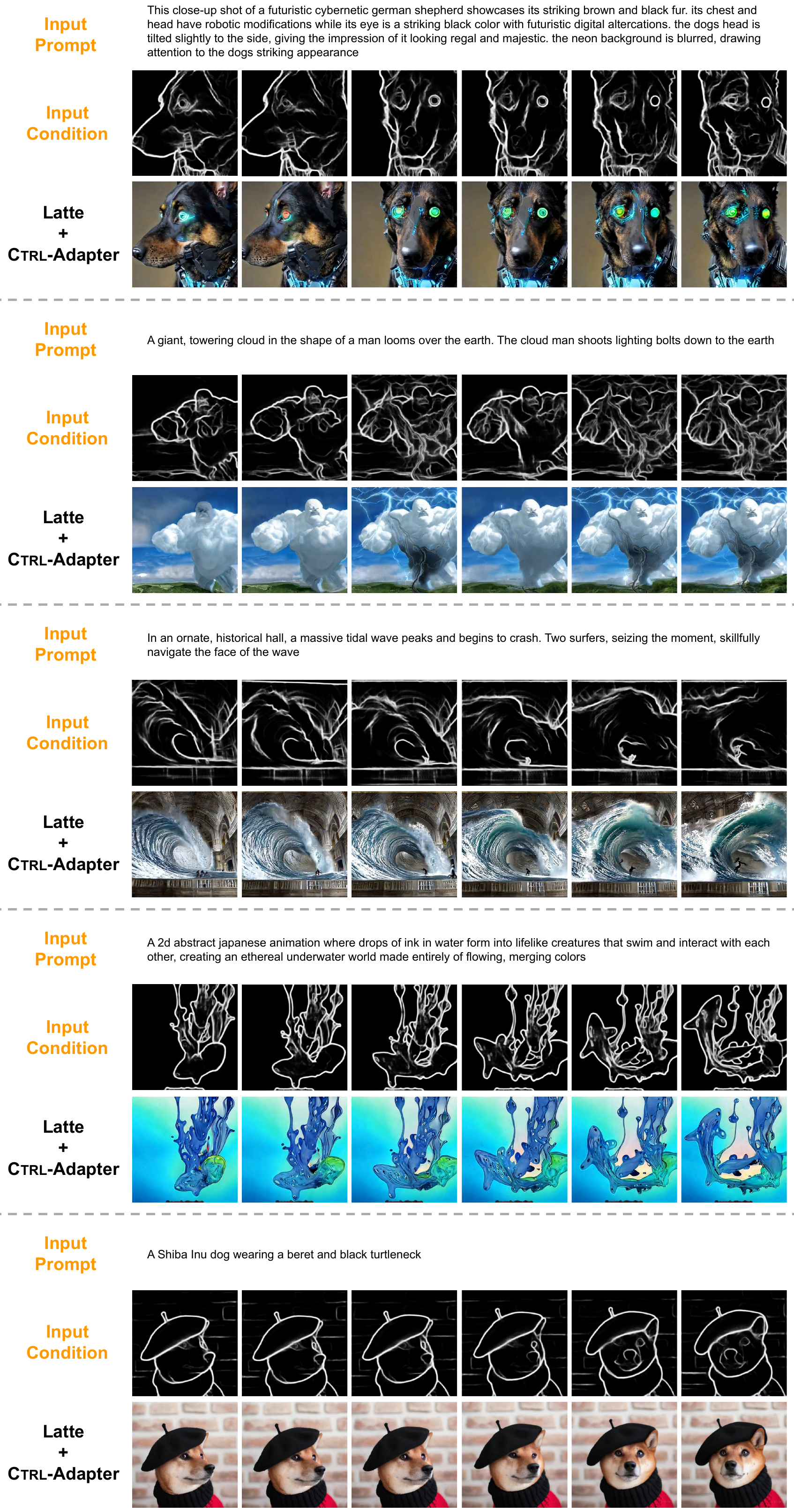}
  \caption{
  Video generation
  with {\color{Blue}\textbf{Latte + \adaptermethod{}}} using {\color{orange} \textbf{soft edge}} as control condition.
  }
\label{fig:video_control_softedge_latte} 
\end{figure}

In \cref{fig:video_control_softedge_latte}, we show video generation results with Latte using soft edge extracted from videos from Sora\footnote{\url{https://openai.com/sora}} and the internet.

\clearpage
\subsection{Multi-Condition Video Generation Visualization Examples}
\label{appendix_subsec:multi_video_gen_examples}

\begin{figure}[h]
  \centering
  \includegraphics[width=.99\linewidth]{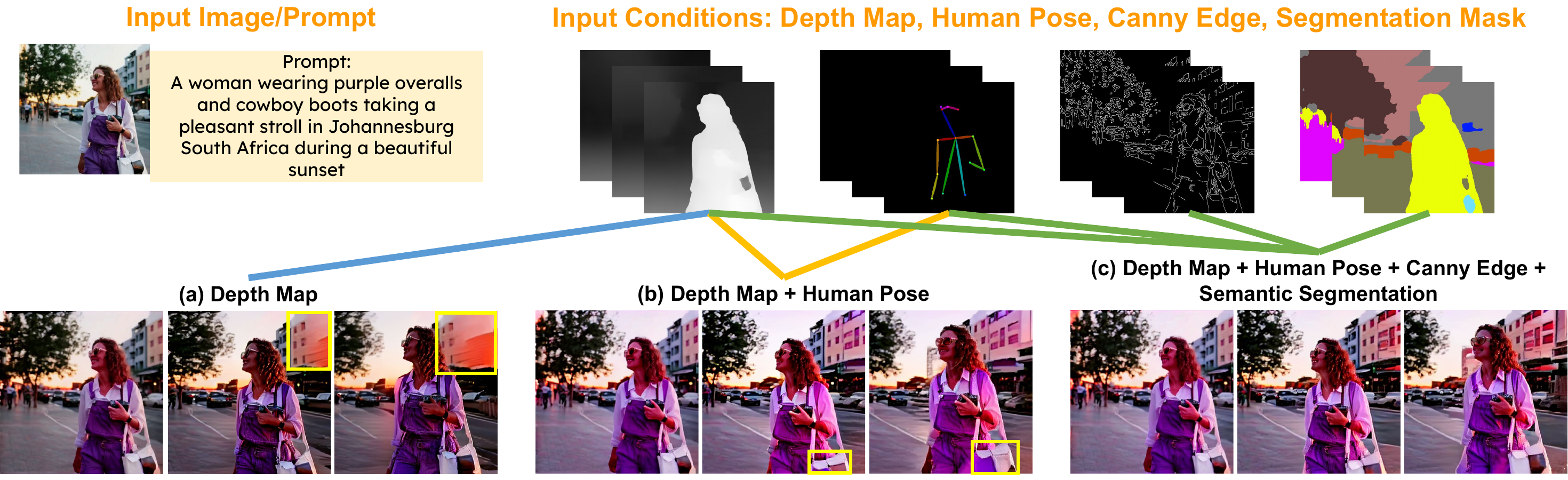}
  \caption{
    Video generation from single and multiple conditions with \adaptermethod{} on I2VGen-XL.
    (a) single condition: depth map;
    (b) 2 conditions: depth map + human pose;
    (c) 4 conditions: depth map + human pose + canny edge + semantic segmentation.
    Adding more conditions can help fix several minor artifacts
    (\eg{}, in (a) -- building is blurred; in (b) -- purse color changes).
  }
\label{fig:multi_source_examples} 
\end{figure}

\cref{fig:multi_source_examples} shows example videos generated with single and multiple conditions.
While all videos correctly capture the high-level dynamics of
`a woman wearing purple strolling during sunset',
the videos generated with more conditions show more robustness in several minor artifacts.
For example,
when only depth map is given (\cref{fig:multi_source_examples} a),
the building behind the person is blurred.
When depth map and human pose are given (\cref{fig:multi_source_examples} b),
the color of the purse changes from white to purple.
When four conditions (depth map, canny edge, human pose, and semantic segmentation) are given, such artifacts are removed (\cref{fig:multi_source_examples} c).

\begin{figure}[h]
  \centering
  \includegraphics[width=.89\linewidth]{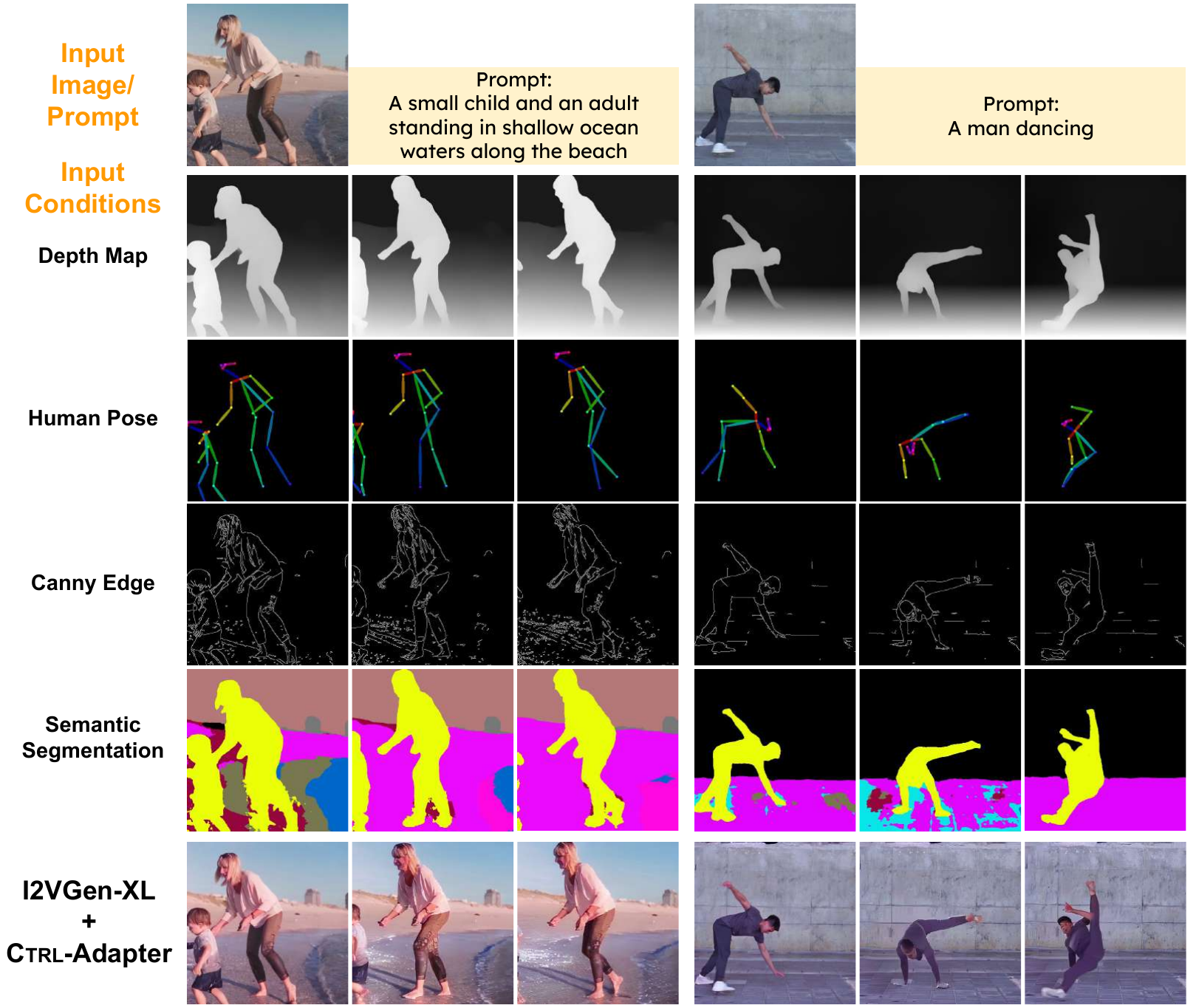}
  \caption{
    Video generated with {\color{Blue}\textbf{I2VGen-XL + \adaptermethod{}}}
 from 4 control conditions: depth map + human pose + canny edge + semantic segmentation map. 
  }
\label{fig:multi_source_examples_2} 
\end{figure}

In \cref{fig:multi_source_examples_2}, we show multi-condition control examples with I2VGen-XL.

\clearpage
\subsection{Image Generation Visualization Examples}
\label{appendix_subsec:image_gen_examples}

\begin{figure}[h]
  \centering
  \includegraphics[width=.9\linewidth]{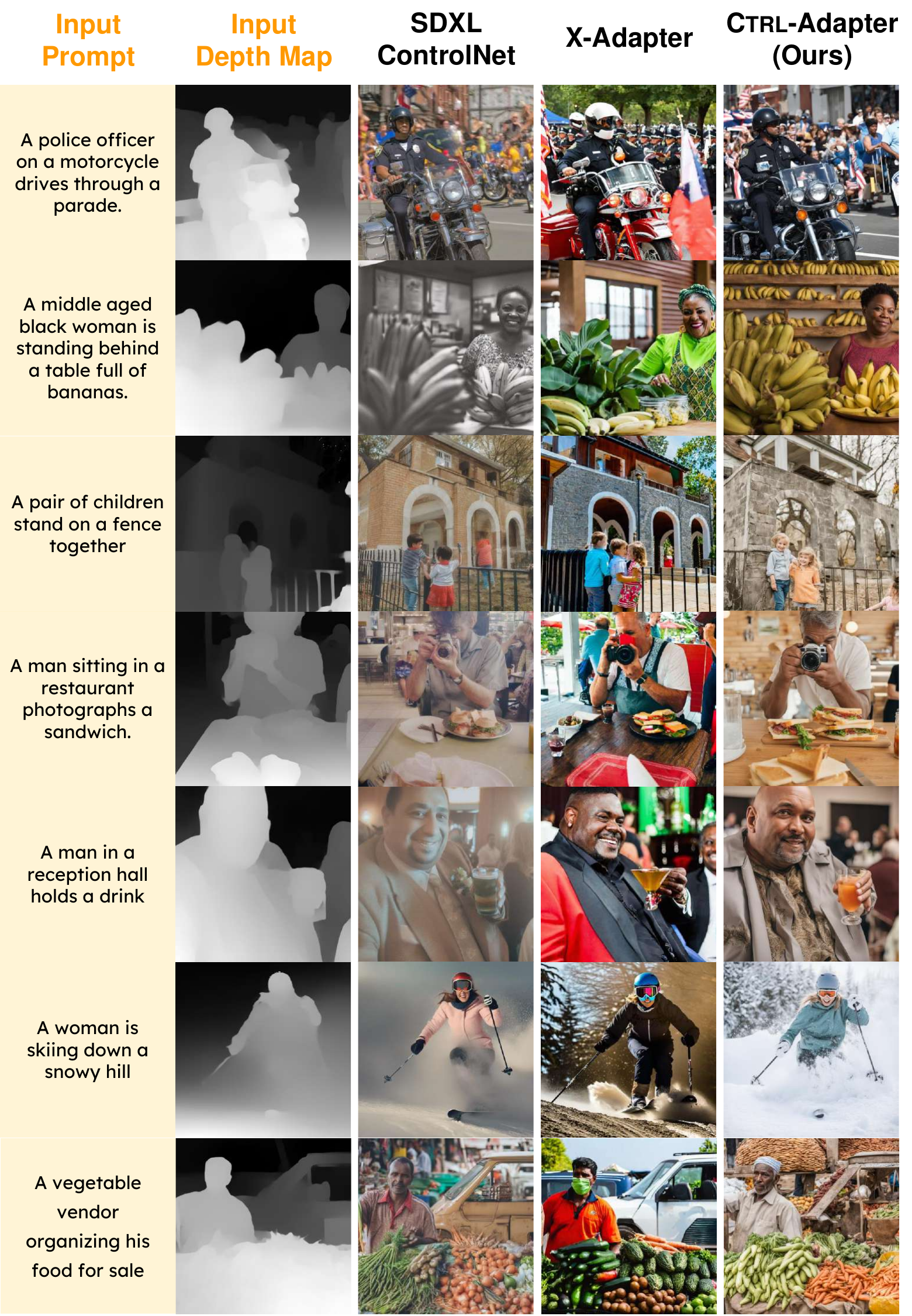}
  \caption{Image generation from different {\color{Blue}\textbf{SDXL}}-based image control methods and \adaptermethod{} on COCO val2017 split using {\color{orange}\textbf{depth map}} as control condition.
  }
\label{fig:appendix_sdxl_depth_visualization} 
\end{figure}

\begin{figure}[h]
  \centering
  \includegraphics[width=.9\linewidth]{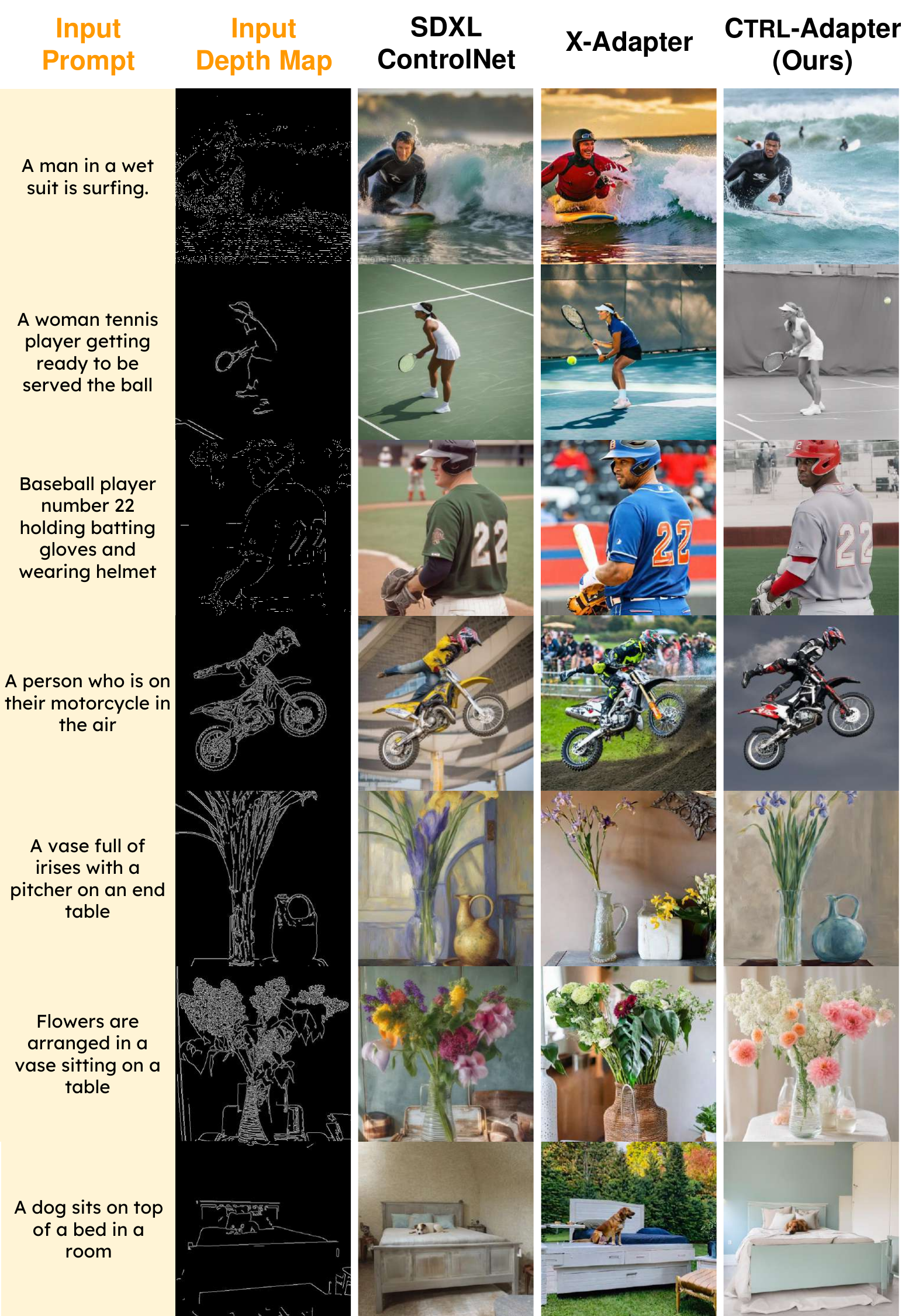}
  \caption{Image generation from different {\color{Blue}\textbf{SDXL}}-based image control methods and \adaptermethod{} on COCO val2017 split using {\color{orange}\textbf{canny edge}} as control condition.
  }
\label{fig:appendix_sdxl_canny_visualization} 
\end{figure}

In \cref{fig:appendix_sdxl_depth_visualization}
and
\cref{fig:appendix_sdxl_canny_visualization},
we show image generation results on COCO val2017 split using depth map and canny edge as control conditions.

\begin{figure}[h]
  \centering
  \includegraphics[width=.8\linewidth]{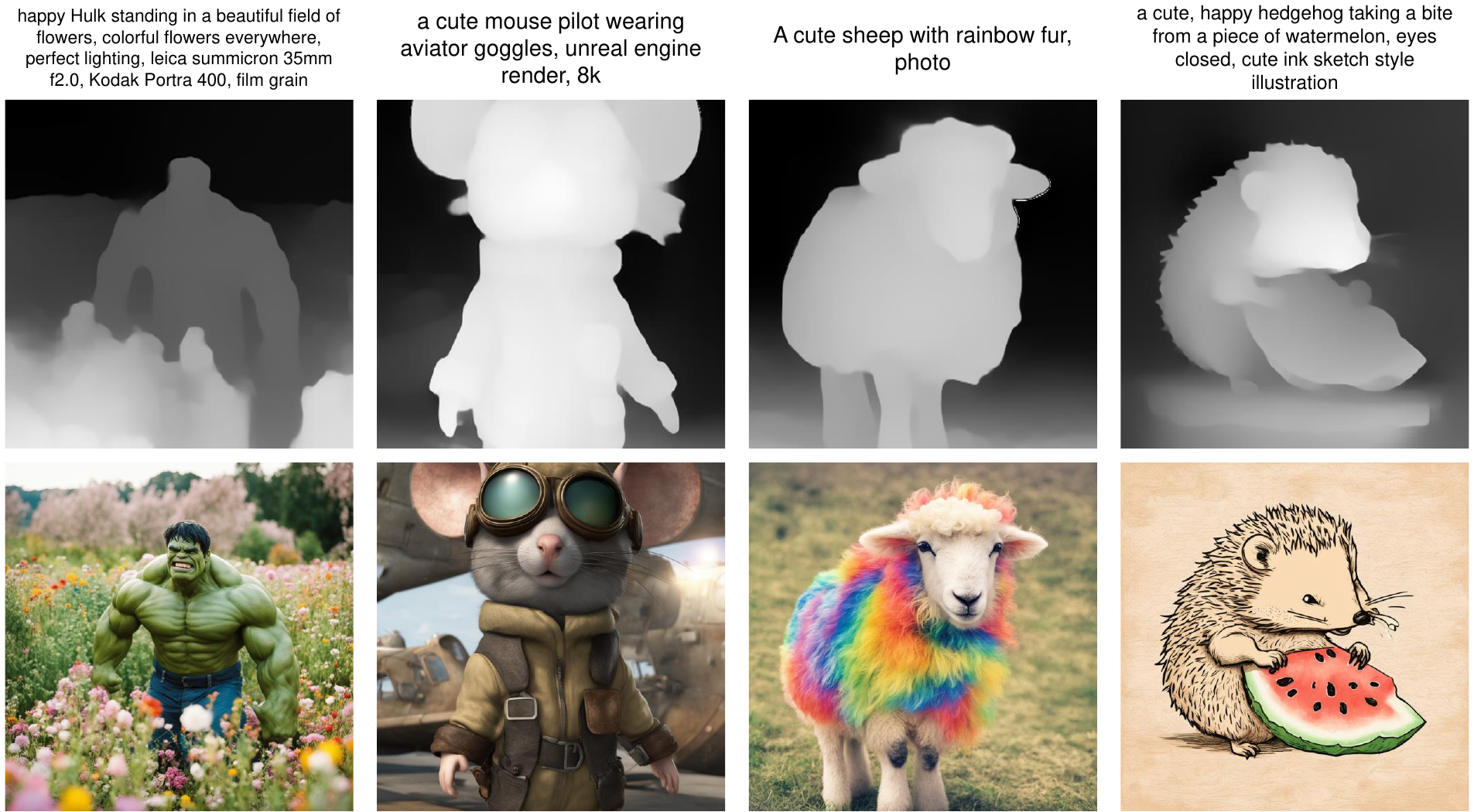}
  \caption{Image generation
   with {\color{Blue} \textbf{SDXL + \adaptermethod{}}} using {\color{orange} \textbf{depth map}} as a control condition.
  }
\label{fig:image_control_depth} 
\end{figure}

\begin{figure}[h]
  \centering
  \includegraphics[width=.8\linewidth]{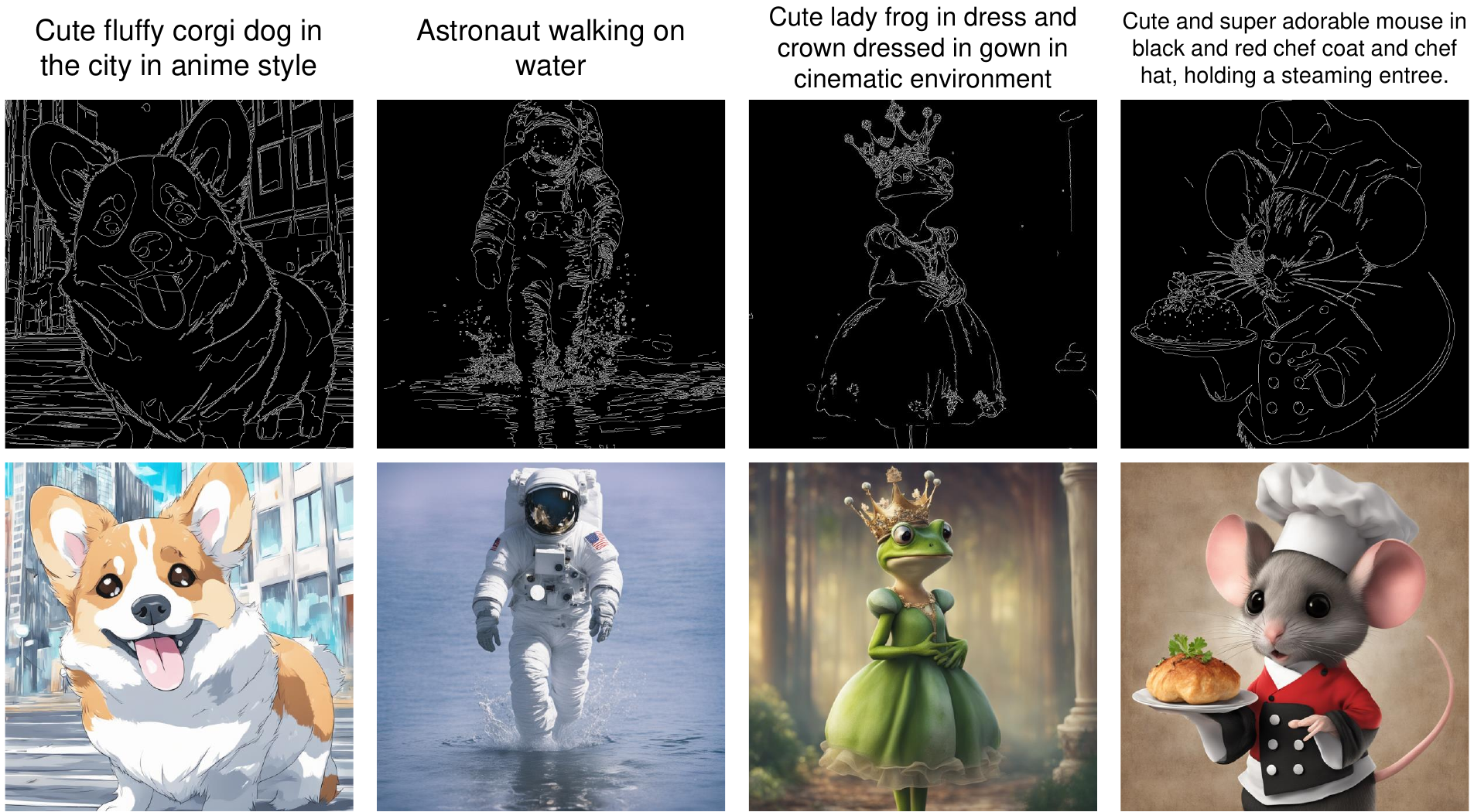}
  \caption{Image generation
   with {\color{Blue} \textbf{SDXL + \adaptermethod{}}} using {\color{orange} \textbf{canny edge}} as a control condition.
  }
\label{fig:image_control_canny} 
\end{figure}

\begin{figure}[h]
  \centering
  \includegraphics[width=.8\linewidth]{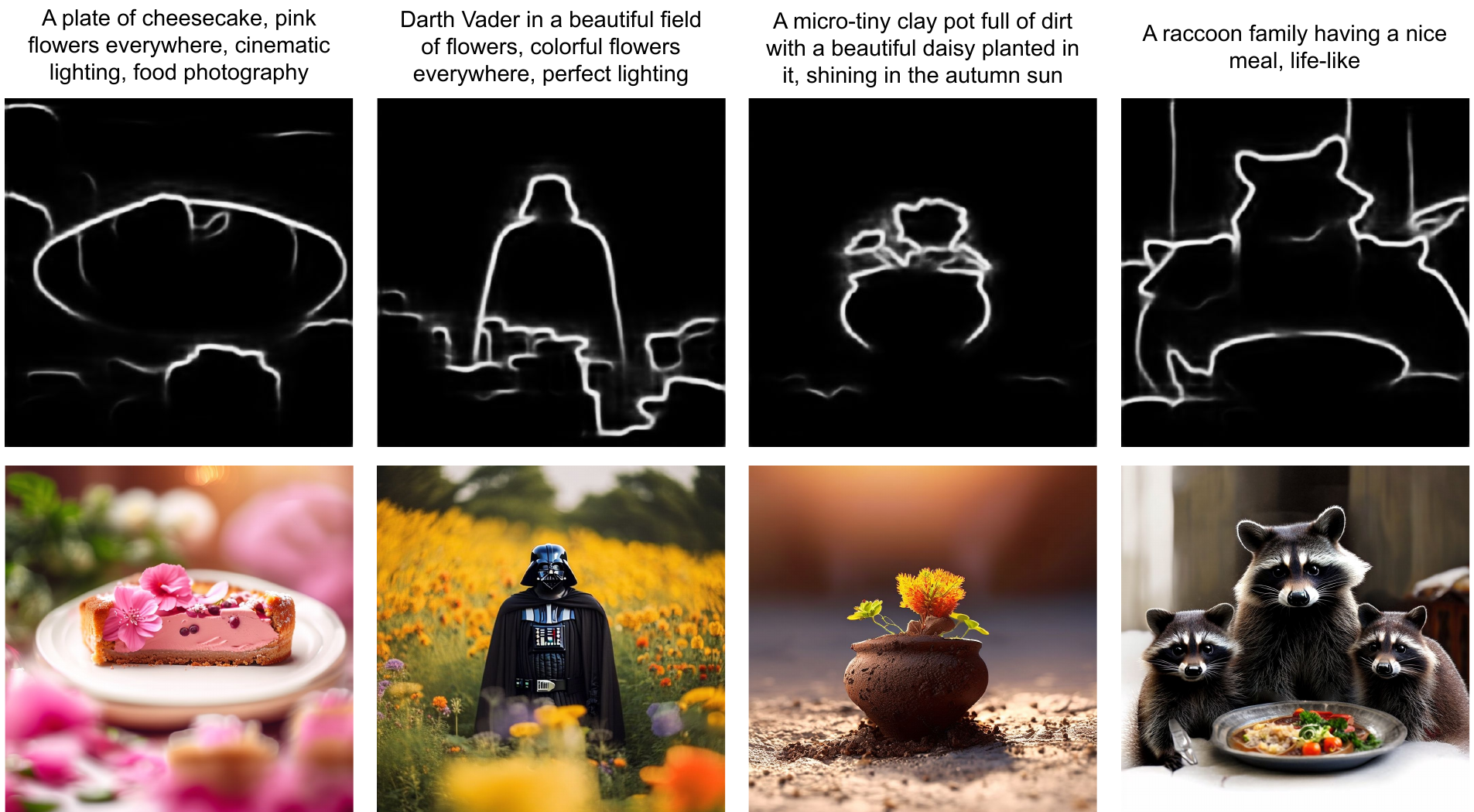}
  \caption{Image generation
   with {\color{Blue} \textbf{PixArt-$\alpha$ + \adaptermethod{}}} using {\color{orange} \textbf{soft edge}} as a control condition.
  }
\label{fig:image_control_softedge} 
\end{figure}

In
\cref{fig:image_control_depth},
\cref{fig:image_control_canny}, and \cref{fig:image_control_softedge},
we show image generation results on
prompts from Lexica\footnote{\url{https://lexica.art/}} 
using depth map, canny edge, and soft edge as control conditions.

\clearpage
\subsection{Visualization Examples for Additional Downstream Tasks}
\label{appendix_subsec:down_stream_task_examples}

Here, we describe in detail how our \adaptermethod{} can be seamlessly integrated into a wide variety of downstream tasks including video editing, video style transfer, and text-guided motion control, as mentioned in \cref{sec:beyond_spatial_control}. Additional visualizations are also shown in this part.

\paragraph{Video editing.}

Video editing can be achieved by combining image and video \adaptermethod{}s. The procedure is as follows: 
\begin{itemize}[leftmargin=1.5em]
    \item Firstly, given a source video, we first extract the control condition(s). We can either extract a single control condition (\eg{}, depth map), or multiple control conditions (\eg{}, depth map, canny edge, segmentation, \etc{}) to enhance performance (as we observe in \cref{appendix_subsec:multi_video_gen_examples} that multi-condition control improves spatial control accuracy) .
    \item Next, given a user-provided prompt together with the extracted control condition(s), we can use image \adaptermethod{} (\ie{}, SDXL + \adaptermethod{}) to generate the first frame of the video.
    \item Finally, we can use video \adaptermethod{} (\ie{}, I2VGen-XL + \adaptermethod{}), with the generated first frame image, text prompt, and extracted control conditions as inputs for final video generation.
\end{itemize}

\begin{figure}[h]
  \centering
  \includegraphics[width=.99\linewidth]{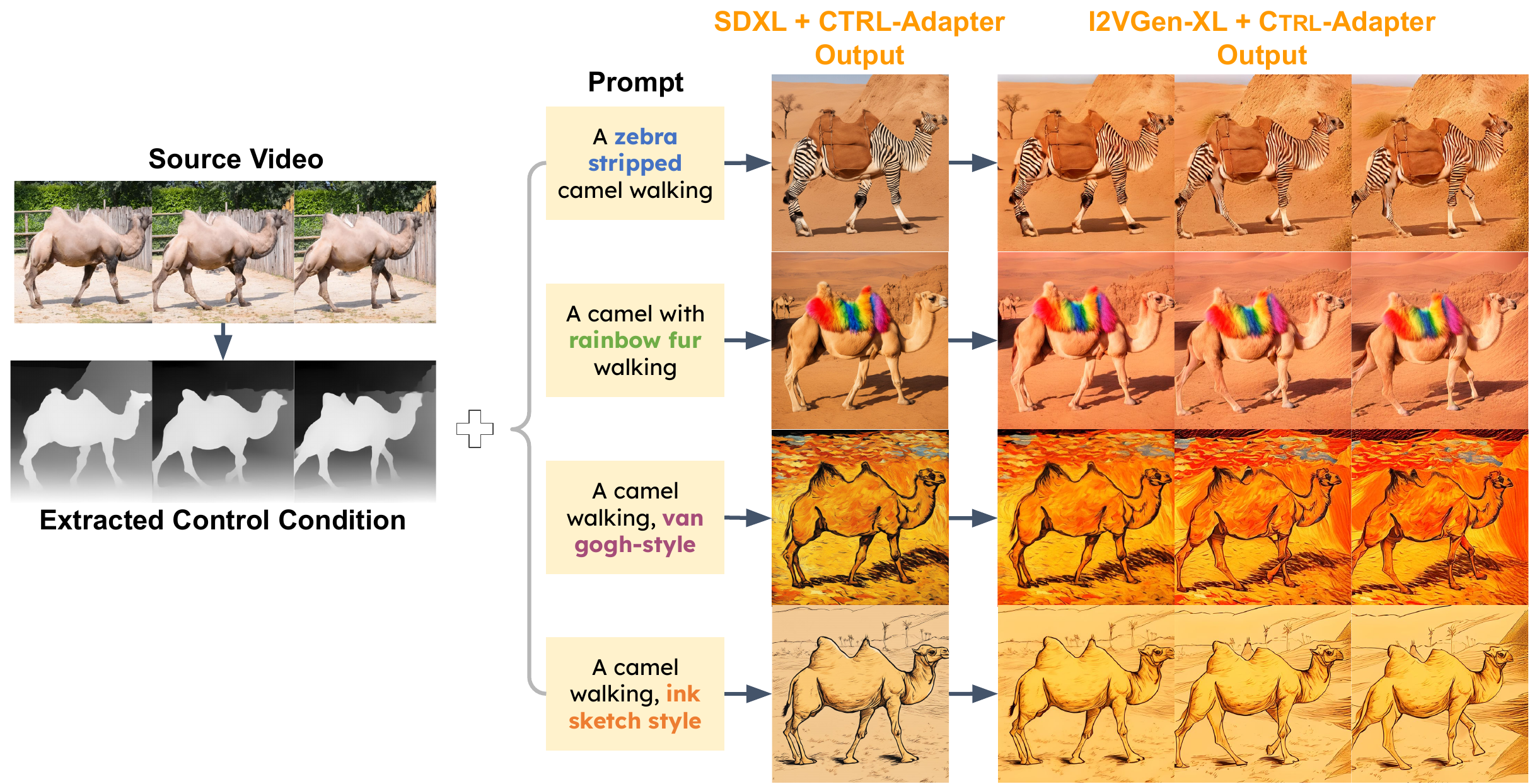}
  \caption{
  Video editing by combining {\color{Blue} \textbf{SDXL}} and {\color{Blue} \textbf{I2VGen-XL}}, where both models are equipped with spatial control via \adaptermethod{}.
  First, we extract conditions (\eg{}, depth map) from the original video.
  Next, we create the initial frame with SDXL + \adaptermethod{}.
  Lastly, we provide the newly generated initial frame and frame-wise conditions to I2VGen-XL + \adaptermethod{} to obtain the final edited video. This video editing framework can edit both object and background.
  }
\label{fig:video_editing_full} 
\end{figure}

In \cref{fig:video_editing_full}, we provide additional visualizations of the camel example in our main paper.

\clearpage
\paragraph{Text-Guided Motion Control.} This task can be achieved by combining video \adaptermethod{} with inpainting \cnet{}. We train such \adaptermethod{} as follows:
\begin{itemize}[leftmargin=1.5em]
    \item Firstly, for each training video, we randomly select a random block in the image, with the width and height of the block uniformly sampled from $0.25$ to $0.75$ of the image size.
    \item Next, we color the block area of the video frames as black color (these processed frames can be regarded as control condition sequences like depth maps). 
    \item Finally, we can train \adaptermethod{} with the frozen inpainting \cnet{} similar as other types of \adaptermethod{}s mentioned in our main paper.
\end{itemize}

\begin{figure}[h]
  \centering
  \includegraphics[width=.99\linewidth]{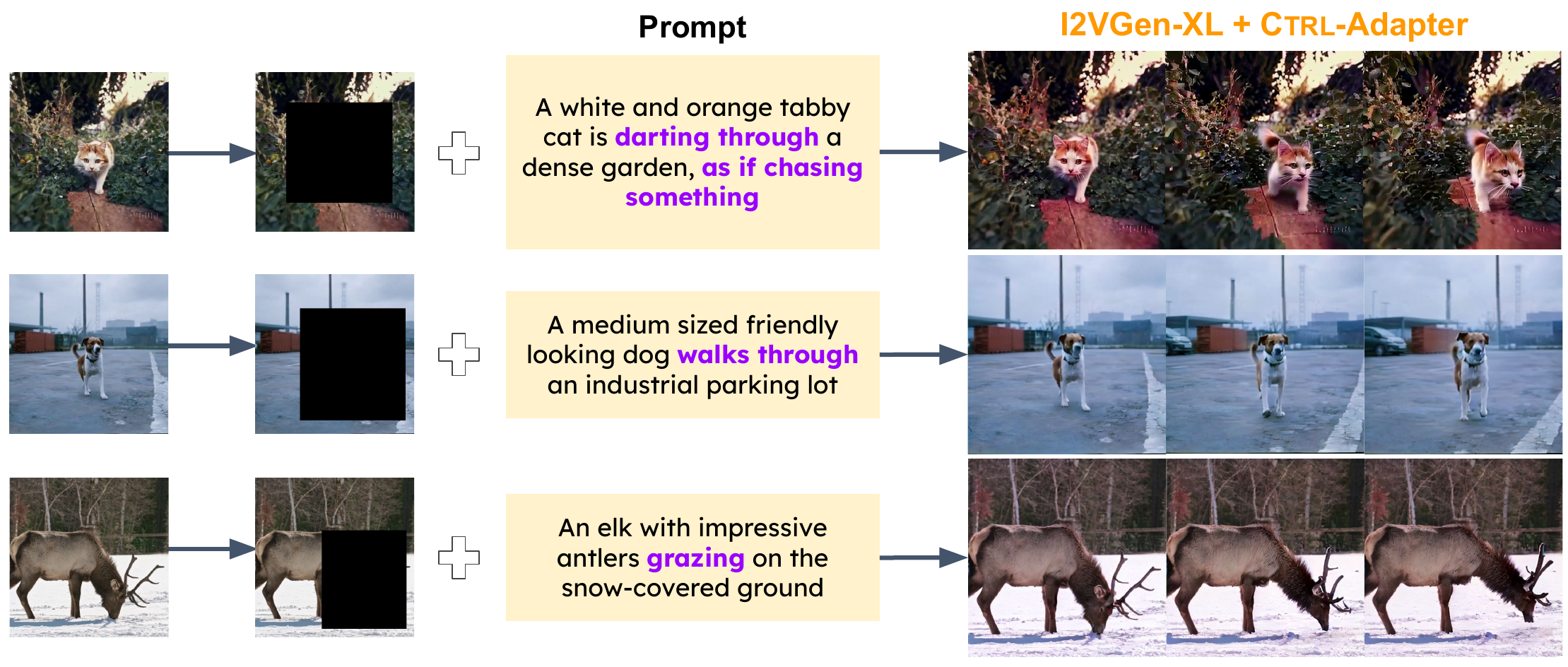}
  \caption{
  Text-guided motion control by combining {\color{Blue} \textbf{inpainting \cnet{}}} with {\color{orange} \textbf{I2VGen-XL + \adaptermethod{}}}. Specifically, inpainting \cnet{} takes the masked frames as well as text prompt as inputs. The output feature maps of inpainting \cnet{} are then given to \adaptermethod{} built on top of I2VGen-XL to generate the final video. Object(s) in the masked area can follow the motion described in the text prompt. The unmasked area can be either static or dynamic.
  }
\label{fig:text_guided_image_animation} 
\end{figure}

In \cref{fig:text_guided_image_animation}, we provide additional visualizations of text-guided image amination.

\begin{figure}[h]
  \centering
  \includegraphics[width=.99\linewidth]{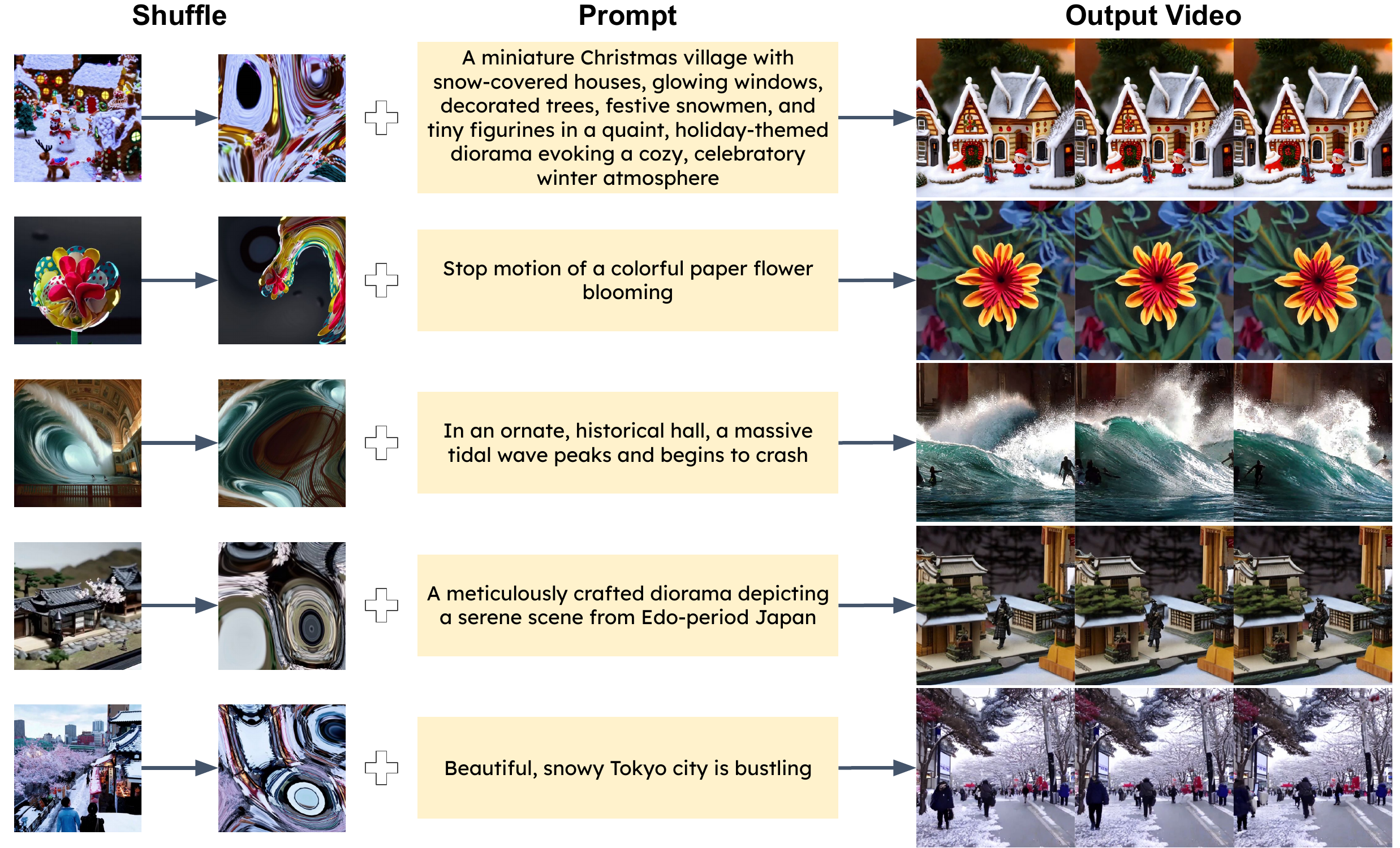}
  \caption{
  Video style transfer by combining {\color{Blue} \textbf{shuffle \cnet{}}} with {\color{orange} \textbf{Latte + \adaptermethod{}}}. Specifically, shuffle \cnet{} takes the shuffled image as well as text prompt as inputs. The output feature maps of shuffle \cnet{} are then given to \adaptermethod{} built on top of Latte to generate the final video. The generated videos maintain similar style as the input image before shuffling.
  }
\label{fig:video_style_transfer} 
\end{figure}

\clearpage
\paragraph{Video style transfer.} This task can be achieved by combining video \adaptermethod{} with shuffle \cnet{}. We train such \adaptermethod{} as follows:
\begin{itemize}[leftmargin=1.5em]
    \item Firstly, for the first frame of each training video, we apply the content shuffle detector implemented in the \texttt{controlnet\_aux}\footnote{\url{https://github.com/huggingface/controlnet_aux}} library, to get a shuffled image. 
    \item Next, we repeat this shuffled image $N$ times, with $N$ equal to the number of output frames of the backbone video diffusion model. These repeated images can be regarded as control condition sequences like depth maps.
    \item Finally, we can train \adaptermethod{} with the frozen shuffle \cnet{} similar as other types of \adaptermethod{}s mentioned in our main paper.
\end{itemize}

in \cref{fig:video_style_transfer}, we provide additional visualizations of video style transfer.

\clearpage
\section{Broader Impacts}
\label{sec:broader_impact}

\adaptermethod{} is motivated by the fact that training \cnet{}s for new diffusion models, especially video diffusion models that need to consider temporal consistency, can be a huge burden for many users. As shown in \cref{fig:training_statistics_comparison}, by adopting pretrained \cnet{}s, training \adaptermethod{} can be significantly faster than training other controllable image or video generation methods. For example, with the same type of compute (\ie{}, A100 80GB GPUs), \adaptermethod{} trained on SDXL depth \cnet{} for 10 GPU hours can outperform SDXL \cnet{} trained for 700 GPU hours. This drastically reduces the carbon emissions footprint by over 70 times.
Therefore, we believe that our work can be a strong contribution to efficient and controllable image and video generation.

While our framework can benefit numerous applications in controllable generation, similar to other image and video generation frameworks, it can also be used for potentially harmful purposes (e.g., creating false information or misleading videos). Therefore, it should be used with caution in real-world applications.

\section{Limitations}
\label{sec:limitations}

Note that since \adaptermethod{} is a method to equip current open-source image and video diffusion models with better control, its performance, quality, and potential visual artifacts largely depend on the capabilities (e.g., motion styles and video length) of the backbone models used.
For example, if a diffusion model cannot handle complex motions, \adaptermethod{} built on top of this backbone might lead to non-optimal complex motion control. 

\section{License}
\label{sec:license}

We use standard licenses from the community
and provide the following links to the licenses for the datasets, codes, and models that we used in this
paper. For further information, please refer to the specific link.

PyTorch~\cite{Ansel_PyTorch_2_Faster_2024}: \href{https://github.com/pytorch/pytorch/blob/main/LICENSE}{BSD-style}

HuggingFace Transformers~\cite{wolf-etal-2020-transformers}: \href{https://github.com/huggingface/transformers/blob/main/LICENSE}{Apache License 2.0}

HuggingFace Diffusers~\cite{von-platen-etal-2022-diffusers}: \href{https://github.com/huggingface/diffusers/blob/main/LICENSE}{Apache License 2.0}

\cnet{}~\cite{zhang2023adding}: \href{https://github.com/lllyasviel/ControlNet?tab=Apache-2.0-1-ov-file}{Apache License 2.0}

SDXL~\cite{podell2023sdxl}: \href{https://github.com/Stability-AI/generative-models?tab=MIT-1-ov-file}{MiT License}

PixArt-$\alpha$~\cite{chen2024pixartalpha}: \href{https://github.com/PixArt-alpha/PixArt-alpha?tab=AGPL-3.0-1-ov-file}{AGPL-3.0 License}

I2VGen-XL~\cite{zhang2023i2vgen}: \href{https://huggingface.co/ali-vilab/i2vgen-xl}{MiT License}

Stable Video Diffusion (SVD)~\cite{blattmann2023stable}: \href{https://github.com/Stability-AI/generative-models?tab=MIT-1-ov-file}{MiT License}

Latte~\cite{ma2024latte}: \href{https://github.com/Vchitect/Latte/blob/main/LICENSE}{Apache License 2.0}

Hotshot-XL~\cite{Mullan_Hotshot-XL_2023}: \href{https://github.com/hotshotco/Hotshot-XL/blob/main/LICENSE}{Apache License 2.0}

LAION dataset~\cite{LAION_POP}: \href{https://github.com/LAION-AI/laion-datasets/blob/main/LICENSE}{MiT License}

Panda70M dataset~\cite{chen2024panda70m}: \href{https://github.com/snap-research/Panda-70M/blob/main/LICENSE}{License}

COCO dataset~\cite{lin2014microsoft}: \href{https://cocodataset.org/#termsofuse}{CC BY 4.0}

DAVIS 2017 dataset~\cite{pont20172017}: \href{https://davischallenge.org/challenge2017/rulesdates.html}{CC BY 4.0}

%% file: timestep_pseudocode.tex
\begin{center}
\begin{minipage}{\textwidth}
\begin{algorithm}[H]
\begin{minted}[linenos,fontsize=\footnotesize]{python}
import torch 

def sample_sigma(u, loc=0., scale=1.):
    """Draw a noise scale (sigma) from the noise schedule of Karras et al. (2022)"""
    sigma_min, sigma_max, rho = 0.002, 700, 7 # values used in the paper
    min_inv_rho, max_inv_rho = sigma_min ** (1 / rho), sigma_max ** (1 / rho)
    sigma = (max_inv_rho + (1-u) * (min_inv_rho - max_inv_rho)) ** rho
    return sigma

def sigma_to_timestep(sigma):
    """Map noise scale to timestep. Here we use the function used in SVD."""
    timestep = 0.25 * sigma.log()
    return timestep

def inverse_timestamp_sample():
    """Sample noise scales and timesteps for ControlNet and diffusion models
    trained with continuous noise sampler. Here we use the setting used for SVD."""
    # 1) sample u from Uniform[0,1]
    u = torch.rand(1)
    # 2) calculate sigma_svd from pre-defined log-normal distribution
    sigma_svd = sample_sigma(u, loc=0.7, scale=1.6)
    # 3) calculate timestep_svd from sigma_svd via pre-defined mapping function
    timestep_svd = sigma_to_timestep(sigma_svd)
    # 4) calculate timestep and sigma for controlnet
    sigma_cnet, timestep_cnet = u, round(1000 * u)
    return sigma_svd, timestep_svd, sigma_cnet, timestep_cnet
\end{minted}
\caption{PyTorch Implementation for Inverse Timestep Sampling
}
\label{alg:timestep_sampling}
\end{algorithm}
\end{minipage}
\end{center}

%% file: main.bbl
\begin{thebibliography}{10}

\bibitem{Ansel_PyTorch_2_Faster_2024}
J.~Ansel, E.~Yang, H.~He, N.~Gimelshein, A.~Jain, M.~Voznesensky, B.~Bao, P.~Bell, D.~Berard, E.~Burovski, G.~Chauhan, A.~Chourdia, W.~Constable, A.~Desmaison, Z.~DeVito, E.~Ellison, W.~Feng, J.~Gong, M.~Gschwind, B.~Hirsh, S.~Huang, K.~Kalambarkar, L.~Kirsch, M.~Lazos, M.~Lezcano, Y.~Liang, J.~Liang, Y.~Lu, C.~Luk, B.~Maher, Y.~Pan, C.~Puhrsch, M.~Reso, M.~Saroufim, M.~Y. Siraichi, H.~Suk, M.~Suo, P.~Tillet, E.~Wang, X.~Wang, W.~Wen, S.~Zhang, X.~Zhao, K.~Zhou, R.~Zou, A.~Mathews, G.~Chanan, P.~Wu, and S.~Chintala.
\newblock {PyTorch 2: Faster Machine Learning Through Dynamic Python Bytecode Transformation and Graph Compilation}.
\newblock In {\em 29th ACM International Conference on Architectural Support for Programming Languages and Operating Systems, Volume 2 (ASPLOS '24)}. ACM, Apr. 2024.

\bibitem{Avrahami2023SpaText}
O.~Avrahami, T.~Hayes, O.~Gafni, S.~Gupta, Y.~Taigman, D.~Parikh, D.~Lischinski, O.~Fried, and X.~Yin.
\newblock {SpaText: Spatio-Textual Representation for Controllable Image Generation}.
\newblock In {\em CVPR}, nov 2023.

\bibitem{blattmann2023stable}
A.~Blattmann, T.~Dockhorn, S.~Kulal, D.~Mendelevitch, M.~Kilian, D.~Lorenz, Y.~Levi, Z.~English, V.~Voleti, A.~Letts, et~al.
\newblock Stable video diffusion: Scaling latent video diffusion models to large datasets.
\newblock {\em arXiv preprint arXiv:2311.15127}, 2023.

\bibitem{opencv_library}
G.~Bradski.
\newblock {The OpenCV Library}.
\newblock {\em Dr. Dobb's Journal of Software Tools}, 2000.

\bibitem{cao2017realtime}
Z.~Cao, T.~Simon, S.-E. Wei, and Y.~Sheikh.
\newblock Realtime multi-person 2d pose estimation using part affinity fields.
\newblock In {\em Proceedings of the IEEE conference on computer vision and pattern recognition}, pages 7291--7299, 2017.

\bibitem{chen2024videocrafter2}
H.~Chen, Y.~Zhang, X.~Cun, M.~Xia, X.~Wang, C.~Weng, and Y.~Shan.
\newblock Videocrafter2: Overcoming data limitations for high-quality video diffusion models, 2024.

\bibitem{chen2024pixart_delta}
J.~Chen, Y.~Wu, S.~Luo, E.~Xie, S.~Paul, P.~Luo, H.~Zhao, and Z.~Li.
\newblock Pixart-$\{$$\backslash$delta$\}$: Fast and controllable image generation with latent consistency models.
\newblock {\em arXiv preprint arXiv:2401.05252}, 2024.

\bibitem{chen2024pixartalpha}
J.~Chen, J.~YU, C.~GE, L.~Yao, E.~Xie, Z.~Wang, J.~Kwok, P.~Luo, H.~Lu, and Z.~Li.
\newblock Pixart-\${\textbackslash}alpha\$: Fast training of diffusion transformer for photorealistic text-to-image synthesis.
\newblock In {\em The Twelfth International Conference on Learning Representations}, 2024.

\bibitem{chen2024panda70m}
T.-S. Chen, A.~Siarohin, W.~Menapace, E.~Deyneka, H.-w. Chao, B.~E. Jeon, Y.~Fang, H.-Y. Lee, J.~Ren, M.-H. Yang, and S.~Tulyakov.
\newblock Panda-70m: Captioning 70m videos with multiple cross-modality teachers.
\newblock In {\em CVPR 2024}, 2024.

\bibitem{chen2023control}
W.~Chen, J.~Wu, P.~Xie, H.~Wu, J.~Li, X.~Xia, X.~Xiao, and L.~Lin.
\newblock Control-a-video: Controllable text-to-video generation with diffusion models.
\newblock {\em arXiv preprint arXiv:2305.13840}, 2023.

\bibitem{dai2023emu}
X.~Dai, J.~Hou, C.-Y. Ma, S.~Tsai, J.~Wang, R.~Wang, P.~Zhang, S.~Vandenhende, X.~Wang, A.~Dubey, et~al.
\newblock Emu: Enhancing image generation models using photogenic needles in a haystack.
\newblock {\em arXiv preprint arXiv:2309.15807}, 2023.

\bibitem{esser2024scaling}
P.~Esser, S.~Kulal, A.~Blattmann, R.~Entezari, J.~Müller, H.~Saini, Y.~Levi, D.~Lorenz, A.~Sauer, F.~Boesel, D.~Podell, T.~Dockhorn, Z.~English, K.~Lacey, A.~Goodwin, Y.~Marek, and R.~Rombach.
\newblock Scaling rectified flow transformers for high-resolution image synthesis, 2024.

\bibitem{Inverse_transform_sampling}
{Estimation lemma}.
\newblock Estimation lemma --- {W}ikipedia{,} the free encyclopedia, 2010.

\bibitem{farneback2003two}
G.~Farneb{\"a}ck.
\newblock Two-frame motion estimation based on polynomial expansion.
\newblock In {\em Image Analysis: 13th Scandinavian Conference, SCIA 2003 Halmstad, Sweden, June 29--July 2, 2003 Proceedings 13}, pages 363--370. Springer, 2003.

\bibitem{Farnebck2003TwoFrameME}
G.~Farneb{\"a}ck.
\newblock Two-frame motion estimation based on polynomial expansion.
\newblock In {\em Scandinavian Conference on Image Analysis}, 2003.

\bibitem{Gafni2022Make-A-Scene}
O.~Gafni, A.~Polyak, O.~Ashual, S.~Sheynin, D.~Parikh, and Y.~Taigman.
\newblock {Make-A-Scene: Scene-Based Text-to-Image Generation with Human Priors}.
\newblock In {\em ECCV}, 2022.

\bibitem{Gal2023TextualInversion}
R.~Gal, Y.~Alaluf, Y.~Atzmon, O.~Patashnik, A.~H. Bermano, G.~Chechik, and D.~Cohen-Or.
\newblock {An Image is Worth One Word: Personalizing Text-to-Image Generation using Textual Inversion}.
\newblock In {\em ICLR}, 2023.

\bibitem{girdhar2023emu}
R.~Girdhar, M.~Singh, A.~Brown, Q.~Duval, S.~Azadi, S.~S. Rambhatla, A.~Shah, X.~Yin, D.~Parikh, and I.~Misra.
\newblock Emu video: Factorizing text-to-video generation by explicit image conditioning.
\newblock {\em arXiv preprint arXiv:2311.10709}, 2023.

\bibitem{goodfellow2020generative}
I.~Goodfellow, J.~Pouget-Abadie, M.~Mirza, B.~Xu, D.~Warde-Farley, S.~Ozair, A.~Courville, and Y.~Bengio.
\newblock Generative adversarial networks.
\newblock {\em Communications of the ACM}, 63(11):139--144, 2020.

\bibitem{guo2024dit}
Q.~Guo and D.~Yue.
\newblock Dit-visualization.
\newblock \url{https://github.com/guoqincode/DiT-Visualization}, 2024.
\newblock Exploring the differences between DiT-based and Unet-based diffusion models in feature aspects using code from diffusers, Plug-and-Play, and PixArt.

\bibitem{guo2023sparsectrl}
Y.~Guo, C.~Yang, A.~Rao, M.~Agrawala, D.~Lin, and B.~Dai.
\newblock Sparsectrl: Adding sparse controls to text-to-video diffusion models.
\newblock {\em arXiv preprint arXiv:2311.16933}, 2023.

\bibitem{guo2023animatediff}
Y.~Guo, C.~Yang, A.~Rao, Y.~Wang, Y.~Qiao, D.~Lin, and B.~Dai.
\newblock Animatediff: Animate your personalized text-to-image diffusion models without specific tuning.
\newblock In {\em International Conference on Learning Representations}, 2024.

\bibitem{gupta2023photorealistic}
A.~Gupta, L.~Yu, K.~Sohn, X.~Gu, M.~Hahn, L.~Fei-Fei, I.~Essa, L.~Jiang, and J.~Lezama.
\newblock Photorealistic video generation with diffusion models, 2023.

\bibitem{resnet2016}
K.~He, X.~Zhang, S.~Ren, and J.~Sun.
\newblock Deep residual learning for image recognition.
\newblock In {\em 2016 IEEE Conference on Computer Vision and Pattern Recognition (CVPR)}, pages 770--778, 2016.

\bibitem{he2022lvdm}
Y.~He, T.~Yang, Y.~Zhang, Y.~Shan, and Q.~Chen.
\newblock Latent video diffusion models for high-fidelity long video generation, 2022.

\bibitem{heusel2017gans}
M.~Heusel, H.~Ramsauer, T.~Unterthiner, B.~Nessler, and S.~Hochreiter.
\newblock Gans trained by a two time-scale update rule converge to a local nash equilibrium.
\newblock {\em Advances in neural information processing systems}, 30, 2017.

\bibitem{ho2022imagen-video}
J.~Ho, W.~Chan, C.~Saharia, J.~Whang, R.~Gao, A.~Gritsenko, D.~P. Kingma, B.~Poole, M.~Norouzi, D.~J. Fleet, et~al.
\newblock Imagen video: High definition video generation with diffusion models.
\newblock {\em arXiv preprint arXiv:2210.02303}, 2022.

\bibitem{ho2020denoising}
J.~Ho, A.~Jain, and P.~Abbeel.
\newblock Denoising diffusion probabilistic models.
\newblock {\em Advances in neural information processing systems}, 33:6840--6851, 2020.

\bibitem{hoogeboom2023simple}
E.~Hoogeboom, J.~Heek, and T.~Salimans.
\newblock simple diffusion: End-to-end diffusion for high resolution images.
\newblock In {\em International Conference on Machine Learning}, pages 13213--13232. PMLR, 2023.

\bibitem{Houlsby2019ParameterEfficientTL}
N.~Houlsby, A.~Giurgiu, S.~Jastrzebski, B.~Morrone, Q.~de~Laroussilhe, A.~Gesmundo, M.~Attariyan, and S.~Gelly.
\newblock Parameter-efficient transfer learning for nlp.
\newblock In {\em ICML}, volume abs/1902.00751, 2019.

\bibitem{hu2023videocontrolnet}
Z.~Hu and D.~Xu.
\newblock Videocontrolnet: A motion-guided video-to-video translation framework by using diffusion model with controlnet.
\newblock {\em arXiv preprint arXiv:2307.14073}, 2023.

\bibitem{khachatryan2023text2video}
L.~Khachatryan, A.~Movsisyan, V.~Tadevosyan, R.~Henschel, Z.~Wang, S.~Navasardyan, and H.~Shi.
\newblock Text2video-zero: Text-to-image diffusion models are zero-shot video generators.
\newblock In {\em ICCV 2023}, 2023.

\bibitem{kim2023diffblender}
S.~Kim, J.~Lee, K.~Hong, D.~Kim, and N.~Ahn.
\newblock Diffblender: Scalable and composable multimodal text-to-image diffusion models.
\newblock {\em arXiv preprint arXiv:2305.15194}, 2023.

\bibitem{kingma2013auto}
D.~P. Kingma and M.~Welling.
\newblock Auto-encoding variational bayes.
\newblock In {\em ICLR}, 2014.

\bibitem{xFormers2022}
B.~Lefaudeux, F.~Massa, D.~Liskovich, W.~Xiong, V.~Caggiano, S.~Naren, M.~Xu, J.~Hu, M.~Tintore, S.~Zhang, P.~Labatut, D.~Haziza, L.~Wehrstedt, J.~Reizenstein, and G.~Sizov.
\newblock xformers: A modular and hackable transformer modelling library.
\newblock \url{https://github.com/facebookresearch/xformers}, 2022.

\bibitem{Li2023BLIP-Diffusion}
D.~Li, J.~Li, and S.~C.~H. Hoi.
\newblock {BLIP-Diffusion: Pre-trained Subject Representation for Controllable Text-to-Image Generation and Editing}.
\newblock In {\em NeurIPS}, 2023.

\bibitem{li2023blip}
J.~Li, D.~Li, S.~Savarese, and S.~Hoi.
\newblock Blip-2: Bootstrapping language-image pre-training with frozen image encoders and large language models.
\newblock In {\em International conference on machine learning}, pages 19730--19742. PMLR, 2023.

\bibitem{li2018storygan}
Y.~Li, Z.~Gan, Y.~Shen, J.~Liu, Y.~Cheng, Y.~Wu, L.~Carin, D.~Carlson, and J.~Gao.
\newblock Storygan: A sequential conditional gan for story visualization.
\newblock In {\em CVPR}, 2019.

\bibitem{li2023gligen}
Y.~Li, H.~Liu, Q.~Wu, F.~Mu, J.~Yang, J.~Gao, C.~Li, and Y.~J. Lee.
\newblock Gligen: Open-set grounded text-to-image generation.
\newblock In {\em Proceedings of the IEEE/CVF Conference on Computer Vision and Pattern Recognition}, pages 22511--22521, 2023.

\bibitem{li2017video}
Y.~Li, M.~R. Min, D.~Shen, D.~Carlson, and L.~Carin.
\newblock Video generation from text.
\newblock In {\em AAAI}, 2017.

\bibitem{lin2023videodirectorgpt}
H.~Lin, A.~Zala, J.~Cho, and M.~Bansal.
\newblock Videodirectorgpt: Consistent multi-scene video generation via llm-guided planning.
\newblock {\em arXiv preprint arXiv:2309.15091}, 2023.

\bibitem{lin2014microsoft}
T.-Y. Lin, M.~Maire, S.~Belongie, J.~Hays, P.~Perona, D.~Ramanan, P.~Doll{\'a}r, and C.~L. Zitnick.
\newblock Microsoft coco: Common objects in context.
\newblock In {\em Computer Vision--ECCV 2014: 13th European Conference, Zurich, Switzerland, September 6-12, 2014, Proceedings, Part V 13}, pages 740--755. Springer, 2014.

\bibitem{long2024videodrafter}
F.~Long, Z.~Qiu, T.~Yao, and T.~Mei.
\newblock Videodrafter: Content-consistent multi-scene video generation with llm.
\newblock {\em arXiv preprint arXiv:2401.01256}, 2024.

\bibitem{loshchilov2018decoupled}
I.~Loshchilov and F.~Hutter.
\newblock Decoupled weight decay regularization.
\newblock In {\em International Conference on Learning Representations}, 2018.

\bibitem{ma2024sit}
N.~Ma, M.~Goldstein, M.~S. Albergo, N.~M. Boffi, E.~Vanden-Eijnden, and S.~Xie.
\newblock Sit: Exploring flow and diffusion-based generative models with scalable interpolant transformers.
\newblock {\em arXiv preprint arXiv:2401.08740}, 2024.

\bibitem{ma2024latte}
X.~Ma, Y.~Wang, G.~Jia, X.~Chen, Z.~Liu, Y.-F. Li, C.~Chen, and Y.~Qiao.
\newblock Latte: Latent diffusion transformer for video generation.
\newblock {\em arXiv preprint arXiv:2401.03048}, 2024.

\bibitem{menapace2024snap}
W.~Menapace, A.~Siarohin, I.~Skorokhodov, E.~Deyneka, T.-S. Chen, A.~Kag, Y.~Fang, A.~Stoliar, E.~Ricci, J.~Ren, and S.~Tulyakov.
\newblock Snap video: Scaled spatiotemporal transformers for text-to-video synthesis, 2024.

\bibitem{mou2023t2i}
C.~Mou, X.~Wang, L.~Xie, Y.~Wu, J.~Zhang, Z.~Qi, Y.~Shan, and X.~Qie.
\newblock T2i-adapter: Learning adapters to dig out more controllable ability for text-to-image diffusion models.
\newblock In {\em AAAI 2024}, 2023.

\bibitem{Mullan_Hotshot-XL_2023}
J.~Mullan, D.~Crawbuck, and A.~Sastry.
\newblock {Hotshot-XL}, Oct. 2023.

\bibitem{sora_2024}
{OpenAI}.
\newblock Video generation models as world simulators, 2024.

\bibitem{parmar2021cleanfid}
G.~Parmar, R.~Zhang, and J.-Y. Zhu.
\newblock On aliased resizing and surprising subtleties in gan evaluation.
\newblock In {\em CVPR}, 2022.

\bibitem{podell2023sdxl}
D.~Podell, Z.~English, K.~Lacey, A.~Blattmann, T.~Dockhorn, J.~M{\"u}ller, J.~Penna, and R.~Rombach.
\newblock Sdxl: Improving latent diffusion models for high-resolution image synthesis.
\newblock In {\em International Conference on Learning Representations}, 2024.

\bibitem{pont20172017}
J.~Pont-Tuset, F.~Perazzi, S.~Caelles, P.~Arbel{\'a}ez, A.~Sorkine-Hornung, and L.~Van~Gool.
\newblock The 2017 davis challenge on video object segmentation.
\newblock {\em arXiv preprint arXiv:1704.00675}, 2017.

\bibitem{qin2023unicontrol}
C.~Qin, S.~Zhang, N.~Yu, Y.~Feng, X.~Yang, Y.~Zhou, H.~Wang, J.~C. Niebles, C.~Xiong, S.~Savarese, et~al.
\newblock Unicontrol: A unified diffusion model for controllable visual generation in the wild.
\newblock In {\em NeurIPS 2023}, 2023.

\bibitem{Ramesh2022UnCLIP}
A.~Ramesh, P.~Dhariwal, A.~Nichol, C.~Chu, and M.~Chen.
\newblock {Hierarchical Text-Conditional Image Generation with CLIP Latents}, 2022.

\bibitem{ran2023x}
L.~Ran, X.~Cun, J.-W. Liu, R.~Zhao, S.~Zijie, X.~Wang, J.~Keppo, and M.~Z. Shou.
\newblock X-adapter: Adding universal compatibility of plugins for upgraded diffusion model.
\newblock In {\em CVPR}, 2024.

\bibitem{ranftl2020towards}
R.~Ranftl, K.~Lasinger, D.~Hafner, K.~Schindler, and V.~Koltun.
\newblock Towards robust monocular depth estimation: Mixing datasets for zero-shot cross-dataset transfer.
\newblock {\em IEEE transactions on pattern analysis and machine intelligence}, 44(3):1623--1637, 2020.

\bibitem{ranjan2017optical}
A.~Ranjan and M.~J. Black.
\newblock Optical flow estimation using a spatial pyramid network.
\newblock In {\em Proceedings of the IEEE conference on computer vision and pattern recognition}, pages 4161--4170, 2017.

\bibitem{Rasley2020DeepSpeedSO}
J.~Rasley, S.~Rajbhandari, O.~Ruwase, and Y.~He.
\newblock Deepspeed: System optimizations enable training deep learning models with over 100 billion parameters.
\newblock {\em Proceedings of the 26th ACM SIGKDD International Conference on Knowledge Discovery \& Data Mining}, 2020.

\bibitem{stablediffusion}
R.~Rombach, A.~Blattmann, D.~Lorenz, P.~Esser, and B.~Ommer.
\newblock High-resolution image synthesis with latent diffusion models.
\newblock {\em 2022 IEEE/CVF Conference on Computer Vision and Pattern Recognition (CVPR)}, pages 10674--10685, 2021.

\bibitem{rombach2022high}
R.~Rombach, A.~Blattmann, D.~Lorenz, P.~Esser, and B.~Ommer.
\newblock High-resolution image synthesis with latent diffusion models.
\newblock In {\em Proceedings of the IEEE/CVF conference on computer vision and pattern recognition}, pages 10684--10695, 2022.

\bibitem{strawberry_svdtemporalcontrolnet_2023}
C.~Rowles.
\newblock {Stable Video Diffusion Temporal Controlnet}.
\newblock \url{https://github.com/CiaraStrawberry/svd-temporal-controlnet}, 2023.

\bibitem{Ruiz2023Dreambooth}
N.~Ruiz, Y.~Li, V.~Jampani, Y.~Pritch, M.~Rubinstein, and K.~Aberman.
\newblock {DreamBooth: Fine Tuning Text-to-Image Diffusion Models for Subject-Driven Generation}.
\newblock In {\em CVPR}, 2023.

\bibitem{simoryu_lora_diffusion}
S.~Ryu.
\newblock Low-rank adaptation for fast text-to-image diffusion fine-tuning, 2022.

\bibitem{Saharia2022Imagen}
C.~Saharia, W.~Chan, S.~Saxena, L.~Li, J.~Whang, E.~Denton, S.~K.~S. Ghasemipour, B.~K. Ayan, S.~S. Mahdavi, R.~G. Lopes, T.~Salimans, J.~Ho, D.~J. Fleet, and M.~Norouzi.
\newblock {Photorealistic Text-to-Image Diffusion Models with Deep Language Understanding}.
\newblock In {\em NeurIPS}, 2022.

\bibitem{schuhmann2022laion}
C.~Schuhmann, R.~Beaumont, R.~Vencu, C.~Gordon, R.~Wightman, M.~Cherti, T.~Coombes, A.~Katta, C.~Mullis, M.~Wortsman, et~al.
\newblock Laion-5b: An open large-scale dataset for training next generation image-text models.
\newblock {\em Advances in Neural Information Processing Systems}, 35:25278--25294, 2022.

\bibitem{LAION_POP}
C.~Schuhmann and P.~Bevan.
\newblock Laion pop: 600,000 high-resolution images with detailed descriptions.
\newblock \url{https://huggingface.co/datasets/laion/laion-pop}, 2023.

\bibitem{Shazeer2017MoE}
N.~Shazeer, A.~Mirhoseini, K.~Maziarz, A.~Davis, Q.~Le, G.~Hinton, and J.~Dean.
\newblock {Outrageously Large Neural Networks: the Sparsely-Gated Mixture-of-Experts Layer}.
\newblock In {\em ICLR}, 2017.

\bibitem{singer2022make}
U.~Singer, A.~Polyak, T.~Hayes, X.~Yin, J.~An, S.~Zhang, Q.~Hu, H.~Yang, O.~Ashual, O.~Gafni, et~al.
\newblock Make-a-video: Text-to-video generation without text-video data.
\newblock In {\em ICLR}, 2023.

\bibitem{stylegan_v}
I.~Skorokhodov, S.~Tulyakov, and M.~Elhoseiny.
\newblock Stylegan-v: A continuous video generator with the price, image quality and perks of stylegan2.
\newblock In {\em CVPR}, 2022.

\bibitem{Sohl-Dickstein2015}
J.~Sohl-Dickstein, E.~A. Weiss, N.~Maheswaranathan, and S.~Ganguli.
\newblock {Deep unsupervised learning using nonequilibrium thermodynamics}.
\newblock In {\em ICML}, 2015.

\bibitem{tumanyan2023plug}
N.~Tumanyan, M.~Geyer, S.~Bagon, and T.~Dekel.
\newblock Plug-and-play diffusion features for text-driven image-to-image translation.
\newblock In {\em Proceedings of the IEEE/CVF Conference on Computer Vision and Pattern Recognition}, pages 1921--1930, 2023.

\bibitem{von-platen-etal-2022-diffusers}
P.~von Platen, S.~Patil, A.~Lozhkov, P.~Cuenca, N.~Lambert, K.~Rasul, M.~Davaadorj, and T.~Wolf.
\newblock Diffusers: State-of-the-art diffusion models.
\newblock \url{https://github.com/huggingface/diffusers}, 2022.

\bibitem{wang2023gen}
F.-Y. Wang, W.~Chen, G.~Song, H.-J. Ye, Y.~Liu, and H.~Li.
\newblock Gen-l-video: Multi-text to long video generation via temporal co-denoising.
\newblock {\em arXiv preprint arXiv:2305.18264}, 2023.

\bibitem{wang2023modelscope}
J.~Wang, H.~Yuan, D.~Chen, Y.~Zhang, X.~Wang, and S.~Zhang.
\newblock Modelscope text-to-video technical report, 2023.

\bibitem{wang2023cogvlm}
W.~Wang, Q.~Lv, W.~Yu, W.~Hong, J.~Qi, Y.~Wang, J.~Ji, Z.~Yang, L.~Zhao, X.~Song, et~al.
\newblock Cogvlm: Visual expert for pretrained language models.
\newblock {\em arXiv preprint arXiv:2311.03079}, 2023.

\bibitem{wang2024videocomposer}
X.~Wang, H.~Yuan, S.~Zhang, D.~Chen, J.~Wang, Y.~Zhang, Y.~Shen, D.~Zhao, and J.~Zhou.
\newblock Videocomposer: Compositional video synthesis with motion controllability.
\newblock {\em Advances in Neural Information Processing Systems}, 36, 2024.

\bibitem{wang2004image}
Z.~Wang, A.~C. Bovik, H.~R. Sheikh, and E.~P. Simoncelli.
\newblock Image quality assessment: from error visibility to structural similarity.
\newblock {\em IEEE transactions on image processing}, 13(4):600--612, 2004.

\bibitem{wolf-etal-2020-transformers}
T.~Wolf, L.~Debut, V.~Sanh, J.~Chaumond, C.~Delangue, A.~Moi, P.~Cistac, T.~Rault, R.~Louf, M.~Funtowicz, J.~Davison, S.~Shleifer, P.~von Platen, C.~Ma, Y.~Jernite, J.~Plu, C.~Xu, T.~L. Scao, S.~Gugger, M.~Drame, Q.~Lhoest, and A.~M. Rush.
\newblock Transformers: State-of-the-art natural language processing.
\newblock In {\em Proceedings of the 2020 Conference on Empirical Methods in Natural Language Processing: System Demonstrations}, pages 38--45, Online, Oct. 2020. Association for Computational Linguistics.

\bibitem{xiao2018unified}
T.~Xiao, Y.~Liu, B.~Zhou, Y.~Jiang, and J.~Sun.
\newblock Unified perceptual parsing for scene understanding.
\newblock In {\em Proceedings of the European conference on computer vision (ECCV)}, pages 418--434, 2018.

\bibitem{xie2021segformer}
E.~Xie, W.~Wang, Z.~Yu, A.~Anandkumar, J.~M. Alvarez, and P.~Luo.
\newblock Segformer: Simple and efficient design for semantic segmentation with transformers.
\newblock {\em Advances in Neural Information Processing Systems}, 34:12077--12090, 2021.

\bibitem{xing2023dynamicrafter}
J.~Xing, M.~Xia, Y.~Zhang, H.~Chen, W.~Yu, H.~Liu, X.~Wang, T.-T. Wong, and Y.~Shan.
\newblock Dynamicrafter: Animating open-domain images with video diffusion priors.
\newblock {\em arXiv preprint arXiv:2310.12190}, 2023.

\bibitem{yang2022reco}
Z.~Yang, J.~Wang, Z.~Gan, L.~Li, K.~Lin, C.~Wu, N.~Duan, Z.~Liu, C.~Liu, M.~Zeng, and L.~Wang.
\newblock Reco: Region-controlled text-to-image generation.
\newblock In {\em Proceedings of the IEEE/CVF Conference on Computer Vision and Pattern Recognition}, 2023.

\bibitem{sung2022vladapter}
M.~B. Yi-Lin~Sung, Jaemin~Cho.
\newblock Vl-adapter: Parameter-efficient transfer learning for vision-and-language tasks.
\newblock In {\em CVPR}, 2022.

\bibitem{yin2023nuwa-xl}
S.~Yin, C.~Wu, H.~Yang, J.~Wang, X.~Wang, M.~Ni, Z.~Yang, L.~Li, S.~Liu, F.~Yang, J.~Fu, M.~Gong, L.~Wang, Z.~Liu, H.~Li, and N.~Duan.
\newblock {NUWA}-{XL}: Diffusion over diffusion for e{X}tremely long video generation.
\newblock In {\em Proceedings of the 61st Annual Meeting of the Association for Computational Linguistics (Volume 1: Long Papers)}, pages 1309--1320, Toronto, Canada, July 2023. Association for Computational Linguistics.

\bibitem{zhang2023adding}
L.~Zhang, A.~Rao, and M.~Agrawala.
\newblock Adding conditional control to text-to-image diffusion models.
\newblock In {\em Proceedings of the IEEE/CVF International Conference on Computer Vision}, pages 3836--3847, 2023.

\bibitem{ControlNet_Depth}
L.~Zhang, A.~Rao, and M.~Agrawala.
\newblock Adding conditional control to text-to-image diffusion models.
\newblock \url{https://huggingface.co/lllyasviel/sd-controlnet-depth}, 2023.

\bibitem{ControlNet_Canny}
L.~Zhang, A.~Rao, and M.~Agrawala.
\newblock Adding conditional control to text-to-image diffusion models.
\newblock \url{https://huggingface.co/lllyasviel/sd-controlnet-canny}, 2023.

\bibitem{zhang2023i2vgen}
S.~Zhang, J.~Wang, Y.~Zhang, K.~Zhao, H.~Yuan, Z.~Qin, X.~Wang, D.~Zhao, and J.~Zhou.
\newblock I2vgen-xl: High-quality image-to-video synthesis via cascaded diffusion models.
\newblock {\em arXiv preprint arXiv:2311.04145}, 2023.

\bibitem{zhang2023controlvideo}
Y.~Zhang, Y.~Wei, D.~Jiang, X.~Zhang, W.~Zuo, and Q.~Tian.
\newblock Controlvideo: Training-free controllable text-to-video generation.
\newblock In {\em ICLR}, 2024.

\bibitem{Zhao_2018_ECCV}
L.~Zhao, X.~Peng, Y.~Tian, M.~Kapadia, and D.~Metaxas.
\newblock Learning to forecast and refine residual motion for image-to-video generation.
\newblock In {\em Proceedings of the European Conference on Computer Vision (ECCV)}, September 2018.

\bibitem{zhao2024uni}
S.~Zhao, D.~Chen, Y.-C. Chen, J.~Bao, S.~Hao, L.~Yuan, and K.-Y.~K. Wong.
\newblock Uni-controlnet: All-in-one control to text-to-image diffusion models.
\newblock {\em Advances in Neural Information Processing Systems}, 36, 2024.

\bibitem{zhou2022magicvideo}
D.~Zhou, W.~Wang, H.~Yan, W.~Lv, Y.~Zhu, and J.~Feng.
\newblock Magicvideo: Efficient video generation with latent diffusion models.
\newblock {\em arXiv preprint arXiv:2211.11018}, 2022.

\end{thebibliography}
